\documentclass[journal]{IEEEtran}

\usepackage{authblk}
\usepackage{graphicx}
\usepackage{hyperref}
\usepackage{xurl}
\usepackage{bm}
\usepackage{amsmath}
\usepackage{amsfonts}
\usepackage{dsfont}
\usepackage{algorithm}
\usepackage{algpseudocode}
\usepackage{amssymb}
\newbox{\myorcidaffilbox}
\sbox{\myorcidaffilbox}{\large\includegraphics[height=1.45ex]{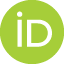}}
\newcommand{\orcidaffil}[1]{%
  \href{https://orcid.org/#1}{\usebox{\myorcidaffilbox}}}
\usepackage{chngcntr}
\usepackage{cite}
\usepackage{xcolor}
\usepackage{booktabs}
\usepackage{anyfontsize}
\usepackage{subcaption}
\usepackage{multirow}
\usepackage{stfloats}
\usepackage{arydshln}
\usepackage{stfloats}
\usepackage{mathtools}

\DeclarePairedDelimiter{\floor}{\lfloor}{\rfloor}
\DeclareMathOperator*{\argmax}{arg\,max}

\usepackage{etoolbox}
\makeatletter
\patchcmd{\@makecaption}
  {\scshape}
  {}
  {}
  {}
\makeatletter
\patchcmd{\@makecaption}
  {\\}
  {.\ }
  {}
  {}
\makeatother

\usepackage{fancyvrb}
\DefineVerbatimEnvironment{HighlightVerbatim}{Verbatim}
{fontsize=\small, frame=single}

\ifCLASSINFOpdf
\else
\fi

\begin{document}

\title{End-to-End Image Compression with Segmentation Guided Dual Coding for Wind Turbines} 

 \author{Raül~Pérez-Gonzalo\textsuperscript{\orcidaffil{0009-0007-6874-0709}},
Andreas~Espersen\textsuperscript{\orcidaffil{0009-0005-9835-541X}},
Søren~Forchhammer\textsuperscript{\orcidaffil{0000-0002-6698-8870}},~\IEEEmembership{Member,~IEEE},
and Antonio~Agudo\textsuperscript{\orcidaffil{0000-0001-6845-4998}}

  \thanks{R. Pérez-Gonzalo and A. Espersen are with the Wind Power LAB, 1150 Copenhagen, Denmark. Email: \{rperez@iri.upc.edu\}}
 \thanks{R. Pérez-Gonzalo and A. Agudo are with the Institut de Robòtica i Informàtica Industrial, CSIC-UPC, 08028 Barcelona, Spain.}
 \thanks{S. Forchhammer is with the Department of Photonics Engineering, Technical University of Denmark (DTU), 2800 Lyngby, Denmark.}
 }


\maketitle

\begin{abstract}
Transferring large volumes of high-resolution images during wind turbine inspections introduces a bottleneck in assessing and detecting severe defects. Efficient coding must preserve high fidelity in blade regions while aggressively compressing the background. In this work, we propose an end-to-end deep learning framework that jointly performs segmentation and dual-mode (lossy and lossless) compression. The segmentation module accurately identifies the blade region, after which our region-of-interest (ROI) compressor encodes it at superior quality compared to the rest of the image. Unlike conventional ROI schemes that merely allocate more bits to salient areas, our framework integrates: (i) a robust segmentation network (BU-Netv2+P) with a CRF-regularized loss for precise blade localization, (ii) a hyperprior-based autoencoder optimized for lossy compression, and (iii) an extended bits-back coder with hierarchical models for fully lossless blade reconstruction. Furthermore, our ROI framework removes the sequential dependency in bits-back coding by reusing background-coded bits, enabling parallelized and efficient dual-mode compression. To the best of our knowledge, this is the first fully integrated learning-based ROI codec combining segmentation, lossy, and lossless compression, ensuring that subsequent defect detection is not compromised. Experiments on a large-scale wind turbine dataset demonstrate superior compression performance and efficiency, offering a practical solution for automated inspections.
\end{abstract}


\IEEEpeerreviewmaketitle

\section{Introduction}

\IEEEPARstart{T}{he} maintenance of wind turbine blades constitutes a big challenge for the wind industry, having an impact on the entire value chain as blade defects can occur as a result of the manufacturing, transportation and installation process~\cite{maintenance}. On top of the production and installation challenges, the wear and tear of wind turbine blades over the 20-25 years of operation represents a significant cost to the industry~\cite{costs,costs2}. Blade defects can for most part be repaired, if discovered in time. Hence, there is a great commercial significance in being able to plan and react to blade defects in due time. In 2028, on-shore and off-shore wind turbine repairs are estimated to address a market of 3.2b dollars~\cite{woodmac} and 18b dollars~\cite{offshore-report}, respectively.

Wind turbine blade inspections are today used for repair planning. Drones are utilized to capture blade pictures which are later transferred to specialists, who process the data and detect blade defects in order to design the repair campaign~\cite{ai-drone}. The development of a faster service from inspection to decision making will have a direct positive impact on the wind industry~\cite{downtime,downtime2}. Time from inspection to repair is crucial as curtailed or stopped wind turbines do not produce electricity.

The large amount of image data generated in wind turbine inspections triggers a bottleneck delivering the assessments. A drone inspection captures $\sim$400 high-resolution images in 15min often in places with low-bandwidth connection~\cite{15min}. A common use case would be uploading the data from an off-shore windfarm. In these cases, current solutions imply delivering the blade assessments later and, therefore, delaying the blade repair campaign, which in the worst case causes loss in production. Other solutions imply compressing the images too much, which reduces its quality excessively. In consequence, structural defects captured in these images could become unrecognizable, making blade inspections worthless.

Since increasing the communication bandwidth is often not a feasible solution, efficient image compression is necessary without compromising the quality of the blade region. However, current industry solutions such as JPEG2000~\cite{J2K} have lower performance than learning-based image coders. For instance, \cite{hyperprior,elic,mlic++} neural lossy methods exhibit significantly better rate-distortion performance than JPEG2000~\cite{J2K}. These works have inspired state-of-the-art learning-based region-of-interest (ROI) coders~\cite{related-ROI-2,related-ROI-3,related-ROI-4,related-ROI-5}. Despite their efficiency, none of these algorithms support lossless coding. 

Therefore, we propose a learning-based ROI coding framework that supports dual compression: lossy and lossless. These compression algorithms have been combined with a segmentation model to derive our end-to-end ROI-eML coder: an entirely Machine Learning (eML) driven algorithm  designed and adapted completely to blade imagery through learning.  

\begin{figure*}[t!]
\centering
\resizebox{18cm}{!}{
\includegraphics[width=7in]{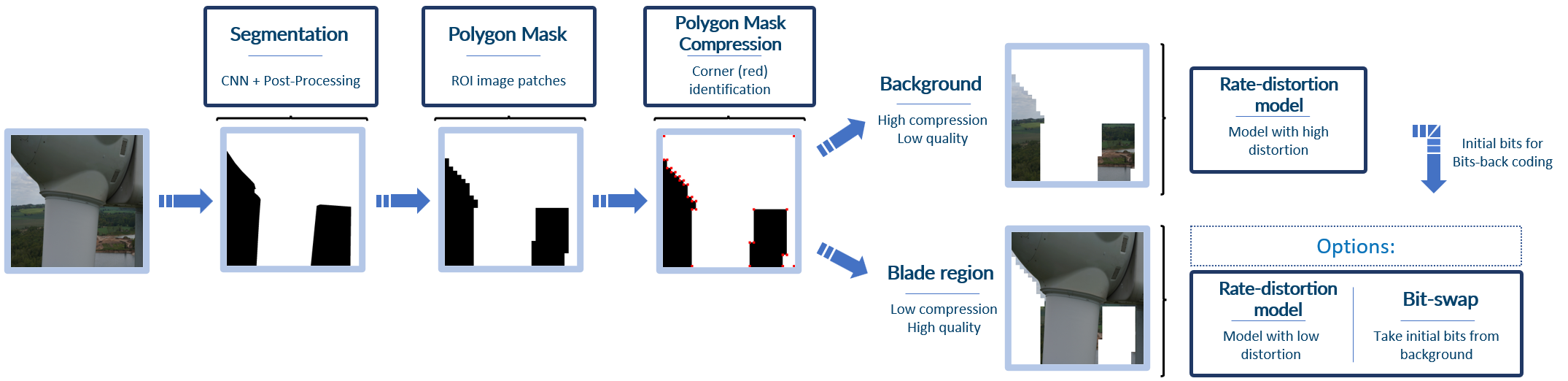}}
\caption{\textbf{Overview of our ROI algorithm for blade image compression}.  Our method begins by extracting a segmentation mask using our proposed BU-Netv2+P. This mask is converted into a polygonal representation by classifying image patches as either blade (ROI) or background. The polygon mask is then efficiently encoded via its corner points. The background is highly compressed with our lossy coder, while the blade region is coded either lossily --in better quality and less distortion-- or losslessly. In ROI lossless case, the background is first encoded to provide its lossy bitstream for parallel bits-back coding.}
\label{fig:roi-schema}
\vspace{-0.4cm}
\end{figure*}

First, the blade region is identified through segmentation by means of a BU-Netv2+P (Blade U-Net with a post-processing strategy) algorithm, inspired by~\cite{PerezGonzaloIcip2023}. A preliminary mask is obtained using a customized BU-Net model, thereafter, this is refined through hole filling~\cite{fill} and grouped into  blade surface to train a simple random forest model~\cite{randomforest}, which maps an RGB pixel and its surrounding neighbors to a binary pixel from the preliminary masks. The trained model is used to generate new masks to ensure an accurate detection of the blade. Also, it enhances a major robustness along the distinct blade surface images. To analyze and compare the robustness of our solution, we introduce the acceptance-ratio curves to demonstrate the feasibility of our segmentation algorithm. 

Then, the image is divided into rectangular patches, classified as blade or background, generating a polygonal mask coded through its corners. As shown in Fig.~\ref{fig:roi-schema}, the blade patches are compressed either lossily introducing some minor distortion or in lossless mode to avoid adversely affecting defect detection. An additional solution that our framework supports is to just compress the blade region as a special case.

The lossy compression algorithm is implemented by means of a variational autoencoder-like model based on non-linear convolutional filters. This method is inspired by the approach presented in HP+EASN~\cite{easn}. The rate-distortion trade-off is optimized end-to-end by applying uniform noise to continuously approximate the rounding quantizer. The distortion term is calculated through the reconstructed image obtained from the autoencoder and the rate term is estimated using a factorized prior probability model over the latent space~\cite{hyperprior}. 

Bits-back coding~\cite{bits-back} leverages hierarchical latent variable models~\cite{hpl} to develop a lossless coder. Similarly to BB-ANS~\cite{bb-ans}, it chains bits-back coding over the distinct image patches of a range Asymmetrical Numeral Systems (ANS) stream~\cite{ANS}. Bits-back coding is recursively applied to each latent variable and its encoding and decoding operations are rearranged to minimize the required initial bits, obtaining Bit-swap~\cite{bit-swap}. In our case, bits-back coding takes advantage of the coded background region by using it as the initial bits required. This releases Bit-swap from the sequential dependency when coding the image patches and enables our algorithm to code the blade region using Bit-swap in parallel.

The experimental results prove the feasibility of our segmentation algorithm. In addition to that, the viability of the lossy and lossless implementations are demonstrated in terms of compression performance and computational time, comparing them with other state-of-the art solutions. Finally, we focus on our ROI coder by analyzing how the blade region influences the resulting compression and running performance.

In brief, the main contributions of this paper are as follows:
\begin{itemize}
    \item As far as we know, this is the first framework that integrates both lossy and lossless learning-based compression. We gather these two algorithms with the segmentation model in an end-to-end manner, obtaining the dual learning-based ROI-eML coder.
    \item We introduce a novel strategy that removes the sequential dependency in bits-back lossless coding by reusing background-coded bits, enabling parallelized and time-efficient compression of the blade region.  
    \item We enhance the BU-Net+P~\cite{PerezGonzaloIcip2023} segmentation model with CRF regularization and on-the-fly random forest ensembling. We further introduce acceptance-ratio curves as a new reliability resource to assess robustness, and propose an efficient segmentation mask-coding algorithm.
    \item We present, for the first time, a fully operational learning-based dual-mode ROI compression pipeline by integrating entropy coding to produce actual bit rates and runtime results. This confirms the framework’s practicality for large-scale wind turbine inspection applications.  
\end{itemize}

The remainder of the paper is organized as follows. In Section ~\ref{sec:related-work}, we present the related learning-based image segmentation and compression methods along with popular traditional coders. After that, Sections~\ref{sec:method-segmentation}-\ref{sec:method-roi} include the underlying details of our image segmentation and compression framework. This part is subdivided into our four implementations: the segmentation algorithm, and the lossy, lossless and ROI algorithms. Sections~\ref{sec:seg-results}-\ref{sec:roi-results} analyze the performance of these four methods. Lastly, Section~\ref{sec:conclusion} outlines our conclusions.

\vspace{-0.3cm}
\section{Related Work} \label{sec:related-work}
\subsection{Image Segmentation} \label{sec:related-segmentation}
Contemporary image segmentation is predominantly grounded in learning-based models, with encoder-decoder architectures forming the backbone of state-of-the-art approaches~\cite{deeplabv3+,sw}. The introduction of U-Net~\cite{unet} marked a turning point in segmentation~\cite{pdda,colorspace}, as it captures high-level context and local details with skip connections. More recently, attention has been incorporated to refine feature extraction and improve segmentation accuracy~\cite{mask2former,cvpr_attention}. Innovative designs such as ResNeSt~\cite{resnest}, which integrates channel-wise attention with multi-path representations, and U-NetFormer~\cite{unetformer}, which applies global-local attention over the U-Net decoder. Advanced architectures now achieve high precision even with lightweight models, achieving favorable trade-offs between accuracy and efficiency in resource-constrained environments~\cite{chitty2023survey,mobilevit,efficientformer}. Furthermore, zero-shot models~\cite{clipseg,sam} demonstrate the flexibility of transformer architectures in achieving robust segmentation across various domains without domain-specific adjustments~\cite{diffseg}.

\vspace{-0.3cm}
\subsection{Image Coding}
There is a plethora of traditional compression algorithms to address the ever-growing demand for efficient image storage and transmission. In general, traditional image coders involves a handmade design to pursue efficient compression. Notably, some worth noting are PNG~\cite{png}, JPEG2000~\cite{J2K}, WebP~\cite{webp}, HEVC Intra~\cite{hevc}, BPG~\cite{bpg}, VTM~\cite{vtm} and JXL~\cite{jxl}. These routines rely on distinct techniques such as dictionary coders, prefix coding, DCT and wavelet, arithmetic coding, predictive coding, variable length coding, chroma subsampling, context modeling, intra prediction, progressive decoding among others.

In recent years, the use of deep learning algorithms has received great attention in lossy image compression. In \cite{rnn}, the authors propose an autoencoder-based recurrent neural network with a binarized latent space that supports variable compression rates for small image patches. The mentioned method just focuses on optimizing the distortion for a given compression rate, while other models more related to our work are end-to-end trained through joint rate-distortion optimization. \cite{entropy-fixed} proposes a non-trainable predefined function to estimate the rate term. Other studies train a probability model for entropy estimation and apply different continuous approximations for quantization~\cite{bls2017,hyperprior}. Autoregressive models such as JA~\cite{ja} add complexity to the training architecture by predicting image context sequentially. Some other works incorporate a non-local attention mechanism~\cite{compression-cvpr25,gmm,mlic++}, conditional probabilities for context modeling~\cite{compression-iccv25,compression-neurips24,elic} or channel-wise context~\cite{mlic++,lic-tcm,elic}.

Deep learning-based lossless image compression primarily relies on predictive coding, either at the pixel level~\cite{lossless-cvpr24,llicti} or in the frequency domain~\cite{lc-fdnet}, modeling the input image $\mathbf{x}$ as a sequence of pixels through autoregressive methods~\cite{pixelcnn,lcic-duplex}. More recent efforts enhance performance by integrating lossy coders with residual modeling, as in DLPR~\cite{dlpr}. In contrast, our method avoids explicit prediction and instead exploits the image distribution for efficient compression. Specifically, we model the conditional data distribution using auxiliary information $\mathbf{z}$ for entropy coding. Our approach employs a hierarchical autoencoder that learns feature representations $\mathbf{z}$, whose joint distribution with the image is captured via non-autoregressive predictors~\cite{lossless-neurips24}. In~\cite{integer-related}, $\mathbf{z}$ is derived as an invertible discrete function of the image and encoded using a fixed prior. Bits-back coding has also been successfully implemented using ANS and VAE latent variables~\cite{bb-ans}, with further improvements from a hierarchical latent model in~\cite{bit-swap}.

Less attention has been given to develop neural ROI coding algorithms. None of the existing methods can apply lossless coding to the ROI and they just modulate the distortion in exchange for the bit rate~\cite{related-ROI-0}. First works append an input channel with the ROI mask to the model architecture~\cite{related-ROI-1}, while newer versions infer the ROI mask by the same compression network~\cite{related-ROI-2}. More recent work develops region-adaptive transforms that use segmentation priors to apply different transforms per region~\cite{roi-segmentation1,roi-segmentation2}. Transformer-based~\cite{roi-transformer} and prompt-conditioned codecs~\cite{roi-text,roi-text2} support variable-rate ROI control, enabling intuitive joint control of rate and ROI via mask and rate tokens, respectively. Another applicability of ROI coding to improve machine-vision tasks by allocating more bits to certain regions~\cite{related-ROI-machine1,related-ROI-machine2}.

\vspace{-0.15cm}
\section{BU-Netv2+P: Blade Segmentation} \label{sec:method-segmentation}

This section presents our supervised blade-aware segmentation approach. To this end, we consider a learning-based model that is further refined with a post-processing algorithm. Our improved version, BU-Netv2, builds upon BU-Net~\cite{PerezGonzaloIcip2023} by integrating Dense CRF regularization and on-the-fly random forest ensembling optimized on the same blade surface images.

\begin{figure*}[t!]
\centering
\resizebox{17.8cm}{!}{
\includegraphics[width=4.8in]{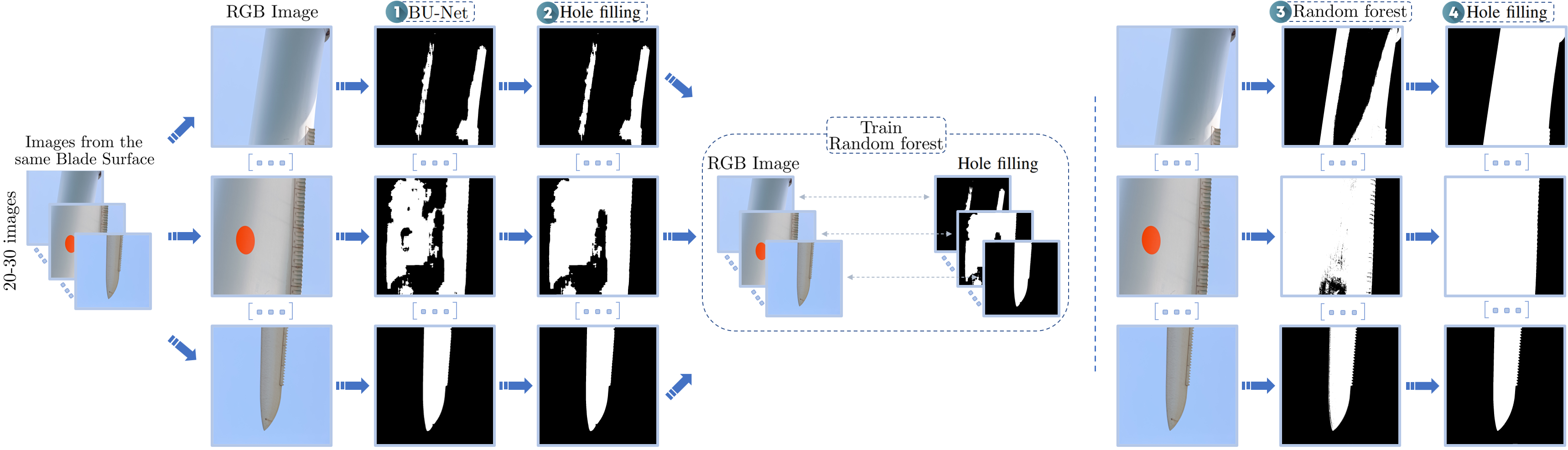}}
\vspace{-0.1cm}
\caption{\textbf{BU-Netv2+P algorithm pipeline to segment blades}: 1) an enhanced BU-Netv2, 2) a hole filling step, 3) a random forest module that exploits the input image set and the preliminary solution and, 4) a repeated hole filling operation.} 
\label{fig:pipeline}
\vspace{-0.5cm}
\end{figure*}

The BU-Netv2 masks are refined through three post-processing modules: a hole filling step, a random forest block that ensembles its outputs with the preliminary BU-Netv2 masks, and a latter hole filling step (identical to the first). A visual representation of this algorithm is displayed in Fig.~\ref{fig:pipeline}. Next, we will introduce every component in our framework.

\vspace{-0.3cm}
\subsection{Problem Formulation}
Let $\mathbf{x}_{i,j} \in \{0, \ldots, 255\}^{H \times W \times 3}$ be the $j$-th image captured from the $i$-th blade surface of a specific wind turbine, with a total of $J_i$ images for that specific blade surface. We aim at identifying the region of the images where the blade is localized. Given the ground-truth segmentation mask $\mathbf{s}_{i,j} \in \{0,1\}^{H \times W}$, we propose a dense-prediction algorithm composed of four modules to estimate the mapping $\mathbf{x}_{i,j}\rightarrow{\mathbf{s}_{i,j}}$. In our formulation, we denote $\hat{\mathbf{s}}_{i,j}$ as our prediction. For simplicity, the modules where the blade surface information is not leveraged will not include these subindixes.

\vspace{-0.35cm}
\subsection{Design and Optimization of the BU-Netv2 Model} 
We implement a customized BU-Net module to classify the input pixels $x_{h,w}$ into the binary entries $\hat s_{h,w}^{BU}$, where $h=\{1,\ldots,H\}$ and $w=\{1,\ldots,W\}$ denote the subindices of the pixel coordinates. The BU-Netv2 follows a standard convolutional architecture composed of four encoder and decoder blocks with information flow through its skip connections. The training dataset follows the work presented in~\cite{PerezGonzaloIcip2023} and is augmented on-the-fly by flipping rotations and a set of small rotations followed by the corresponding zoom; details in Supplementary Section II A-B. Similarly, the output mask is enhanced by averaging the predictions of the four flipping rotations. Therefore, the final output of this block ${\hat{\mathbf{s}}}^{BU}$ is obtained by quantifying a probability map of the blade class by a threshold $\tau^{BU}$. This preliminary mask is later fed as the input of our post-processing algorithm.

We implement a customized BU-Net module to classify the input pixels $x_{h,w}$ into binary predictions $\hat{s}^{BU}_{h,w}$, where $h \in \{1, \ldots, H\}$ and $w \in \{1, \ldots, W\}$ denote the pixel coordinates. The BU-Netv2 uses a standard encoder-decoder architecture with skip connections following~\cite{PerezGonzaloIcip2023}. The training data is augmented on-the-fly using flipping and small affine transformations (rotations and zoom). To increase robustness during inference, we apply the trained model to the four flipped versions of the input image $\mathbf{x}$: original, horizontally flipped, vertically flipped, and both. Let $T_i(\mathbf{x})$ denote the $i$-th flipped input, and let $\hat{\mathbf{s}}^{(i)}$ be the corresponding predicted mask. The final prediction is obtained by averaging the inverse-transformed outputs $\hat{\mathbf{s}}^{BU} = \frac{1}{4} \sum_{i=1}^{4} T_i^{-1} \left( \hat{\mathbf{s}}^{(i)} \right)$, where $T_i^{-1}(\cdot)$ denotes the inverse flipping transformation.

However, we have to adapt the loss function to improve the experimental results with this type of dataset. In particular, our loss function contains two terms: a weighted focal loss to assess the quality of the segmentation, and a dense CRF regularization loss to obtain a smooth solution.

\subsubsection{Focal Loss} \label{sec:focal-loss}
Let $\sigma$ denote the sigmoid function, we weight each pixel by the confidence we have in the prediction of that pixel~\cite{focal}. In particular, $\gamma$ is the parameter that controls the rate by reducing the loss of easy cases. Additionally, to hinder benefiting the majority class (the background) relative to the minority one, we balance each class by the weight $\alpha$:

\vspace{-0.3cm}
\begin{align*} 
    \mathcal L_{focal}(\mathbf{s}, \mathbf{\hat s} ) = -\sum_{h,w} \left[ \alpha \left(1-\sigma(\hat s_{h,w})\right)^\gamma s_{h,w}\log\sigma(\hat s_{h,w})  \right. \nonumber \\ 
   \left.  + (1 - \alpha) \left(\sigma(\hat s_{h,w}))\right)^\gamma (1-s_{h,w})\log(1-\sigma(\hat s_{h,w}))    \right].
\end{align*}

\subsubsection{Dense CRF Loss} \label{sec:crf-loss} 
Additionally, a contiguity penalty is added by means of an embedded CRF loss~\cite{denseCRF} through the pairwise potentials $\psi$. In particular, we model pairwise potentials as weighted Gaussian kernels:
{\small \begin{equation} \label{eq:pairwise-potentials}
 \psi \left( \hat s_{h,w}, \hat  s_{\tilde h, \tilde w} \right) = \left[ \hat \sigma^{\tau}( \hat s_{h,w} ) \neq \sigma^{\tau} ( \hat s_{\tilde h, \tilde w}  ) \right] \langle \mathbf{\rho} , \mathbf{k} (\mathbf{\hat f}_{h,w}, \mathbf{\hat f}_{\tilde h, \tilde w}) \rangle ,
\end{equation}} 
where $\mathbf{\rho}$ are linear combination weights, $\mathbf{k}$ is a Gaussian kernel, and $[ \cdot \hspace{-0.1cm} \neq \hspace{-0.1cm} \cdot ]$ indicates the Potts model as the label compatibility function. For simplicity, we have omitted the subindices $h,w,\tilde h, \tilde w$ that determinate $\mathbf{\rho}$. $\sigma^{\tau}$ binarizies the output of the sigmoid function by thresholding it with a predefined parameter $\tau^{BU}$. Moreover, $ \mathbf{\hat f}_{h,w} $ denotes a 5-dimensional feature vector that incorporates the spatial location $(h, w)$ and the color intensity of $\mathbf{x}$. The degrees of nearness and similarity are adjusted by dividing the feature components by the Gaussian bandwidth parameters $\eta_{\texttt{loc}}$ and $\eta_{\texttt{rgb}}$, respectively. 

To make Eq.~\eqref{eq:pairwise-potentials} differentiable, we relax the Potts model following the strategy presented in~\cite{relaxed-crf} as: 
\begin{equation*} 
\mathcal L_{CRF}(\mathbf{\hat s}) =  \sum_{\tilde h,\tilde w} \sum_{h,w}  \sigma( \hat s_{h,w}) (1- \sigma(\hat s_{\tilde h, \tilde w})) \langle \mathbf{\rho} , \mathbf{k} (\mathbf{\hat f}_{h,w}, \mathbf{\hat f}_{\tilde h, \tilde w}) \rangle .
\end{equation*}

\vspace{-0.1cm}
The pairwise potentials provide an image-dependent smoothing term that encourages assigning similar labels to pixels that resemble in location and color intensity. The CRF loss requires calculating $H^2W^2$ terms, therefore, to reduce its prohibitive cost, we follow a mean-field approximation strategy to compute fast high-dimensional Gaussian filtering~\cite{permutohedral}. Note that despite this complexity, we can leverage the graphical model CRF without increasing the inference time, thanks to only being calculated during the training.

\subsubsection{Total Loss} \label{sec:total-loss} 
Both focal and dense CRF losses are linearly combined to define our training objective, where $\lambda$ adjusts the load of the CRF regularization term.

\vspace{-0.2cm}
\subsection{Post-processing scheme}
\subsubsection{Hole Filling} \label{hole_filling}
\begin{figure}[!t]
\centering
\includegraphics[width=3.3in]{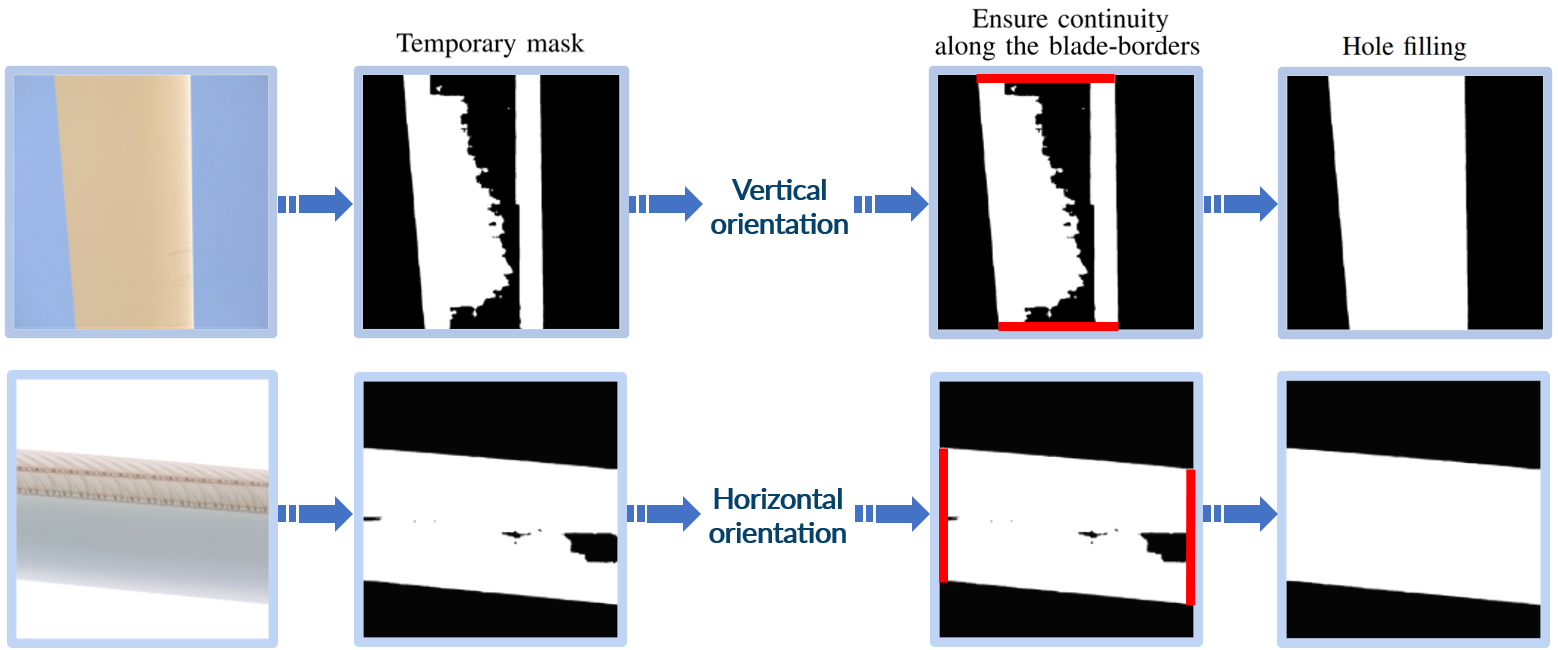}
\caption{\textbf{Hole filling scheme.} Example of artifact removal inside the segmented blade region ($\mathbf{\hat s}^{BU}$). The method detects blade orientation to locate border edges (in red) and applies a standard hole-filling algorithm~\cite{fill} to produce the mask $\mathbf{\hat s}^{H}$.} 
\label{fig:holefilling}
\vspace{-0.5cm}
\end{figure}

The BU-Netv2 mask $\mathbf{\hat s}^{BU}$ can include non-realistic predictions, including some irregularities inside the blade region (see Fig.~\ref{fig:holefilling}). By leveraging our spatial prior that a solution of this type is not possible, we apply an enhanced hole filling algorithm to suppress these artifacts. Taking inspiration from~\cite{fill}, our approach addresses the holes that are surrounded by blade pixels and image borders, in contrast of only filling in the holes that are entirely surrounded by blade pixels. To this end, we focus on localizing first the blade pixels on the image boundaries: the blade-borders. 

As we know the blade traverses the image horizontally (from left-to-right) or vertically (top-to-bottom), we can calculate its orientation by means of an accumulated gradient. Diagonal orientations are not present in our dataset to a degree that would compromise the hole filling logic, i.e., where the blade spans from one image corner to the opposite one. In case the accumulated gradient along the $x$-axis is higher (lower) than along the $y$-axis, the blade orientation is estimated as vertical (horizontal). In particular, we employ the Sobel filter to compute the accumulated gradients. 

In our dataset (see Section I.A in Supplementary), blade images are predominantly aligned either horizontally (left-to-right) or vertically (top-to-bottom), with only slight deviations. 

Once the orientation is identified, we can distinguish where the blade image borders are. Therefore, we ensure those borders are continuously connected by replacing the background pixels within the blade area by blade pixels. Ultimately, we carry on with the standard hole filling algorithm~\cite{fill} to generate the output $\mathbf{\hat s}^{H}$ of this step. A schematic illustration of the hole filling procedure is pictured in Fig.~\ref{fig:holefilling}. Due to its robustness, the hole filling step is applied also after performing the random forest step that it will be introduced below. Therefore, we distinguish the hole filled BU-Netv2 mask and random forest one by the notation $\mathbf{\hat s}^{H1}$ and $\mathbf{\hat s}^{H2}$, respectively. 

\subsubsection{Random Forest Ensemble} \label{sec:random_forest}
To generate smooth estimations, we propose leveraging the temporary masks $\mathbf{\hat s}^{H1}$ to train an unsupervised random forest model, which is later combined with the BU-Netv2 estimations via soft-voting. Specifically, our approach aims to overcome the inability of previous steps to capture global context by learning consistent mapping patterns across all images of the same blade surface.

In particular, the training data for the random forest corresponds to the set of images $\{\mathbf{x}_{i,j}\}_{j \in J}$ and their hole-filled masks $\{\mathbf{\hat s}^{H1}_{i,j}\}_{j \in J}$ for a fixed blade index \(i\). These images share similar properties, such as illumination, contrast, and background texture, since they belong to the same blade and drone inspection. This local homogeneity reduces the complexity of the underlying distribution, allowing the random forest to act as a robust denoising model that can generalize to difficult cases not well handled by BU-Netv2 alone.

The random forest learns whether a pixel $x_{h,w}^{H1}$ and its local neighborhood $\texttt{neigh}( x_{h,w}^{H1})$ belong to the blade region, and outputs the probability:

\vspace{-0.5cm}
\begin{equation*}
    p^{RF}\big(\hat s_{h,w} \big) = \frac{1}{|RF|} \sum_{t \in RF} p^t\big( \hat s_{h,w} | x_{h,w}^{H1},  \texttt{neigh}( x_{h,w}^{H1}) \big)\text{ } ,
\end{equation*}
\vspace{-0.4cm}

\noindent
where $RF$ is the set of decision trees $t$. It is worth pointing out that this model only needs to be adapted to some image features (such as brightness and contrast) of the images of a specific $i$-th blade. As a consequence, that factor reduces substantially the complexity of the distribution to be learned. Hence, the random forest acts as a denoising step; besides, it provides a fine-grained solution along the boundaries.

Once the model is trained, it is applied to obtain the probability estimations $p^{RF}(\mathbf{\hat s}_{i,j})$ which are combined with the BU-Netv2 probability maps $\sigma(\mathbf{\hat s}_{i,j})$ via soft voting. This ensembling represents the output $\hat{\mathbf{s}}^{RF}_{i,j}$ of this step in our algorithm after thresholding it by $\tau^{RF}=0.37$. 

Ultimately, as this step could include some small holes in the solution especially in images with blade shadows, without loss of generality we apply a new hole filling step (see Section~\ref{hole_filling}) to refine the solution, obtaining $\hat{\mathbf{s}}\equiv\hat{\mathbf{s}}^{H2}$.

\vspace{-0.15cm}
\section{HP+EASN-deep: Lossy Coding} \label{sec:method-hp+easn}

As our ROI algorithm can be combined with any neural lossy coder (see Section~\ref{sec:method-roi}), we present the methodology necessary to implement such an end-to-end coder. We introduce end-to-end rate-distortion optimization which combines a non-linear analysis transform, a uniform quantizer and a non-linear synthesis transformation. We show how variational autoencoders can be applied to learn compressed latent representations, and demonstrate that uniform noise successfully continuously approximates a rounding quantizer. Section III-A and III-B from Supplementary details the training specifications for HP+EASN-deep~\cite{easn} and its discretization of the continuous latent space to apply entropy coding, respectively.

\vspace{-0.35cm}
\subsection{Design and Optimization of the Variational Model} \label{sec:hp+easn-model}
\subsubsection{Relaxed Rate-Distortion Optimization} \label{sec:rate-distortion}

Given an encoder-decoder architecture, we would like to balance the compression rate with the decompression error. Let $\mathbf{x}$ be the input image, we aim to minimize the rate-distortion trade-off by optimizing the encoder $h_{\boldsymbol{\phi}}$ and the decoder $g_{\boldsymbol{\theta}}$ as:
\begin{align}  \label{eq:rate-distortion}
 &- \mathbb{E}_{p_{\boldsymbol{\theta}}(\mathbf{x})} \left[ \log p_{\boldsymbol{\phi}}(\mathbf{z}) \right] + \zeta \ \mathbb{E}_{p_{\boldsymbol{\theta}}(\mathbf{x})} \left[  d(\hat{\mathbf{x}}, \mathbf{x}) \right]   
\\ = &- \mathbb{E}_{p_{\boldsymbol{\theta}}(\mathbf{x})} \left[ \log p_{\boldsymbol{\phi}}(h_{\boldsymbol{\phi}}(\mathbf{x})) \right] + \zeta \ \mathbb{E}_{p_{\boldsymbol{\theta}}(\mathbf{x})} \left[ d(g_{\boldsymbol{\theta}}(h_{\boldsymbol{\phi}}(\mathbf{x})), \mathbf{x}) \right], \nonumber 
\end{align} 
where $\mathbf{z} = h_{\boldsymbol{\phi}}(\mathbf{x})$ is the latent variable after encoding $\mathbf{x}$, $\mathbb{E}[\cdot]$ is the expectation operator, $p_{\boldsymbol{\theta}}$ and $p_{\boldsymbol{\phi}}$ are the probabilistic models of $\mathbf{x}$ and $\mathbf{z}$ with their respective weights $\boldsymbol{\theta}$, $\boldsymbol{\phi}$, $d(\hat{\mathbf{x}}, \mathbf{x})$ indicates a distortion metric between the reconstructed $\hat{\mathbf{x}} = g_{\boldsymbol{\theta}}(\mathbf{z})$ and input image $\mathbf{x}$, and $\zeta$ is a scalar that governs the trade-off between the compression ratio and the distortion. 

To combine the autoencoder with an entropy coder, continuous latent representations are quantized to a finite set of discrete values $Q(\mathbf{z}) = \bar{\mathbf{z}}$. Let $\bar{p}_{\boldsymbol{\phi}}(\bar{\mathbf{z}})$ be the corresponding quantized prior distribution, then the quantized rate-distortion trade-off can be written as $- \mathbb{E}_{p_{\boldsymbol{\theta}}(\mathbf{x})} \left[ \log \bar{p}_{\boldsymbol{\phi}}(\bar{\mathbf{z}}) \right] + \zeta \ \mathbb{E}_{p_{\boldsymbol{\theta}}(\mathbf{x})} \left[  d(\hat{\mathbf{x}}, \mathbf{x}) \right]$.

In this case, the reconstructed image differs from Eq.~\eqref{eq:rate-distortion} as it has been transformed from the latent variables $\hat{\mathbf{x}} = g_{\boldsymbol{\theta}}(Q(h_{\boldsymbol{\phi}}(\mathbf{x})))$. Note that the first term is minimized when the quantized prior $\bar{p}_{\boldsymbol{\phi}}(\bar{\mathbf{z}}) $ is identical to the true marginal distribution $\bar{p}_{\boldsymbol{\theta}}(\bar{\mathbf{z}}) $. However, due to the intractability of the true marginal distribution, we minimize a cross-entropy of these terms following a variational inference approach, i.e., by introducing an approximate prior $\bar{p}_{\boldsymbol{\phi}}(\bar{\mathbf{z}}) \approx \bar{p}_{\boldsymbol{\theta}}(\bar{\mathbf{z}})  $.

As optimal quantization is intractable ~\cite{quantization-intractable}, rather than attempting to optimize $Q(\mathbf{z})$, we take a uniform quantizer and expect that the model would fit appropriately. Specifically, we divide the latent space in unit bins and place the quantized values as the center points, such that $\bar{\mathbf{z}} = Q(\mathbf{z}) = \text{round}(\mathbf{z})$.

Introducing a uniform quantizer leads to discontinuities at integer values and zero gradients elsewhere. To use an automatic differentiation framework, we approximate the quantizer with additive uniform noise during the training. Independent uniform noise successfully models quantization error, because their marginal moments closely resemble~\cite{moments}. Let $\tilde{\mathbf{z}} = \mathbf{z} + \mathcal{U}(-\frac{1}{2},\frac{1}{2})$ denote the noisy approximation, the corresponding noisy prior $\tilde{p}_{\boldsymbol{\phi}}(\tilde{\mathbf{z}})$ is then calculated by convolving the prior density model with a noisy uniform distribution as:

\vspace{-0.4cm}
\begin{equation}
    \bigg( p_{\boldsymbol{\phi}}(\mathbf{z}) * \mathcal{U}(-\tfrac{1}{2},\tfrac{1}{2}) \bigg) (\tilde{\mathbf{z}}) 
     = c(\tilde{\mathbf{z}}+ \tfrac{1}{2}) - c(\tilde{\mathbf{z}}-\tfrac{1}{2}), \nonumber
\end{equation} 
where $c$ is the cumulative distribution function of the prior $p_{\boldsymbol{\phi}}(\mathbf{z})$. Since the prior is defined through a non-parametric cumulative density following a factorized probability model \cite{hyperprior}, we can effectively compute its continuous relaxation and, therefore, the corresponding relaxed loss during the training. Indeed, the noisy prior model $\tilde{p}_{\boldsymbol{\phi}}(\tilde{\mathbf{z}})$ is effectively a continuous proxy for the quantized distribution $\bar{p}_{\boldsymbol{\theta}}(\mathbf{z})$ at integer locations:

\vspace{-0.5cm}
\begin{equation} 
\tilde{p}_{\boldsymbol{\phi}}(\mathbf{z}) = c(\mathbf{z}+ \tfrac{1}{2}) - c(\mathbf{z}-\tfrac{1}{2}) = \bar{p}_{\boldsymbol{\theta}}(\mathbf{z}) \iff \mathbf{z} \in \mathbb{Z}^M . \nonumber 
\end{equation}
\vspace{-0.4cm}

By adapting this continuous relaxation, we end up with the following trainable rate-distortion loss $\mathcal{L}_{\boldsymbol{\theta},\boldsymbol{\phi}}$:

\vspace{-0.25cm}
\begin{equation} 
\label{eq:relaxed-rate-distortion} 
- \mathbb{E}_{p_{\boldsymbol{\theta}}(\mathbf{x})} \left[ \log \tilde{p}_{\boldsymbol{\phi}}(\tilde{\mathbf{z}}) \right] + \zeta \ \mathbb{E}_{p_{\boldsymbol{\theta}}(\mathbf{x})} \left[  d(\hat{\mathbf{x}}, \mathbf{x}) \right] . 
\end{equation} 
\vspace{-0.4cm}

The relaxed rate-distortion function can be seen as the log likelihood of a variational model, as discussed in~\cite{bls2017}.

\begin{figure*}[!t]
\centering
\begin{tabular}{c:c}
    \hspace{0.2cm}\rotatebox{90}{\includegraphics[width=0.33\linewidth]{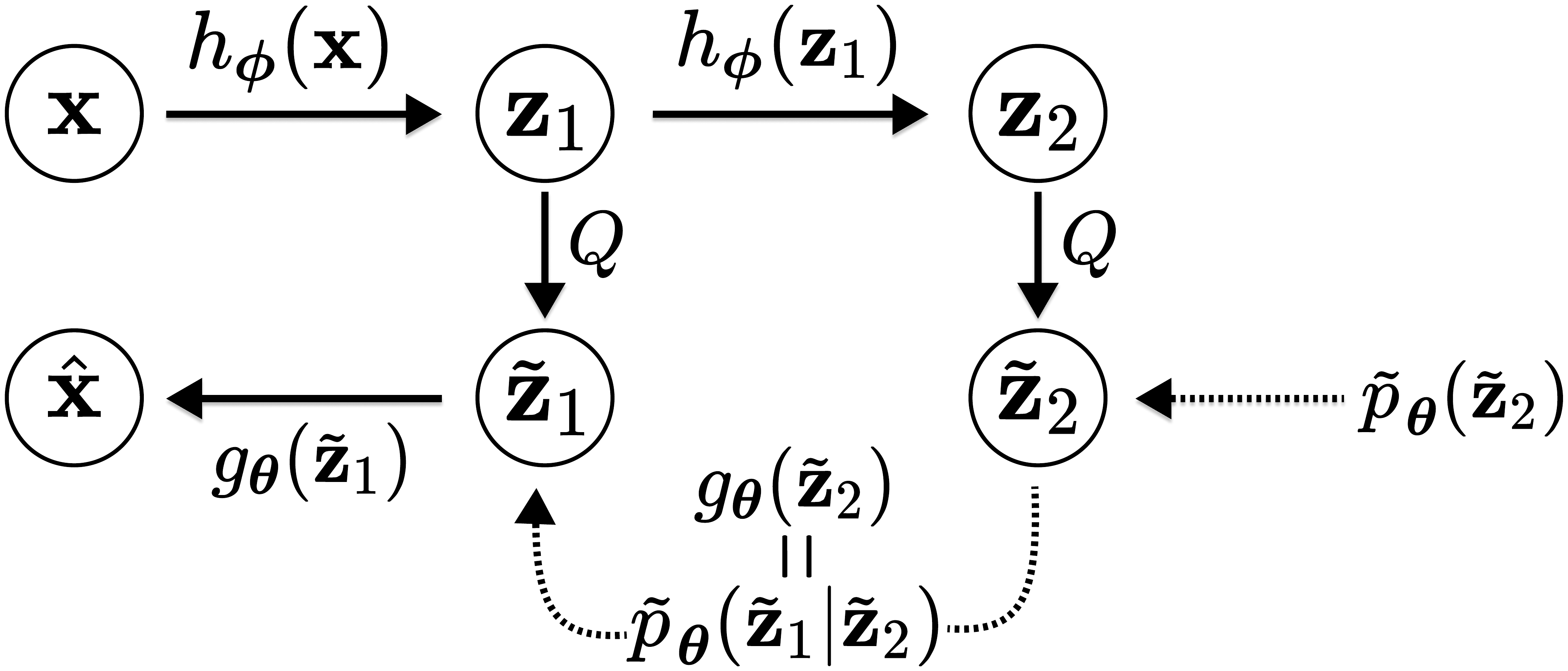}}\hspace{0.2cm} & \hspace{0.2cm}\includegraphics[width=0.78\linewidth]{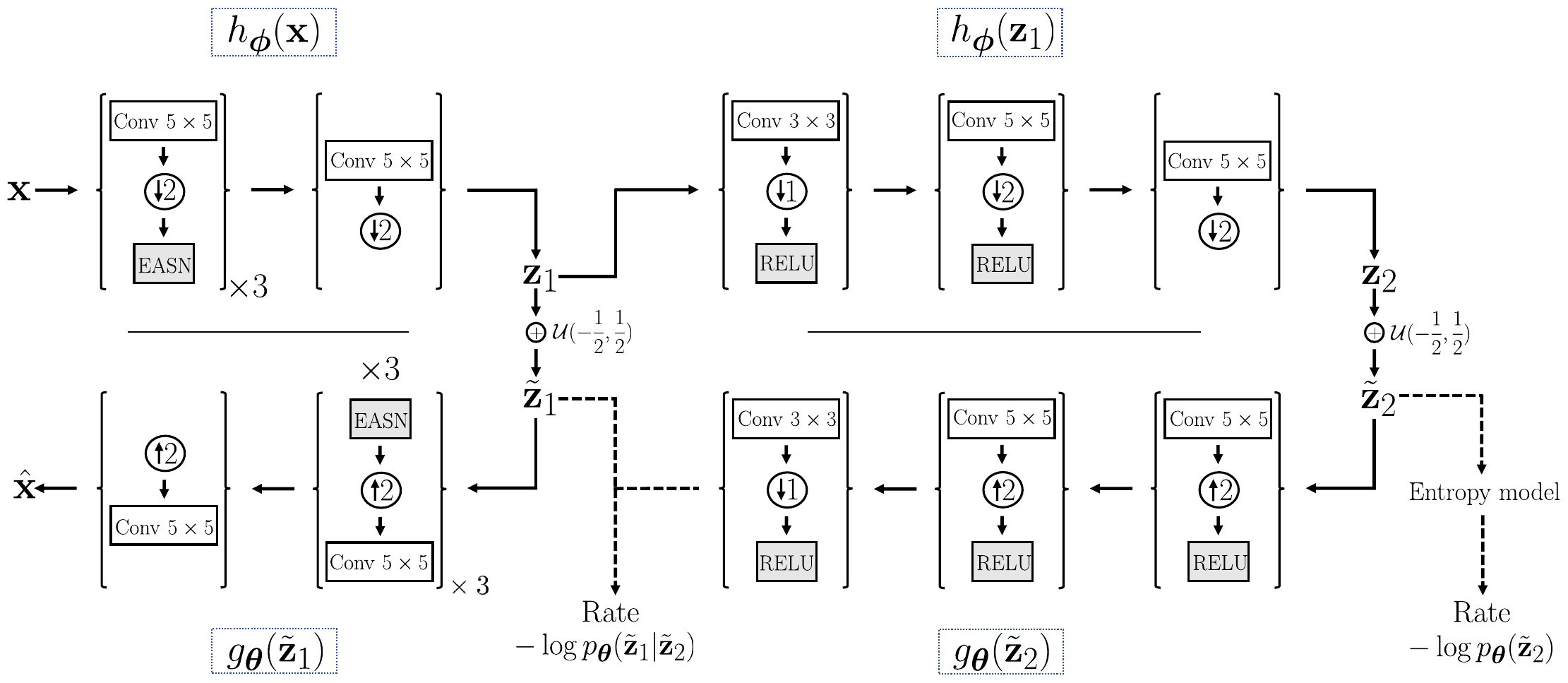} \hspace{0.2cm}\\
\end{tabular}%
\vspace{-0.25cm}
\caption{\textbf{Overview of HP+EASN-deep hierarchical model and architecture.} 
Left-side illustrates the 2-level nested variable model, where latent variables $\mathbf{z}_1$ and $\mathbf{z}_2$ are obtained during encoding $h_{\boldsymbol{\phi}}$ and quantized $Q$. Right-side shows the architecture, a symmetric 2-layer autoencoder composed of convolutional and up/downsampling blocks with non-linear activations.}
\label{fig:hyperprior-architecture}
\vspace{-0.45cm}
\end{figure*}

\subsubsection{Hierarchical Latent Variable Model}

By increasing the number of layers of latent variables, we can extend a typical variational inference model with a single latent variable. More precisely, common variational models can successfully approximate complex dependencies in $p_{\boldsymbol{\theta}}(\mathbf{x})$ by introducing a latent variable $\mathbf{z}_1$;  even when the prior $p(\mathbf{z}_1)$ and the likelihood $p_{\boldsymbol{\theta}}(\mathbf{x}|\mathbf{z}_1)$ distributions are relatively simple. In particular, the distribution $p_{\boldsymbol{\theta}} (\mathbf{x})$ is implicitly expressed in terms of the likelihood $p_{\boldsymbol{\theta}} (\mathbf{x} | \mathbf{z}_1)$ and the explicit prior $p(\mathbf{z}_1)$ as:

\vspace{-0.3cm}
\begin{equation*}
    p_{\boldsymbol{\theta}} (\mathbf{x}) = \int p_{\boldsymbol{\theta}} (\mathbf{x} | \mathbf{z}_1) p(\mathbf{z}_1) d\mathbf{z}_1 .
\end{equation*}

Instead of explicitly defining $p(\mathbf{z}_1)$, we can introduce an additional latent variable $\mathbf{z}_2$ to implicitly express $p(\mathbf{z}_1)$ using the likelihood $p_{\boldsymbol{\theta}} (\mathbf{z}_1 | \mathbf{z}_2)$ and the explicit prior $p(\mathbf{z}_2)$, obtaining:

\vspace{-0.3cm}
\begin{equation*}
p_{\boldsymbol{\theta}} (\mathbf{z}_1) = \int p_{\boldsymbol{\theta}} (\mathbf{z}_1 | \mathbf{z}_2) p(\mathbf{z}_2) d\mathbf{z}_2 .
\end{equation*}
\vspace{-0.4cm}

The new latent variable $\mathbf{z}_2$ can help to model the complex dependencies of $\mathbf{z}_1$. Consequently, we can gather the distinct latent variables $\{\mathbf{z}_1, \mathbf{z}_2 \}$ with a conditional nested structure, as shown in Fig. \ref{fig:hyperprior-architecture}-left, and express the model evidence as:

\vspace{-0.4cm}
\begin{align*}
    p_{\boldsymbol{\theta}} (\mathbf{x}) = \int p_{\boldsymbol{\theta}} (\mathbf{x} | \mathbf{z}_1) p_{\boldsymbol{\theta}} (\mathbf{z}_1 | \mathbf{z}_2)  p_{\boldsymbol{\theta}} (\mathbf{z}_2)  
    \ d\mathbf{z}_2 d\mathbf{z}_1 .
\end{align*}
\vspace{-0.4cm}

Introducing a conditional dependency of $\mathbf{z}_1$ using a nested variable $\mathbf{z}_2$ was introduced for neural lossy compression in \cite{hyperprior} and denoted as a hyperprior. In our case, the prior of the deepest latent variable $p(\mathbf{z}_2)$ is defined as a factorized prior probability model~\cite{bls2017}. Besides that, $p_{\boldsymbol{\phi}}(\mathbf{z}_1 | \mathbf{z}_2)$ is modeled as a conditional Gaussian distribution with zero mean. Both distributions are assumed to be statistically independent along its components. Due to incorporating two latent variables, a compressed representation of an image $\mathbf{x}$ will be comprised by the quantized latent variables ${\tilde{\mathbf{z}}_1}$ and ${\tilde{\mathbf{z}}_2}$, yielding to the following relaxed rate-distortion loss $\mathcal{L}_{\boldsymbol{\theta},\boldsymbol{\phi}} $ in Eq.~\eqref{eq:relaxed-rate-distortion}:

\vspace{-0.4cm}
\begin{equation}  - \mathbb{E}_{p_{\boldsymbol{\theta}}(\mathbf{x})} \left[ \log \tilde{p}_{\boldsymbol{\phi}}({\tilde{\mathbf{z}}_1} | {\tilde{\mathbf{z}}_2}) + \log \tilde{p}_{\boldsymbol{\phi}}({\tilde{\mathbf{z}}_2}) \right] + \zeta \ \mathbb{E}_{p_{\boldsymbol{\theta}}(\mathbf{x})} \left[  d(\hat{\mathbf{x}}, \mathbf{x}) \right] . \nonumber \end{equation}

\subsection{Model Architecture}
We model the encoder $h_{\boldsymbol{\phi}}$ and decoder $g_{\boldsymbol{\theta}}$ symmetrically to mathc input and output dimensionality. Both transforms are implemented as a cascade of three sequential blocks for each latent. Each encoder block is comprised of a convolution, a downsampling operation and a non-linear activation function. Here, we take the Expanded Adaptive Scaling Normalization (EASN) with learnable parameters as the non-linearity~\cite{easn}. Analogously, each decoder block is comprised of the inverse operations in reversed order: an EASN, an upsampling operation and a convolution. A complete picture of the network architecture is shown in Fig.~\ref{fig:hyperprior-architecture}-right. 

In particular, all utilized convolutions are 2-dimensional with a 5$\times$5 kernel. The padding hyperparameter of each convolution is chosen in such a way that the input and the output dimensions match. All convolutions operate with $128$ filters, except the ones interacting with $\mathbf{z}_1$ which operate with $192$ filters. In all blocks, we apply a 2-factor down/upsampling. To reduce computational time, the down/upsampling operations are implemented jointly with the linear convolutions. For instance, the downsampling operation can be easily incorporated by modulating the stride of the convolution. Training specifications are included in Supplementary Section IV-A.

\vspace{-0.35cm}
\subsection{Expanded Adaptive Scaling Normalization (EASN)}

An EASN~\cite{easn} is a non-linear function that incorporates local normalization via its learnable parameters, allowing coding transforms to successfully model local patterns of natural images. EASN strengthens GDN~\cite{gdn} learning capacity by increasing the degree of freedom of its parameters. This higher representational efficiency allows probabilistic models to better model complex data, even with shallower networks.

An EASN is typically applied after linearly transforming the input. It normalizes the resulting output $\mathbf{y}$ by a sigmoid function. Hence, it eradicates the even-symmetry of the GDN normalization function. EASN better exploits the spatial correlations by introducing an input mapping function $m$ and operates as a residual layer to facilitate the flow of gradients through the network as $\text{EASN}(\mathbf{y}) = m(\mathbf{y}) \cdot \sigma(\mathcal{F}(\mathbf{y})) + \mathbf{y}$, where $\mathcal{F}$ is a transform function that along with $m$ is optimized.

\vspace{-0.35cm}
\section{Bit-swap: Lossless Coding} \label{sec:method-bitswap}

To start with, Section~\ref{sec:bitswap-model} explains the neural lossless coder Bit-swap~\cite{bit-swap}, which is further parallelized in Section~\ref{sec:method-roi}. It employs hierarchical latent probabilistic models for granting more flexibility to model high-dimensional data thanks to the nested structure. We derive the hierarchical Evidence Lower BOund (ELBO) as the maximization goal. Finally, we extend the bits-back coding to nested latent variable and present a methodology to significantly reduce the required initial random bits. Detailed architecture, training procedure and discretization approach is included in Supplementary Material.

\vspace{-0.3cm}
\subsection{Design and Optimization of the Variational Model} \label{sec:bitswap-model}

\subsubsection{Hierarchical Evidence Lower Bound} 

We are interested in learning a probabilistic model of our observed data $p_{\boldsymbol{\theta}}(\mathbf{x})$, where  $\boldsymbol{\theta}$ denotes the model parameters. To deal with this learning problem, fully-observed models are extended into a model with deep hierarchical latent variables~\cite{deep-latent1,deep-latent2}, typically denoted by $\mathbf{z}_{1:L} = \{\mathbf{z}_1, \mathbf{z}_2, \ldots, \mathbf{z}_{L}\}$. By marginalizing over $\mathbf{z}_{1:L}$ following a nested structure,  $p_{\boldsymbol{\theta}}(\mathbf{x})$ can be represented as $L$ likelihood distributions and a prior distribution: 

\vspace{-0.4cm}
\begin{align*}
    p_{\boldsymbol{\theta}} (\mathbf{x}) = \int p_{\boldsymbol{\theta}} (\mathbf{x} | \mathbf{z}_1) p_{\boldsymbol{\theta}} (\mathbf{z}_1 | \mathbf{z}_2)   \ldots p_{\boldsymbol{\theta}} (\mathbf{z}_{L-1} | \mathbf{z}_L) p (\mathbf{z}_L)  
    \ d\mathbf{z}_{1:L} .
\end{align*}
\vspace{-0.4cm}

Note that every latent distribution $p_{\boldsymbol{\theta}} (\mathbf{z}_{l-1})$ has been implicitly marginalized by the subsequent latent variable $\mathbf{z}_{l}$, whose distribution has the same dependency to the next subsequent latent variable $\mathbf{z}_{l+1}$. Each latent layer can be understood as granting more flexibility to our model. 

In practice, $p_{\boldsymbol{\theta}}(\mathbf{x})$ is intractable as the marginalization over the latent variables is prohibitively expensive. Let $\mathbf{z}_{0}:= \mathbf{x}$, by applying the Bayes' rule, the intractability is overcome by approximating the true posteriors $ q_{\boldsymbol{\phi}} (\mathbf{z}_{l} | \mathbf{z}_{l-1}) \approx p_{\boldsymbol{\theta}} (\mathbf{z}_{l} | \mathbf{z}_{l-1}) $ for $l \in \{1, \ldots, L\}$, which are chosen from a variational family.

Then, the final optimization goal can be derived through Jensen’s inequality \cite{VAE}, the so-called hierarchical ELBO: $\mathcal{L}_{\boldsymbol{\theta},\boldsymbol{\phi}}(\mathbf{x}, \mathbf{z}_{1:L} )$. The hierarchical ELBO can be rearranged to obtain the typical variational trade-off that balances the reconstruction error and the prior regularizer, which enforces the posterior $q_{\boldsymbol{\phi}}(\mathbf{z}_{L} | \mathbf{z}_{L-1})$ to match the prior $p(\mathbf{z}_{L})$ as:
\begin{equation}
\begin{aligned} \label{eq:hierarchical-elbo}
  &\log p_{\boldsymbol{\theta}} (\mathbf{x}) \geq - \mathcal{L}_{\boldsymbol{\theta},\boldsymbol{\phi}}(\mathbf{x}, \mathbf{z}_{1:L} ) = 
 \\ 
 & \overbrace{ \mathbb{E}_{q_{\boldsymbol{\phi}} (\mathbf{z}_{1:L} | \mathbf{x})} \left[ \log p_{\boldsymbol{\theta}} (\mathbf{x}, \mathbf{z}_{1:L-1} | \mathbf{z}_{L}) \right] }^{ \textrm{Reconstruction Quality}} - \overbrace{ \mathcal{D}_{\operatorname{KL}}\left(q_{\boldsymbol{\phi}} (\mathbf{z}_{1:L} | \mathbf{x})\ |\ p(\mathbf{z}_L)\right) }^{ \textrm{Regularization}},
\end{aligned}
\end{equation}
where $\mathcal{D}_{\operatorname{KL}} := \mathbb{E}_{q_{\boldsymbol{\phi}}(\mathbf{z} | \mathbf{x})} \left[ \log p(\mathbf{z}_L) - \log q_{\boldsymbol{\phi}} (\mathbf{z}_{1:L} | \mathbf{x})  \right] $ is the Kullback-Leibler divergence that measures how two probability distributions differ from each other with $q_{\boldsymbol{\phi}} (\mathbf{z}_{1:L} | \mathbf{x}) = \prod_{l=1}^{L} q_{\boldsymbol{\phi}} (\mathbf{z}_{l} | \mathbf{z}_{l-1})$ and $p_{\boldsymbol{\theta}} (\mathbf{x}, \mathbf{z}_{1:L-1} | \mathbf{z}_{L}) = \prod_{l=0}^{L-1} p_{\boldsymbol{\theta}} (\mathbf{z}_{l} | \mathbf{z}_{l+1})$.

The deepest latent prior $p(\mathbf{z}_L)$ is defined as a standard logistic distribution statistically independent along its components. The inference $q_{\boldsymbol{\phi}} (\mathbf{z}_{l} | \mathbf{z}_{l-1})$ and likelihood $p_{\boldsymbol{\theta}} (\mathbf{z}_{l} | \mathbf{z}_{l+1})$ are defined as conditional independent logistic distributions, except for $p_{\boldsymbol{\theta}} (\mathbf{x} | \mathbf{z}_{1})$ which is categorical, so its discretization captures the main probability mass of the image pixels~\cite{pixelcnn++}.

\vspace{-0.4cm}
\subsection{Model Architecture}
We model the logistic distributions by a variational autoencoder of $L$ nested latent variables. The distinct latent variables are sampled following a Markov process of the form $\mathbf{z}_L \rightarrow \mathbf{z}_{L-1} \rightarrow \ldots \rightarrow  \mathbf{z}_2 \rightarrow \mathbf{z}_1 \rightarrow  \mathbf{x} $.

For the backbone of the model, we use residual blocks~\cite{resnet} which consist of the following sequential operations: a non-linear activation function, a convolutional layer, a non-linear activation function and a convolutional whose output is summed to the initial input. In our case, Exponential Linear Unit (ELU) are chosen for the activation functions.  The amount of residual blocks is modified to maintain constant the amount of networks parameters when varying $L$. 

\begin{figure}[t!]
\centering
\resizebox{8.9cm}{!} {
\includegraphics[width=3.3in]{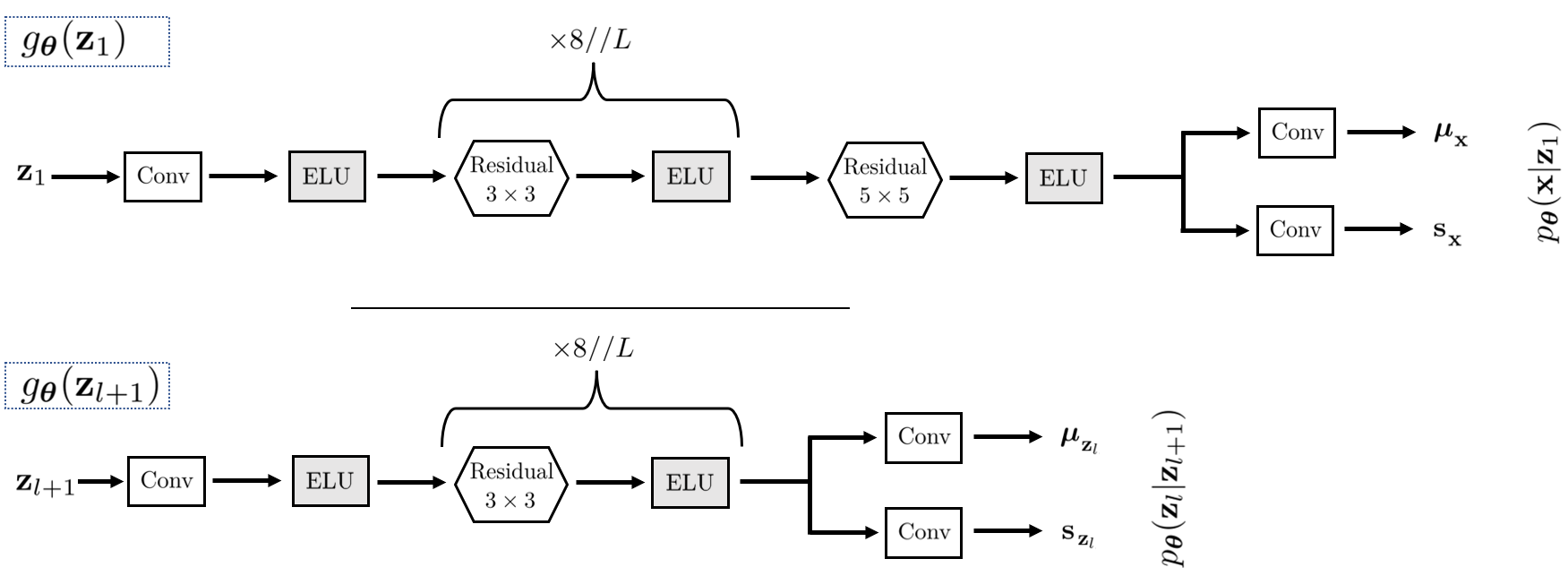}}
    \caption{\textbf{Network architecture corresponding to the decoder transforms}:  $p_{\boldsymbol{\theta}} (\mathbf{x} | \mathbf{z}_{1})$ and $p_{\boldsymbol{\theta}} (\mathbf{z}_{l} | \mathbf{z}_{l+1})$ for $l \in \{1, \ldots, L-1\}$. The arrows indicate the direction of the forward pass.}
    \label{fig:inference} \vspace{-0.5cm}
\end{figure}

We model the logistic distributions by a 2-dimensional convolutional neural network for every pair of logistic distribution parameters $\{(\boldsymbol{\mu}_l, \boldsymbol{\sigma}_l)\}_{l \in \{0,\ldots,L\}} $ for $h_{\boldsymbol{\phi}}$ and $g_{\boldsymbol{\theta}}$. The particular decoder transforms $g_{\boldsymbol{\theta}}$ are depicted in Fig.~\ref{fig:inference}. Furthermore, the encoder transform $h_{\boldsymbol{\phi}}(\mathbf{z}_l)$ has the same network architecture as its corresponding decoder  transform $g_{\boldsymbol{\theta}}(\mathbf{z}_{l-1})$ for $l \in \{1, \ldots, L\}$, obtaining a symmetric autoencoder.

All utilized convolutions are 2-dimensional with a 3$\times$3 kernel, except for the top latent encoder and decoder transforms where the kernel is 5$\times$5. The padding hyperparameter of each convolution is chosen in such a way that the input and the output dimensions match. All convolutions operate with 256 filters, except for the ones that determine the logistic parameters, which operate with 8 filters.

\vspace{-0.4cm}
\subsection{Compression Scheme} \label{sec:bitswap-scheme}
After training the model, we can successfully perform bits-back coding as our compression scheme. We derive an optimized version of bits-back that minimizes the required initial bits. Similarly as our lossy neural coder, we discretized the learned probability densities to apply an entropy coder.

\begin{figure}[!t]
\centering
\resizebox{8.9cm}{!} {
\includegraphics[width=3.3in]{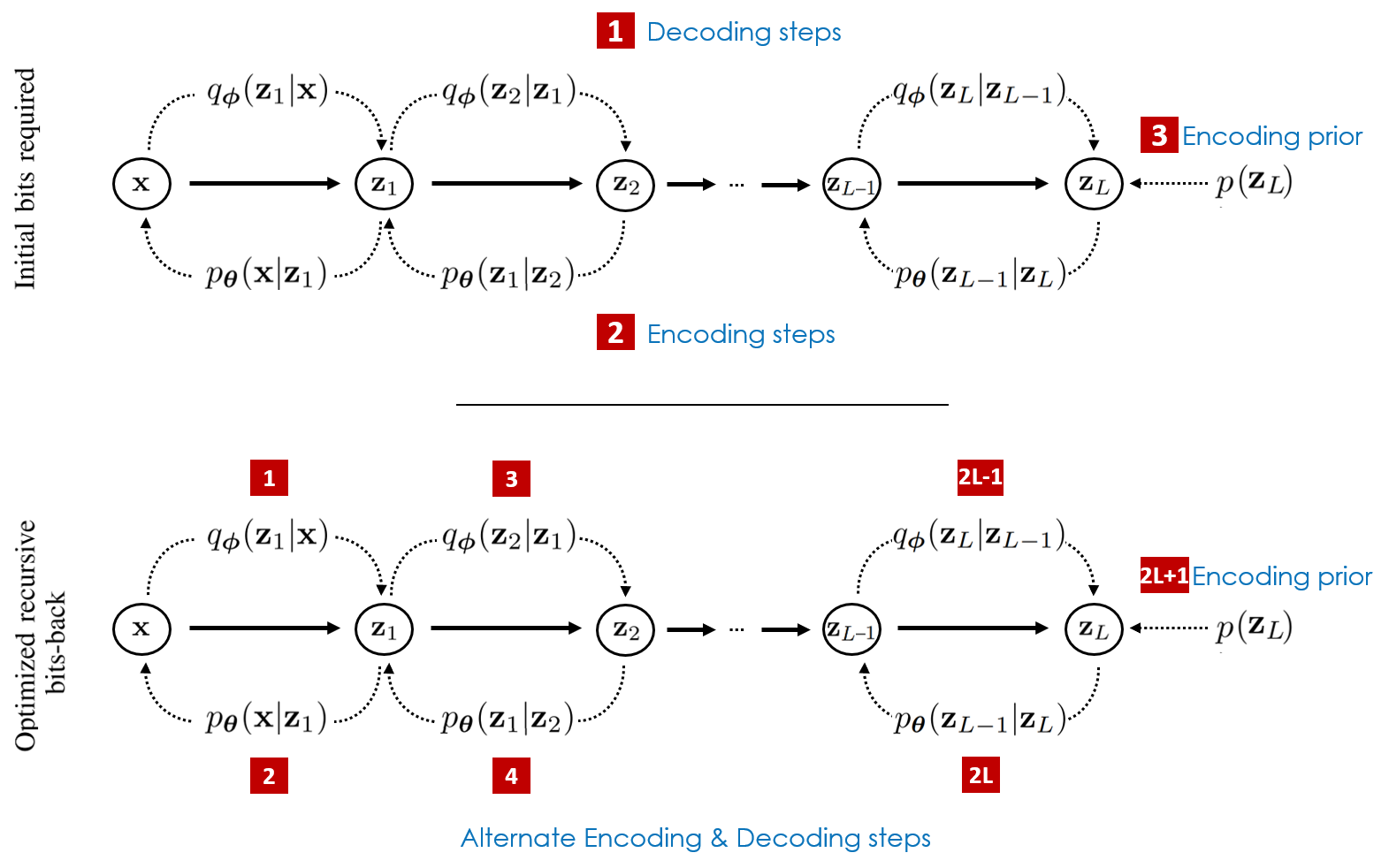}}
    \caption{\textbf{Order of the encoding and decoding operations} for a standard hierarchical bits-back and for its optimized scheme on a $L$-level nested variable model for lossless compression.}
    \label{fig:bitswap-order}  \vspace{-0.45cm}
\end{figure}

\subsubsection{Optimized Recursive Bits-back Scheme} Bits-back coding is a lossless compression method based on latent variable models~\cite{bits-back}. It sequentially codes the distinct datapoints and focuses on minimizing the net code length when coding the subsequent datapoint. In particular, the algorithm capitalizes the previously compressed datapoints to obtain the latent variables $\mathbf{z}_{l}$, building an efficient codec whose net code length coincides with the negative ELBO of the latent variable model.

Bits-back coding can be extended to nested latent variable models. In essence, it leverages the hierarchical dependency between the latent variables to recursively decode the subsequent latent variable and to have access to each probability distribution. Then, it encodes every latent variable using the corresponding likelihood or prior probability distribution. 

Assume we would like to transmit $\mathbf{x}$, we have access to $q_{\boldsymbol{\phi}}(\mathbf{z}_{l} | \mathbf{z}_{l-1}), p(\mathbf{z}_L),  p_{\boldsymbol{\theta}}(\mathbf{z}_{l-1} | \mathbf{z}_{l})$ for $l \in \{1, \ldots, L \}$ and to a stack-like entropy encoder. Suppose also there are $Bit_{\text{init}}$ bits from previous datapoints already compressed (or alternatively randomly initialized bits if $\mathbf{x}$ is the first datapoint to compress), then this method is made concrete by: 
\begin{enumerate}
    \item Subsequently decode the latent variables $\mathbf{z}_l$ for $l>0$ from the bitstream using $q_{\boldsymbol{\phi}}(\mathbf{z}_{l} | \mathbf{z}_{l-1})$, subtracting $- \sum_{l=1}^{L} \log q_{\boldsymbol{\phi}} (\mathbf{z}_{l} | \mathbf{z}_{l-1})$ bits to the code length.
    \item Subsequently encode in reverse order the latent variables $\mathbf{z}_l$ for $l<L$ to the bitstream using $p_{\boldsymbol{\theta}}(\mathbf{z}_{l-1} | \mathbf{z}_{l})$, adding $- \sum_{l=0}^{L-1} -\log p_{\boldsymbol{\theta}}(\mathbf{z}_{l-1} | \mathbf{z}_{l})$ bits to the code length.
    \item Encode $\mathbf{z}_L$ to the bitstream using $p(\mathbf{z}_L)$, adding $-\log p(\mathbf{z}_L)$ bits to the code length.
\end{enumerate}

Note that the expected increase in the code length corresponds to the negative hierarchical ELBO (see Eq.~\eqref{eq:hierarchical-elbo}). Thus it implies that optimizing a VAE with nested latent variables is equivalent to minimizing the net bitstream length when compressing with the bits-back scheme. 

This naive nested algorithm faces a challenge: the required initial bits $Bit_{\text{init}}^{\text{recursive}}$ grow with the number of latent variables, because each decoding step consumes additional initial bits. In particular, the depth $L$ of the nested latent variable model compromises the initial cost bitstream by $Bit_{\text{init}}^{\text{recursive}} = - \sum_{l=1}^{L} \log q_{\boldsymbol{\phi}} (\mathbf{z}_{l} | \mathbf{z}_{l-1})$. This inefficiency is mitigated by reordering encoding and decoding steps~\cite{bit-swap}, as shown in Fig.~\ref{fig:bitswap-order}. This optimized nested algorithm basically consists of encoding a variable immediately when possible, so the remaining decoding steps subtract bits from the encoded variables of the current datapoint and not from the initial bits.

In particular, the optimized scheme requires $- \log q_{\boldsymbol{\phi}}(\mathbf{z}_{1} | \mathbf{x})$ on the first decoding operation when sampling $\mathbf{z}_{1} \sim q_{\boldsymbol{\phi}}(\mathbf{z}_{1} | \mathbf{x})$. For the remaining decoding steps, at least $-\log p_{\boldsymbol{\theta}}(\mathbf{z}_{l-1} | \mathbf{z}_{l})$ bits are available to decode $\mathbf{z}_{l+1} \sim q_{\boldsymbol{\phi}}(\mathbf{z}_{l+1} | \mathbf{z}_{l})$, so the amount of initial bits required is bounded. Therefore, the optimized recursive bits-back scheme does not require more initial bits than its conventional version: $Bit_{\text{init}}^{\text{optimized}} \leq \chi < Bit_{\text{init}}^{\text{recursive}}=$ 

\noindent 
$- \log q_{\boldsymbol{\phi}}(\mathbf{z}_{1} | \mathbf{x}) + \sum_{l=1}^{L-1} \hspace{-0.15cm} -\log q_{\boldsymbol{\phi}}(\mathbf{z}_{l+1} | \mathbf{z}_{l})$; $\chi = \log q_{\boldsymbol{\phi}}(\mathbf{z}_{1} | \mathbf{x}) +$ 

\noindent 
$ \sum_{l=1}^{L-1} \max \big( 0, -\log q_{\boldsymbol{\phi}}(\mathbf{z}_{l+1} | \mathbf{z}_{l}) + \log p_{\boldsymbol{\theta}}(\mathbf{z}_{l-1} | \mathbf{z}_{l})\big)$.

Since arithmetic coding is inefficient with bits-back due to opposite decoding order~\cite{bits-back}, we use ANS (Section IV-C from Supplementary) to match its stack-like decoding.

\vspace{-0.25cm}
\section{ROI-eML: ROI Coding} \label{sec:method-roi}
This section explains the methodology followed to implement our proposed neural ROI coding: ROI-eML; entirely ML-driven as it relies solely on machine learning techniques. Our ROI coder is based on segmenting the input image in the blade and background regions to apply different compression coders in each region. ROI-eML gathers our segmentation model with our lossy and lossless coders to build an efficient ROI framework that does not include any significant overhead. Moreover, it removes the dependency of Bit-swap to generate initial random bits, enabling it to execute in parallel and, consequently, making Bit-swap time-efficient.

 \vspace{-0.4cm}
\subsection{Compression Scheme} \label{sec:method-roi-compression}
The ROI compression approach consists of first obtaining a segmentation mask $\mathbf{\hat s}$ of the blade (following Section~\ref{sec:method-segmentation}). Then, the input image is subdivided into image patches of equal dimensionality. To guarantee all image patches have the same dimensions, the image and mask are mirror padded. The image patches are classified as either blade or background. When a single pixel of an image patch belongs to the blade, the image patch is considered as a blade patch, because we want to assure no detail of a blade damaged is compromised. Hence, we end with a polygonal-shaped mask that can be compressed by identifying its corners $C = \{ (h,w) \,\mid\, \hat s_{h,w} \neq \hat s_{h+1,w} \text{ and } \hat s_{h,w} \neq \hat s_{h,w+1} \}$, which are the pixel coordinates with a non-negative difference along both axis dimensions. The polygon is encoded to identify which patches fall within the ROI when decompressing the patches. 

Then, we can encode the blade and background patches using different coders. In case the blade region is also coded lossily, two independent rate-distortion models from Section~\ref{sec:method-hp+easn} are used to compress the image patches in parallel. On the contrary, if the blade is compressed using our lossless coder from Section~\ref{sec:method-bitswap}, we first encode the background regions using our lossy coder, so the coded background patches can be used as the initial bitstreams for bits-back.

The initial bits for each blade patch coded losslessly are taken from a single coded background patch. In this way, we create distinct initial bitstreams, so the different blade patches can be coded in parallel. Thus, we remove the sequential dependency of bits-back coding, leading to a ROI algorithm that can computationally scale with the number of image patches in the lossy and lossless scenarios. In case there are more blade patches than background ones, they are encoded sequentially following the same procedure as Section~\ref{sec:bitswap-scheme}.

Our algorithm does not include any significant overhead information, because it only includes the image dimensions, the coded mask and which two models are employed. The ROI patch size is set to the maximum of the lossy and lossless coders to ensure compatibility, allowing the smaller-patch model to further subdivide the image.

\begin{figure*}[!t]
\centering
\resizebox{17.8cm}{!}{
\begin{tabular}{@{}cc@{}}
\includegraphics[width=3.3in]{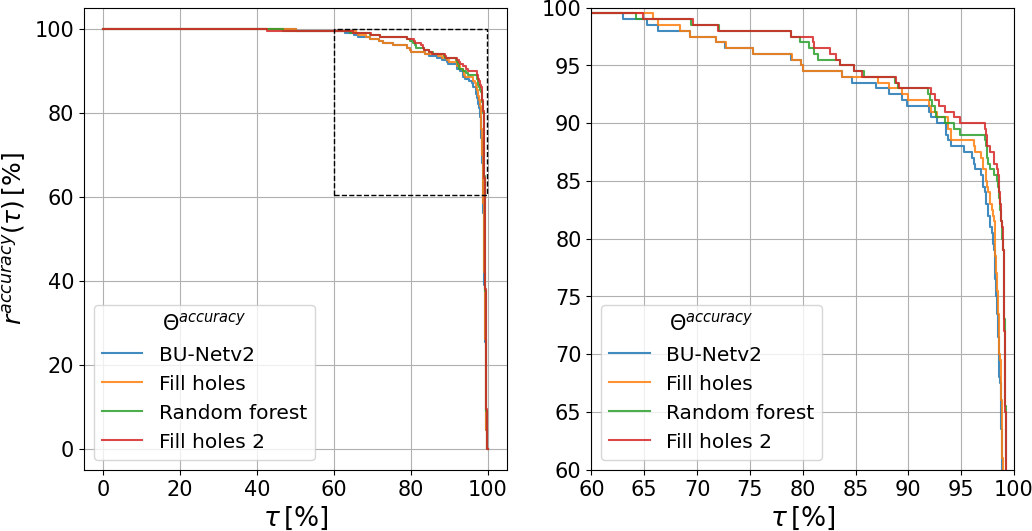}&
\hspace*{-0.15cm} \includegraphics[width=3.3in]{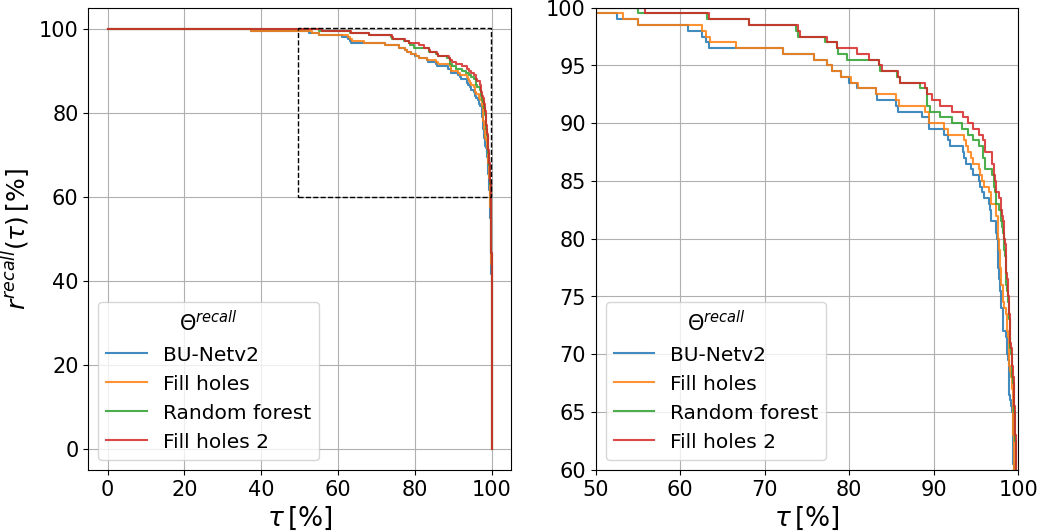}
\end{tabular}}
\vspace{-0.2cm}
    \caption{\textbf{Acceptance-ratio curves of each segmentation step.} $\Theta^{accuracy}$ and $\Theta^{recall}$ similarity metrics are displayed in the top and bottom sides, respectively. On the right column, the lefts plots are zoomed on the black dashed area.}
    \label{fig:seg-auc-post} \vspace{-0.35cm}
\end{figure*}

\begin{table}[!t]
\centering
\resizebox{9cm}{!}{
\begin{tabular}{lcccc}
\toprule
     \multicolumn{1}{c}{Performance} &  \multicolumn{1}{c}{BU-Netv2} &  \multicolumn{1}{c}{Fill holes} &  \multicolumn{1}{c}{Random forest} &  \multicolumn{1}{c}{Fill holes 2}\\
    \multicolumn{1}{c}{metric} &  \multicolumn{1}{c}{${\hat{\mathbf{s}}}^{BU}$} &   \multicolumn{1}{c}{$\hat{\mathbf{s}}^{H1}$} &  \multicolumn{1}{c}{$\hat{\mathbf{s}}^{RF}$} &  \multicolumn{1}{c}{$\hat{\mathbf{s}}\equiv\hat{\mathbf{s}}^{H2}$} \\ \midrule
   Accuracy [\%] & 96.82  & 96.98 &  97.54 & 97.61\\
   Recall [\%] &  96.41 & 96.61 & 97.54 & 97.67 \\
   F1-score [\%] & 95.60 & 95.72 & 96.44 & 96.50 \\
\bottomrule
\end{tabular}}
\vspace{-0.1cm}
\caption{\textbf{Performance results of each segmentation step}.} \label{tab:seg-accuracy} \vspace{-0.5cm}
\end{table}

\vspace{-0.15cm}
\section{Experimental Results on BU-Netv2+P} \label{sec:seg-results}

This section presents quantitative evaluations demonstrating the reliability, generality, robustness, and efficiency of our segmentation algorithm, with results shown for each step. We also introduce acceptance-ratio curves to assess robustness. Performance metrics are reported after rescaling segmentation masks to the original image resolution (see Supplementary Section II-A). Experiments use the dataset from \cite{PerezGonzaloIcip2023}, outlined in Supplementary Section I-A, with further qualitative and computational analysis in Supplementary Sections II D-F.

\vspace{-0.3cm}
\subsection{Acceptance-Ratio Curves}

Given any arbitrary similarity measure $\texttt{sim} \in [0,1]$ reported over a dataset $\mathcal{D}$, we aim to study the dispersion of $\texttt{sim}$. To compare the robustness of distinct approaches, we would like to identify instances with really low $\texttt{sim}$. A robust solution must perform reasonably well across all cases, rather than excelling on some while failing completely on others.

To this end, we define the acceptance-ratio $r^{\texttt{sim}}(\tau) =  \frac{ |\{ \mathbf{x} \in \mathcal{D} \,\mid\, \texttt{sim}(\mathbf{s}, \hat{\mathbf{s}}) > \tau \}| }{ |\mathcal{D}| }$ which measures the proportion of instances whose similarity is above a predefined threshold $\tau \in [0,1]$. This ratio could be understood as the proportion of instances that are considered (or accepted) as duly segmented. Thus, $\tau$ governs the extent of our criteria. Notice that it corresponds to the opposite of the cumulative density function of $\texttt{sim}$.

Since the threshold $\tau$ is chosen arbitrarily, when we wish to discuss the robustness of the performance, we must take into account all conceivable threshold values. Therefore, we define the acceptance-ratio curve $\Theta^{\texttt{sim}} = \{ (\tau, r^{\texttt{sim}}(\tau)) \,\mid\, \tau \in [0,1] \}$ as the parametric curve of $r^{\texttt{sim}}(\tau)$ given the distinct possible thresholds. The $\Theta^{\texttt{sim}}$ of distinct segmentation algorithms can be visualized to compare their robustness. The acceptance-ratio curves disaggregate each image into easily comparable curves, as their Area-Under-the-Curve AUC$(\Theta^{\texttt{sim}})$ represents the expectancy of the similarity measure:

\vspace{-0.5cm}
\begin{equation} \label{eq:auc-ar}
    \text{AUC}(\Theta^{\texttt{sim}}) = \int_{0}^{1} r^{\texttt{sim}}(\tau) \, d\tau = \mathbb{E}_{p(\tau)} \left[ \texttt{sim}(\mathbf{s}, \hat{\mathbf{s}}) \right]  .
\end{equation}
\vspace{-0.4cm}

When segmentation methods perform similarly (close expectancies), the most robust is the one with higher $r^{\texttt{sim}}(\tau)$ at lower thresholds $\tau$. Thus, acceptance-ratio curves offer a practical way to visualize robustness for any similarity metric.

\vspace{-0.4cm}
\subsection{Quantitative Results} \label{sec:seg-quant}
We evaluated the distinct algorithm steps over the test set of our approach in terms of accuracy, recall and F1-score. These performance metrics are expounded in Table~\ref{tab:seg-accuracy}.

First, we observe that BU-Netv2 already delivers high performing results, demonstrating how powerful are the new incorporated loss and the fine tuning process performed. In this way, the post-processing steps do not play such essential role as compared to BU-Netv1+P~\cite{PerezGonzaloIcip2023}. Particularly, despite improving the performance, the hole filling steps represent a small quantitative gain, but they are still valuable to accomplish excellent masks; see Section II-D of Supplementary.

The standard random forest represents a smaller improvement if the ensemble with the BU-Netv2 is not performed. For instance, without the ensemble strategy, the performance would indeed decrease, resulting in an accuracy of 97.28\%, a recall of 93.85\% and an F1-score of 95.11\%. 

Besides, Fig.~\ref{fig:seg-auc-post} effectively demonstrates that post-processing steps enhance the robustness of our network. The acceptance-ratio curves for the accuracy and recall depict that similarity metrics are always on top of the curves of the preceding algorithm step. Hence, after applying the subsequent post-processing step, we successfully improve the segmentation masks of most of the test images. Additionally, Fig. 3 of Supplementary proves our segmentation method's generalizability, with high performance across all windfarm images.

\vspace{-0.3cm}
\subsection{Quantitative Comparative Results} 

\begin{figure}[t!]
\centering
 \resizebox{8.9cm}{!}{
\begin{tabular}{c}
\includegraphics[width=4.in]{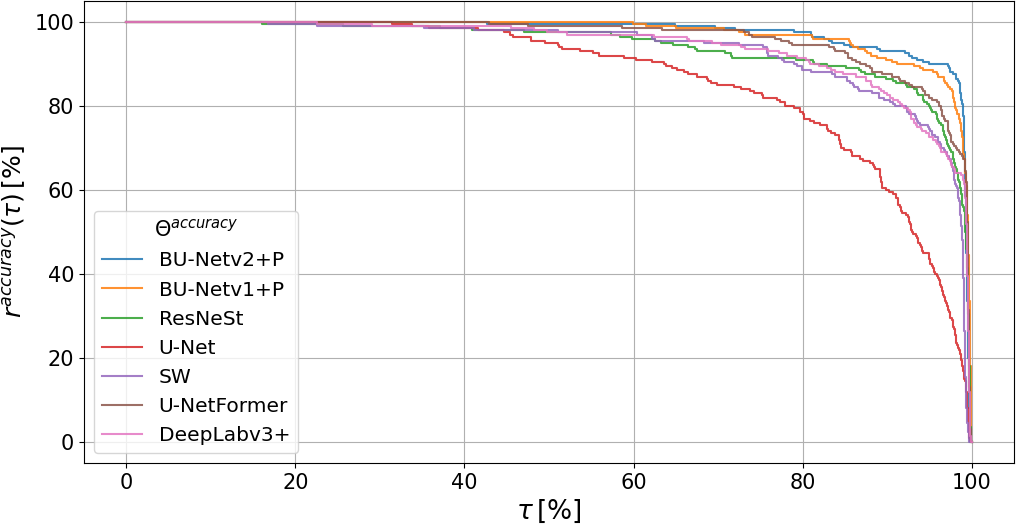} \\
\includegraphics[width=4.in]{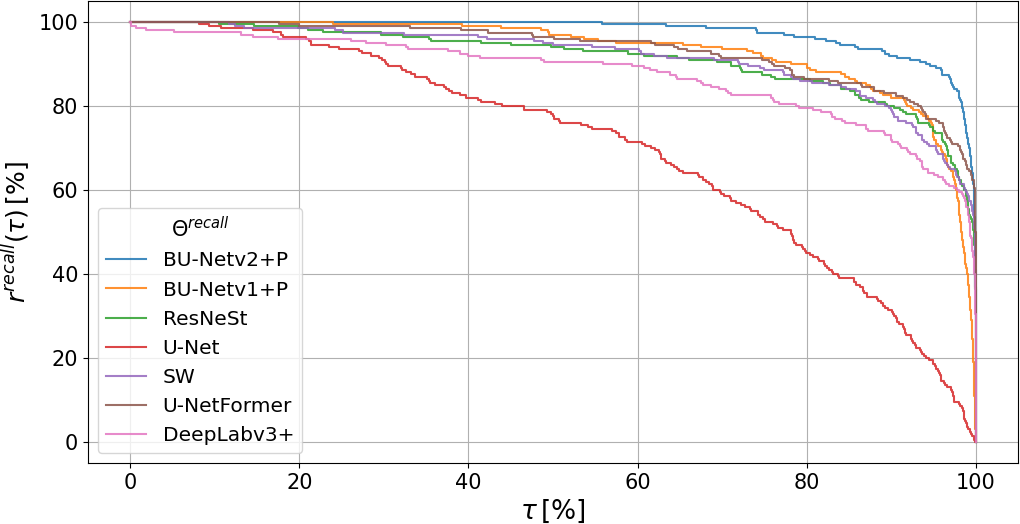}
\end{tabular}}
\vspace{-0.2cm}
    \caption{\textbf{Acceptance-ratio curves of comparison techniques.} $\Theta^{accuracy}$ and $\Theta^{recall}$ similarity metrics are displayed in the left and right parts, respectively.}
    \label{fig:auc} \vspace{-0.45cm}
\end{figure}

For comparison, we benchmark our proposed algorithm with the original U-Net~\cite{unet} along with other state-of-the-art encode-decoder networks: DeepLabv3+~\cite{deeplabv3+}, SW~\cite{sw} and ResNeSt~\cite{resnest}. Moreover, we also report the transformer U-NetFormer~\cite{unetformer}, inspired in the architecture that we have employed, and the competing algorithm BU-Netv1+P~\cite{PerezGonzaloIcip2023}. The average quantitative results are gathered in Table~\ref{tab:seg-unet-compare} and their acceptance-ratio curves in Fig.~\ref{fig:auc}. 

To make a fair comparison, we optimized those models following the same strategy of the original works. The main parameters that were tuned are the resolution of the input image, the loss, the backbone architecture, whether to employ atrous convolutions~\cite{deeplab}, the convolution output stride, and the data augmentation strategy. For instance, ResNet-50 is used as the backbone architecture of DeepLabv3+~\cite{deeplabv3+}, SW~\cite{sw} and ResNeSt~\cite{resnest}, while U-NetFormer~\cite{unetformer} employs ResNet-18.

\begin{table}[t!]
\centering
\resizebox{9 cm}{!} {
\begin{tabular}{lcccccc}
\toprule
    \multicolumn{1}{c}{Method} & \multicolumn{1}{c}{Accuracy}  & \multicolumn{1}{c}{Recall} & \multicolumn{1}{c}{F1-score} & mIoU & Relative & Relative \\
    \multicolumn{1}{c}{} & [\%] & [\%] & [\%] & [\%]  & accuracy & recall \\
    \midrule
    
   U-Net~\cite{unet} & 86.24 & 68.93 & 77.95 & 75.94 & 1 & 1 \\
   DeepLabv3+~\cite{deeplabv3+} & 94.14 & 87.38 & 89.03 & 87.47 & 1.09 & 1.27 \\ 
   SW~\cite{sw} & 93.48 & 91.71 & 91.37 & 87.44 & 1.08 & 1.33 \\ 
   ResNeSt~\cite{resnest} & 94.23 & 91.47 & 92.77 & 89.63 & 1.09 & 1.33 \\ 
   U-NetFormer~\cite{unetformer} & 96.20 & 93.51 & 94.42 & 91.75 & 1.12 & 1.36 \\
   BiRefNet~\cite{birefnet} & 95.65 & 92.52 & 94.57 & 92.38 & 1.11 & 1.35 \\ 
   SAM*~\cite{sam} & 94.36 & 91.22 & 92.60 & 91.66 & 1.10 & 1.34 \\ 
   DiffSeg*~\cite{diffseg} & 96.37 & 85.73 & 86.40 & 72.14 & 1.05 & 1.12 \\ 
   EfficientFormer~\cite{efficientformer} & 96.42 & 93.63 & 94.55 & 93.81 & 1.12 & 1.37 \\ 
   MobileViT~\cite{mobilevit} & 96.14 & 93.33 & 94.38 & 93.47 & 1.11 & 1.36 \\
   Mask2Former~\cite{mask2former} & 96.68 & 93.89 & 94.76 & 93.72 & 1.12 & 1.37 \\
   BU-Netv1+P~\cite{PerezGonzaloIcip2023} & 97.39 & 93.35 & 95.73 & 93.80 & \textbf{1.13} & 1.35 \\ 
   BU-Netv2+P ($\hat{\mathbf{s}}$) & {\bf 97.61} & {\bf 97.67} & {\bf 96.50} & 94.58 & \textbf{1.13} & \textbf{1.42} \\ 
\bottomrule
\end{tabular}
}
\vspace{-0.2cm}
\caption{\textbf{Quantitative comparison on blade segmentation} with respect to state-of-the-art methods. Models marked with an asterisk only propose segmentation regions without a class; results assume a perfect region classifier.} \label{tab:seg-unet-compare}  \vspace{-0.4cm} 
\end{table}

While the U-Net~\cite{unet} achieves a notable accuracy thanks to properly identifying the background, it struggles in identifying properly the blade region, obtaining a poor recall of 68.93\%. On the other hand, our method overcomes the blade-background imbalances, and it escapes from the tendency of inferring the most prevalent class (the background). Hence, we gain in accuracy and prominently in recall. The BU-Netv2 achieves 97.67\% of recall without dropping its precision, outperforming substantially the other methods and its previous version~\cite{PerezGonzaloIcip2023}. Note that our method represents a huge relative improvement of 42\% compared to the baseline U-Net recall.

Fig.~\ref{fig:auc}-bottom shows an in-depth analysis of the recall similarity along the test data. We observe the acceptance-ratio curve of our method for the recall is way above the other curves, which are conglomerated together. This proves the robustness of the solution compared to competing techniques. For instance, we can notice our algorithm accomplishes a recall above 60\% in all the test images, while the rest of the methods can only assure a recall above 40\%. All in all, we demonstrate that the U-Net~\cite{unet} properly tailored with a post-processing can yield top performance results.

\vspace{-0.35cm}
\subsection{Computational Cost} \label{sec:seg-time}

The inference time for segmentating a blade image is tested for our segmentation algorithm. The experiments are run on a virtual machine with a NVIDIA RTX 3080 Ti GPU and a 20-core processor Intel core i9-10900 KF at 3.70 GHz. The provided results assume the model weights are already loaded, emulating a drone inspection. Note that the algorithm's computational time has not been optimized. This hardware setup is used for benchmarking purposes only. In practice, compression may be performed offboard or post-inspection.

The overall running time for segmenting an image corresponds to 0.089s. The step that consumes most time is the random forest, primarily for its training, as it gathers around 30 input images. The random forest training represents 0.039s and, along with the input image assembly and ensemble prediction, entails two-thirds of the overall running time (in particular, 0.058s). The second step that spends more running time is the BU-Netv2, as it predicts the images one by one, simulating a drone inspection. Note that, after all, the BU-Netv2 performs the predictions with the same running time as the U-Net~\cite{unet}, since they are based on the same architecture. The hole filling steps do not signify any significant time. Ultimately, there is some marginal overhead time, such as reading the input image and, masking and storing the blade image. The average and cumulative runtimes for each step in our algorithm are illustrated in Section II-F of Supplementary.

 \vspace{-0.25cm}
\section{Experimental Results on HP+EASN-deep} \label{sec:lossy-results}

\begin{figure*}[!t]
\centering
\resizebox{1.0\linewidth}{!}{
\begin{tabular}{c}
\hspace{-0.3cm}\includegraphics[width=\linewidth]{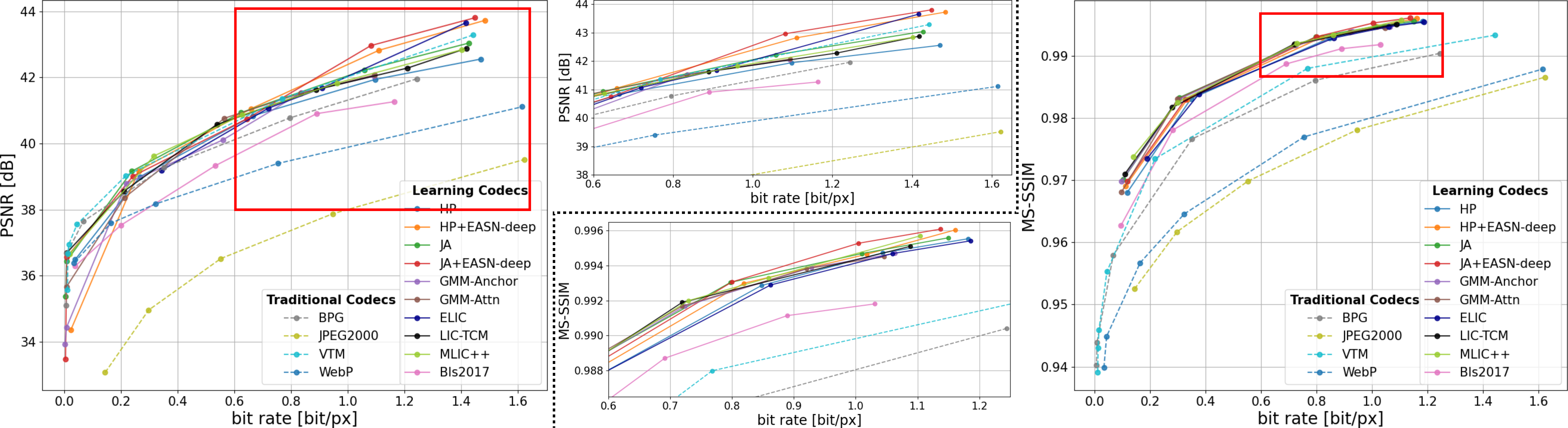}
\end{tabular}}
\vspace{-0.35cm}
    \caption{\textbf{Rate-distortion curves for state-of-the-art lossy coders}. Learning-based curves for each distortion loss metric: PSNR (left) or MS-SSIM (right) with red regions zoomed (middle). Non-learning-based coders are depicted with dashed lines.}
    \label{fig:lossy-performance} \vspace{-0.45cm}
\end{figure*} 

In this section, we analyze the distinct state-of-the-art lossy coders in order to determine that HP+EASN-deep is the best option for our industry application problem. Therefore, we are seeking a method that accomplishes higher compression rates with a reasonable loss in the image quality. The distortion introduced should not exceed a certain limit, as those images would be assessed by blade specialists to outline the repair plan. Lastly, because on-site we do not possess high computer capacity, another important requirement is that compressing a full-resolution image is performed in acceptable time. The compression dataset is formed of 64,438 high-resolution images, where the training set is 80\% and validation and test sets are 10\% each (details in Section I-B of Supplementary).

\vspace{-0.3cm}
\subsection{Compression Performance}
We evaluate learning-based methods against the current industry standard JPEG2000~\cite{J2K} and other traditional coders, whose coding settings are detailed in Section III-E of Supplementary. Such comparison is performed using the rate-distortion curves, shown in Fig.~\ref{fig:lossy-performance}. These curves describe the average image quality for a given bit rate. Quality is quantified using two measures: PSNR and MS-SSIM. 

Since models are trained on 256$\times$256 image patches, full-resolution images are mirror padded before patch extraction. Alternative padding methods, like zero padding, were tested but they introduce artifacts not present in the training.

In PSNR, Fig.~\ref{fig:lossy-performance}-left shows that learning-based methods generally outperform traditional coders, except at very low bit rates where VTM leads. However, this image quality is unacceptable on these low bit rates to perform blade assessments. As bit rate increases, learning-based algorithms improves significantly, with HP+EASN-deep and JA+EASN-deep performing best. Since their relative performance varies by bit rate, we further evaluate their practical use by analyzing computation time in Section\ref{sec:lossy-time}.

In terms of MS-SSIM (Fig.~\ref{fig:lossy-performance}-right), all learning-based coders significantly outperform traditional ones, with similar performance across most methods except Bls2017\cite{bls2017}. Once again, it becomes crucial to take into account the coding time.

Table~\ref{tab:lossy-compression-time} includes the BD-rate for both quality metrics. HP+EASN-deep leads in PSNR, showcasing superior performance than autoregressive methods, which are typically sequential and slow. In MS-SSIM, we observe that learning approaches depict similar BD-rate, now with autoregressive methods leading, as they are more consistent in lower bit rates.

Section III-D of Supplementary material provides a detailed study on visual artifacts introduced by HP+EASN-deep and how they could affect the detection of blade defects.

\vspace{-0.15cm}
\subsection{Computational Cost} \label{sec:lossy-time}

We compare the inference time of various state-of-the-art coders under the same infrastructure as in Section~\ref{sec:seg-time}, on our blade compression test set. Table~\ref{tab:lossy-compression-time} reports the minimum, maximum, and average coding times in seconds. Since higher bit rates increase runtime due to entropy coding, we report the slope of this growth. Alternatively, Fig.~9 in the Supplementary visually represents the running times reported in Table~\ref{tab:lossy-compression-time}. Ideally, a top-performing coder for on-site scenarios with limited computing resources should offer fast compression and decompression with minimal sensitivity to bit rate.

HP+EASN-deep is the coding scheme that better fulfills these requirements. As mentioned, HP+EASN-deep along with JA+EASN-deep displays a remarkable compression performance. Its average compression time is closed to the other leading coders, with the exception of Bls2017~\cite{bls2017} that exhibits an astonishingly rapid timing. Moreover, it displays a low slope growth in compression and the average decompression time is not substantially high. In contrast, we have that JA+EASN-deep is several times slower, because of its autoregressive (sequential) contextual prediction process.

\begin{table*}[t!]
\centering
\resizebox{17.8 cm}{!} {
\begin{tabular}{lcccccccccccc}
\toprule
     & \multicolumn{2}{c}{\textbf{Performance}} & \multicolumn{4}{c}{\textbf{Compression Time}} & \multicolumn{4}{c}{\textbf{Decompression Time}} \\
    \multicolumn{1}{c}{Coding} & BD-rate & BD-rate &  \multicolumn{1}{c}{Minimum}  & \multicolumn{1}{c}{Maximum} & \multicolumn{1}{c}{Mean} & Slope & \multicolumn{1}{c}{Minimum}  & \multicolumn{1}{c}{Maximum} & \multicolumn{1}{c}{Mean} & Slope \\
    \multicolumn{1}{c}{scheme} & [PSNR] & [MS-SSIM] & time [s] & time [s] & time [s]  & growth [s/bit] & time [s] & time [s] & time [s]  & growth [s/bit] \\
    \midrule  
   HP \cite{hyperprior} & 15.35 &  5.81  & 5.76 &    6.17 &    5.96 &    0.25 &   6.51 &  15.60 &  12.83 &   5.57 \\
 HP+EASN-deep \cite{easn} & \textbf{0} & 0 & 6.46 &    6.95 &    6.64 &    0.26 &   7.86 &  15.35 &  12.74 &   4.00 \\
           JA \cite{ja} & 2.55 & \textbf{-6.04}  & 200.68 &  204.19 &  203.01 &    1.96 & 442.74 & 458.26 & 451.08 &   8.69 \\
 JA+EASN-deep \cite{easn} & 2.14 & -1.11  & 200.53 &  205.75 &  203.42 &    2.83 & 443.54 & 458.08 & 453.10 &   7.88 \\
   GMM-Anchor \cite{gmm} & 18.14 & -0.40  & 227.91 &  235.96 &  233.04 &    5.51 & 480.62 & 495.27 & 487.40 &  10.02 \\
     GMM-Attn \cite{gmm} & \underline{2.09} & -1.49  & 247.97 &  258.12 &  252.81 &    6.78 & 500.60 & 523.14 & 508.58 &  15.06 \\
    ELIC \cite{elic} &  12.00 &  7.44     & 112.78 & 128.72 & 123.37 & 20.70 & 35.38 & 36.58 & 36.06 & 1.56 \\
    LIC-TCM \cite{lic-tcm}  & 9.04 & \underline{-2.54}  & 39.06 & 41.38 & 40.55 & 6.28 & 45.35 & 54.72 & 52.02 & 25.40 \\
    MLIC++ \cite{mlic++} & 7.54 & -2.42   & 50.86 & 54.33 & 53.24 & 2.34 & 74.38 & 79.26 & 77.41 & 3.28\\
      Bls2017 \cite{bls2017} & 431.75 & 30.89  & 0.27 &    0.35 &    0.30 &    0.07 &   2.85 &  11.94 &   9.12 &   7.36 \\
          BPG \cite{bpg} &  34.47  &   93.02  & 2.47 &    8.80 &    4.75 &    3.25 &   1.08 &  12.78 &   5.91 &   6.01\\
     JPEG2000 \cite{J2K} &  58.05 &    286.16  & 7.02 &    7.03 &    7.02 &    0.01 &  24.40 &  58.48 &  38.92 &  16.66\\
          VTM \cite{vtm} &  4.06 &    56.96  & 102.57 & 9943.02 & 4584.58 & 5309.94  &   1.42 &   4.32 &   2.73 &   1.57\\
         WebP \cite{webp} & 147.89  &    244.32  & 2.49 &    9.40 &    4.17 &    3.59 &   1.26 &  10.87 &   5.56 &   4.99 \\
\bottomrule
\end{tabular}
}
\vspace{-0.15cm} 
\caption{\textbf{BD-rate, compression and decompression times of a full-resolution image} for state-of-the-art lossy coders.} \label{tab:lossy-compression-time} \vspace{-0.45cm}
\end{table*}

\vspace{-0.2cm}
\section{Experimental Results on Bit-swap} \label{sec:lossless-results}

Bit-swap is evaluated against other competing lossless coding methods. While it achieves strong compression, its sequential patch-wise coding results in longer runtimes, limiting its industrial use. Section IV-B of Supplementary compares standard and optimal recursive bits-back, and IV-G analyzes the optimal number of hierarchical latents. We employ the same dataset as for our lossy experiments (Section~\ref{sec:lossy-results}).

\vspace{-0.35cm}
\subsection{Compression Performance}
Table~\ref{tab:lossless-compression-performance} compares the bit rates of various lossless coding schemes. Traditional coding settings are detailed in Supplementary Section IV-H. Bit-swap achieves a competitive performance with a bit rate of 8.98 bit/pix, outperforming traditional standard methods, as well as more recent learned approaches like LC-FDNet\cite{lc-fdnet} and LLICTI~\cite{llicti}. Although DLPR~\cite{dlpr} marginally improves upon Bit-swap (8.93 bit/pix), Bit-swap remains highly effective. Detailed ablations showcase top-performing results with $L=2$ and a patch size of $64\times64$ as shown in Supplementary Section IV-G.

\begin{table}[t!]
\centering
\resizebox{7.25 cm}{!} {
\begin{tabular}{lcc}
\toprule
    \multicolumn{1}{c}{Coding} & \multicolumn{1}{c}{Mean}  & \multicolumn{1}{c}{Std}  \\
    \multicolumn{1}{c}{scheme} & bit rate [bit/pix] & bit rate [bit/pix]  \\
    \midrule
   Bit-swap & \underline{8.98} &  1.70   \\
   PNG \cite{png} & 11.33 & 1.64 \\
   BPG \cite{bpg} &  11.61  &  2.19 \\
   JPEG2000 \cite{J2K} &  10.41  & 1.69 \\
   WebP \cite{webp} &  10.17  &  1.77 \\
   JXL \cite{jxl} &  9.77  &  1.72 \\
   LC-FDNet \cite{lc-fdnet} & 9.16 & 1.59 \\
   LLICTI \cite{llicti} & 10.24 & 1.96 \\
   DLPR \cite{dlpr} & \textbf{8.93} & 1.71 \\

\bottomrule
\end{tabular}
}
\vspace{-0.15cm}
\caption{\textbf{Compression performance of a full-resolution image} for the distinct state-of-the-art lossless coders. Std denotes the standard deviation across test images.} \label{tab:lossless-compression-performance} \vspace{-0.35cm}
\end{table}

\vspace{-0.35cm}
\subsection{Computational Cost}

Our lossless algorithm has been tested under the same conditions as Section~\ref{sec:lossy-time}. When comparing the coding inference times for various state-of-the-art lossless coding schemes in Table~\ref{tab:lossless-compression-time}, we observe that the Bit-swap model shows longer compression and decompression times, being excessively slower compared to the rest. In particular, it exceeds a one-hour duration by a significant margin, which is not the case for the competing techniques. This slower processing speed arises due to coding each image patch sequentially, so the generated amount of random bits is minimized. This huge computational cost makes Bit-swap unfeasible and its parallel implementation becomes crucial for the industrial setting.

\vspace{-0.25cm}
\section{Experimental Results on ROI-eML} \label{sec:roi-results}

In this section, we present the observational results for our proposed ROI-eML framework, which gathers the exposed learning-based algorithms: BU-Netv2+P for image segmentation, HP+EASN-deep for lossy coding and Bit-swap for lossless coding. The empirical findings demonstrate that ROI-eML retains the compression performance of the presented coding algorithms, while provides a fast ROI framework that supports both compression modes.

\begin{table}[t!]
\centering
\resizebox{7.25 cm}{!} {
\begin{tabular}{lcccc}
\toprule
    & \multicolumn{2}{c}{\textbf{Compression}} & \multicolumn{2}{c}{\textbf{Decompression}} \\
    \multicolumn{1}{c}{Coding} & \multicolumn{1}{c}{Mean}  & \multicolumn{1}{c}{Std} & \multicolumn{1}{c}{Mean}  & \multicolumn{1}{c}{Std}  \\
    \multicolumn{1}{c}{scheme} & time [s] & time [s] & time [s] & time [s] \\
    \midrule
   Bit-swap &  5688  &  21.56  &  4367  & 19.64 \\
   PNG \cite{png} &  11.35  & 2.28  &  9.8  & 1.96 \\
   BPG \cite{bpg} &  7.22  &  0.49 &  14.03  & 1.99 \\
   JPEG2000 \cite{J2K} &   7.03 & 0.89  &  18.78  & 4.91 \\
   WebP \cite{webp} &  62  &  13.55 &  9.4  &  2.33 \\
   JXL \cite{jxl} &  757.8  &  34.10 &  9.80  & 2.48\\
   LC-FDNet \cite{lc-fdnet} & 160.73 & 7.73 & - & - \\
   LLICTI \cite{llicti} & 32.48 & 5.00 & 36.52 & 4.77 \\ 			
   DLPR \cite{dlpr} & 297.82 & 3.59 & 350.57 & 2.54 \\
\bottomrule
\end{tabular}
}
\vspace{-0.15cm}
\caption{\textbf{Compression and decompression times of a full-resolution image} for state-of-the-art lossless coders. Std denotes the standard deviation across test images.} \label{tab:lossless-compression-time}
\vspace{-0.45cm}
\end{table}

\vspace{-0.45cm}
\subsection{Compression Performance} \label{sec:roi-performance}

ROI-eML introduces some extra bits to encode the segmentation mask (Section~\ref{sec:method-roi-compression}). This overhead information from the mask is negligible, since it can be transmitted by specifying in which patches the blade region is narrowed/widened. Therefore, the mask can be encoded using $\approx 100$ bytes, which for a full-resolution image constitutes only $\approx 10^{-5}$ bit/px. This process is done automatically using a run-length coder.

Since the compressed segmentation mask does not introduce any significant extra bits, the bit rate of a full-resolution image coded using ROI-eML depends solely on the proportion of foreground and background regions. In case that the blade region represents most part of the image, the resulting bit rate will be significantly higher, because the blade is always coded using a higher bit rate model. Similarly, the quality measures depend on the proportion of each region. However, while the bit rate simply corresponds to the weighted average based on the proportion of area of each region, PSNR and MS-SSIM cannot be directly estimated in such a way, because they are not linear functions. Particular examples with its visual representation and its compression performance per region are included in Section V-A of Supplementary material.

\begin{figure*}[t!] 
    \centering
    \resizebox{17.cm}{!}{
    \begin{tabular}{@{}c@{}}
 \includegraphics[width=.998\linewidth]{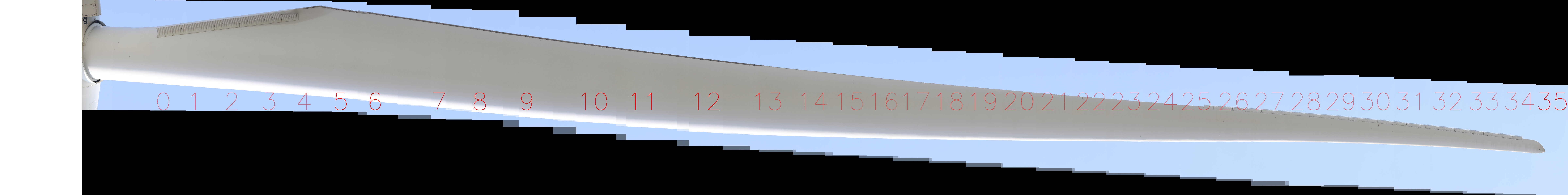}  \vspace{-0.1cm}\\
\includegraphics[width=1\linewidth]{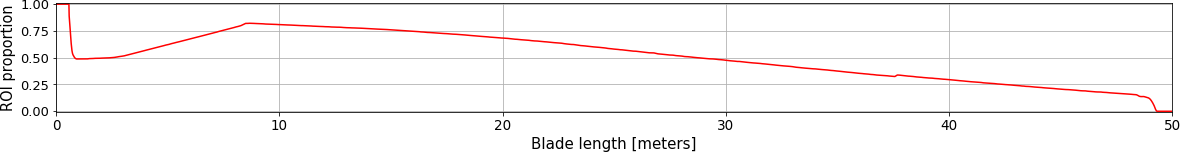}
\end{tabular}}
\vspace{-0.3cm}
    \caption{\textbf{Proportion of ROI width along the pressure side of a wind turbine blade}. Red numbers denote the $i$-th image.}
    \label{fig:mosaic} \vspace{-0.45cm}
\end{figure*}

It is important to note that the size of the blade region is considerably different depending on which fraction of the blade is captured. Figure~\ref{fig:mosaic} shows an example of a whole blade illustrating that the shoulder of the blade is wider than the tip and, therefore, the picture includes a higher portion of the blade region. In particular, the average relation between the background and the blade area is 2:1.

\subsection{Visualization of the ROI Compressed Images}  \label{app:roi-visual}

\begin{figure*}[t!]
\begin{center}
\resizebox{17.8cm}{!} {
\begin{tabular}{@{}c@{}c@{}c@{}c@{}}
\begin{subfigure}[!h]{0.95\textwidth}
\includegraphics{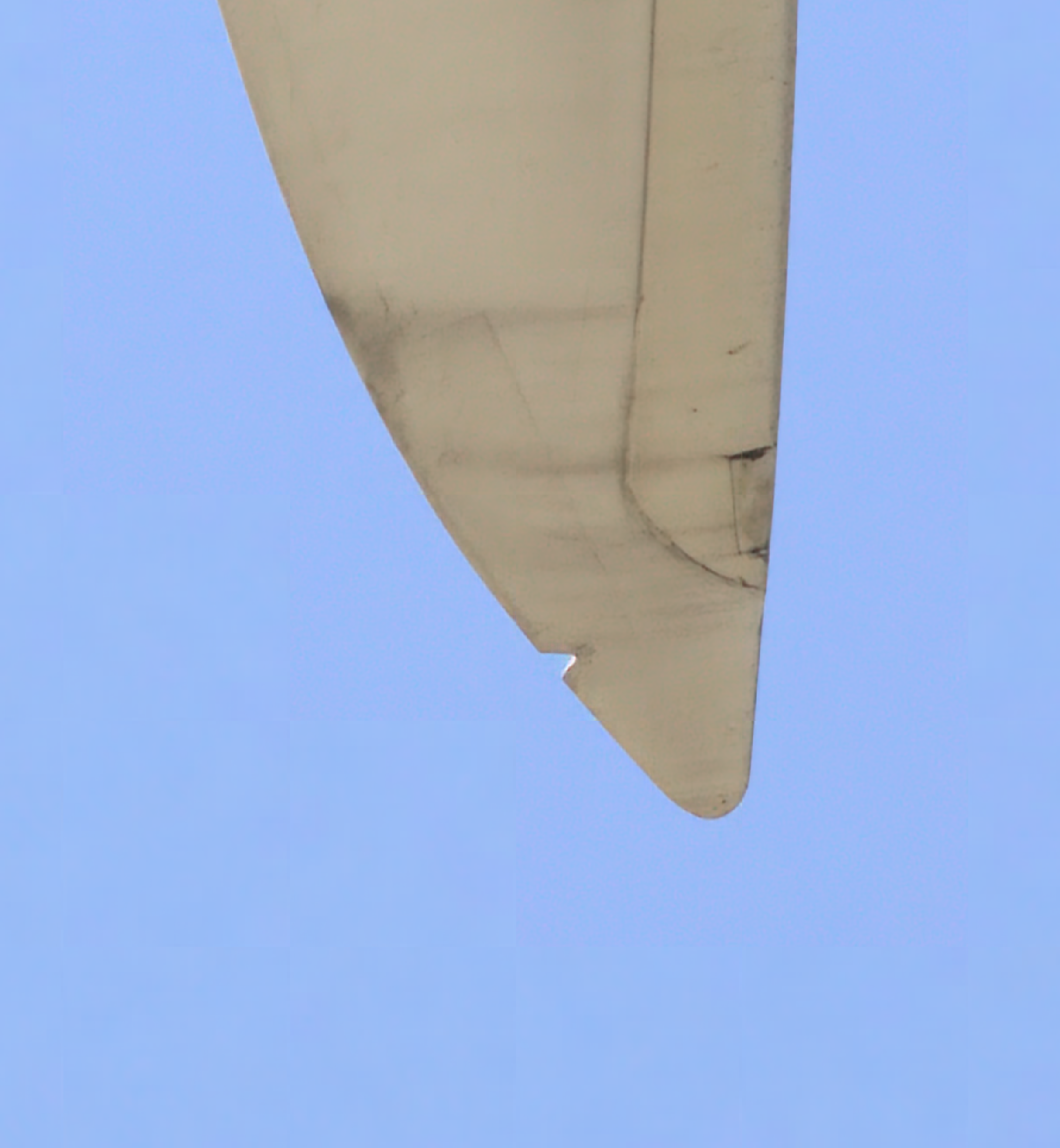}
\caption*{\centering \parbox{\linewidth}{\centering\Huge Blade: 38.22 / 0.9723 / 0.2933 \\ Background: 35.77 / 0.9371 / 0.0108 \\ General: 35.99 / 0.9412 / 0.0398 }}
\end{subfigure}
&
\begin{subfigure}[!h]{0.77\textwidth}
\includegraphics{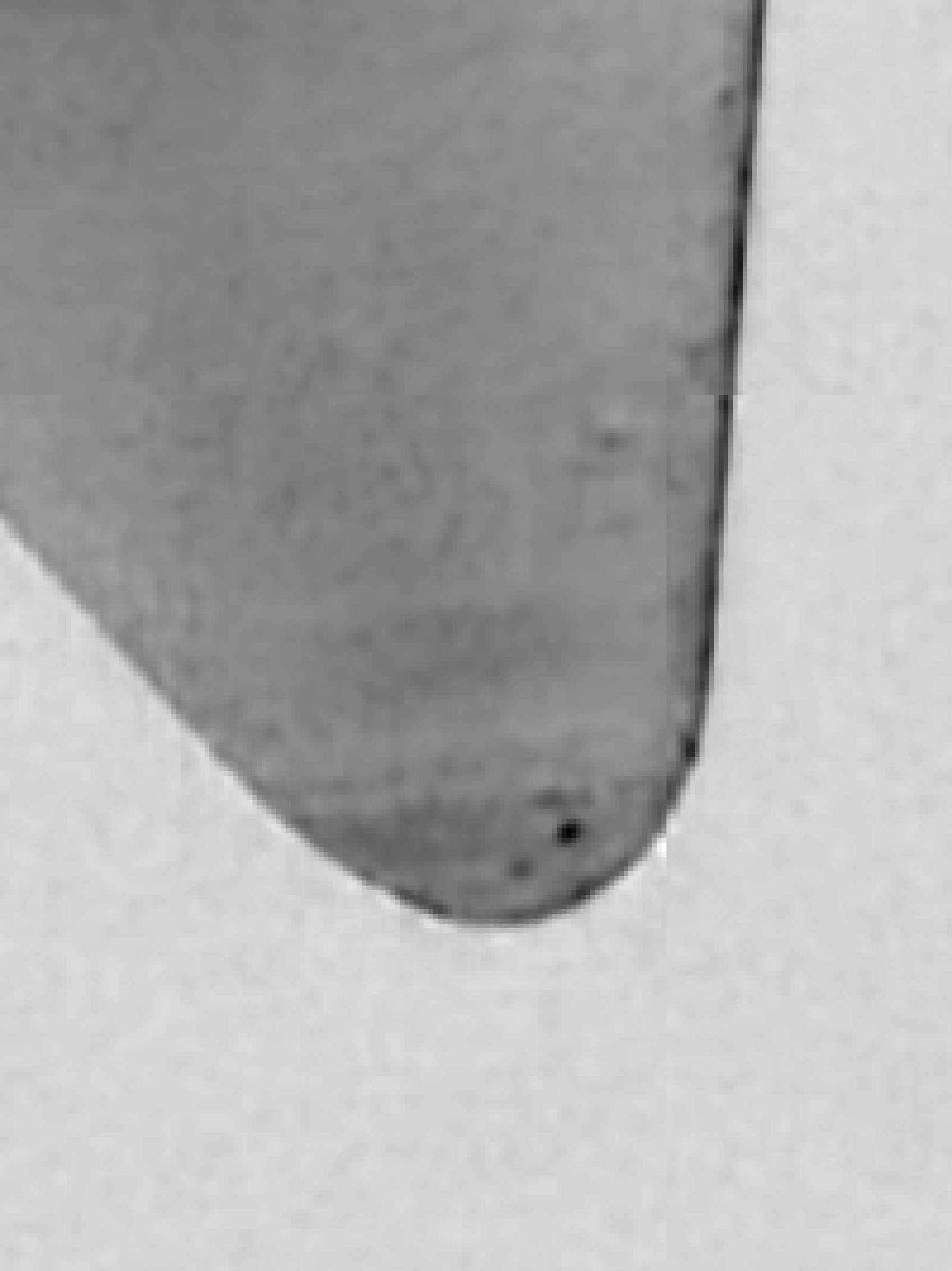}
\caption*{\centering \parbox{\linewidth}{\centering\Huge Zoomed gray-scaled image \\ Lossy blade region \\ \textcolor{white}{.}  }}
\end{subfigure}
&
\begin{subfigure}[!h]{0.95\textwidth}
\includegraphics{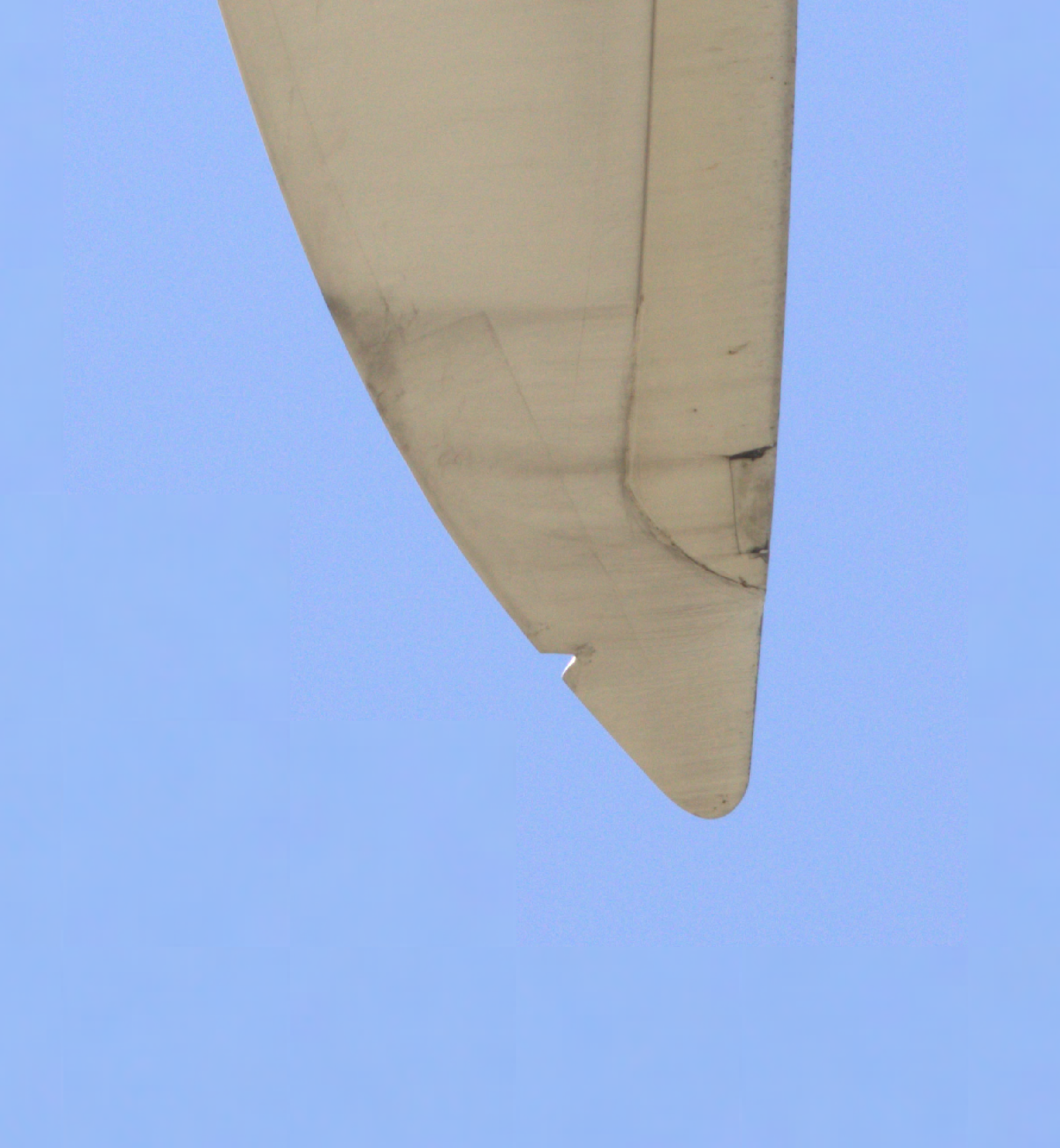}
\caption*{\centering \parbox{\linewidth}{\centering\Huge Blade: 9.2612 bit/px \\ Background: 35.77 / 0.9371 / 0.0108 \\ General: 36.30 / 0.9444 / 0.9625 }}
\end{subfigure}
&
\begin{subfigure}[!h]{0.77\textwidth}
\includegraphics{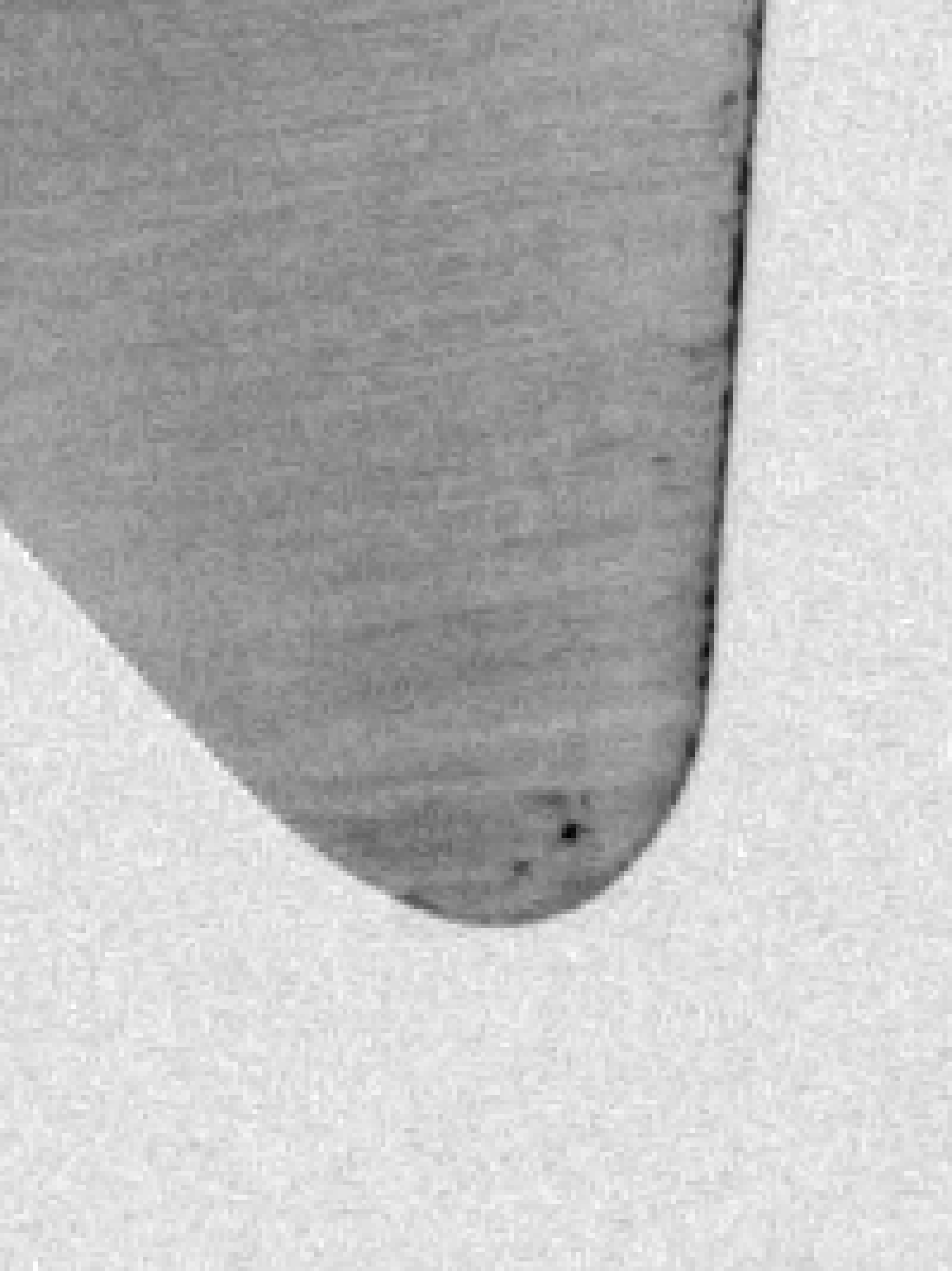}
\caption*{\centering \parbox{\linewidth}{\centering\Huge Zoomed gray-scaled image \\ Lossless blade region \\ \textcolor{white}{.} }}
\end{subfigure}
\\ & & & \\
\begin{subfigure}[!h]{0.95\textwidth}
\includegraphics{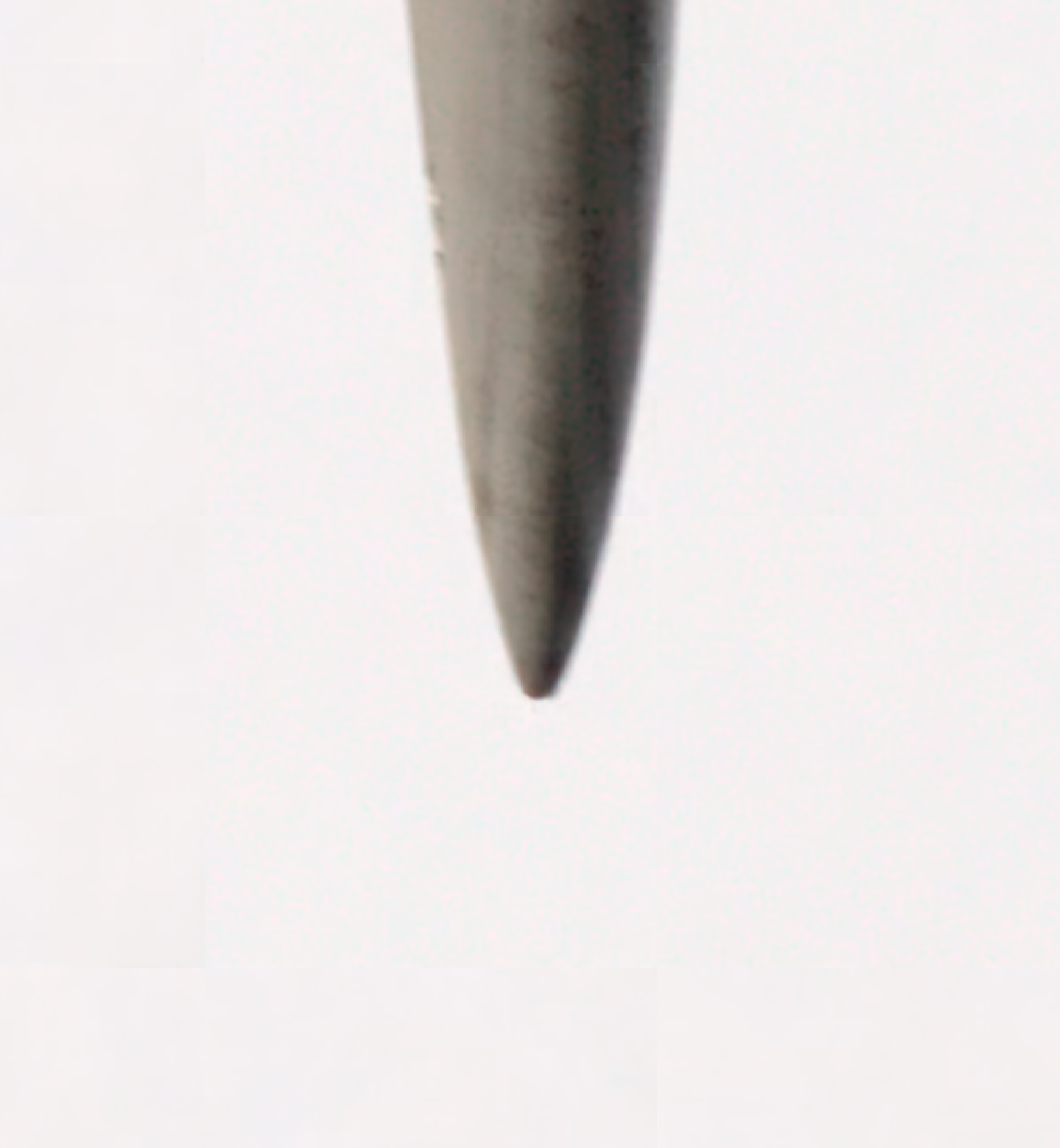}
\caption*{\centering \parbox{\linewidth}{\centering\Huge Blade: 38.09 / 0.9739 / 0.2683 \\ Background: 36.54 / 0.9409 / 0.0197 \\ General: 36.62 / 0.9429 / 0.0332 }}
\end{subfigure}
&
\begin{subfigure}[!h]{0.77\textwidth}
\includegraphics{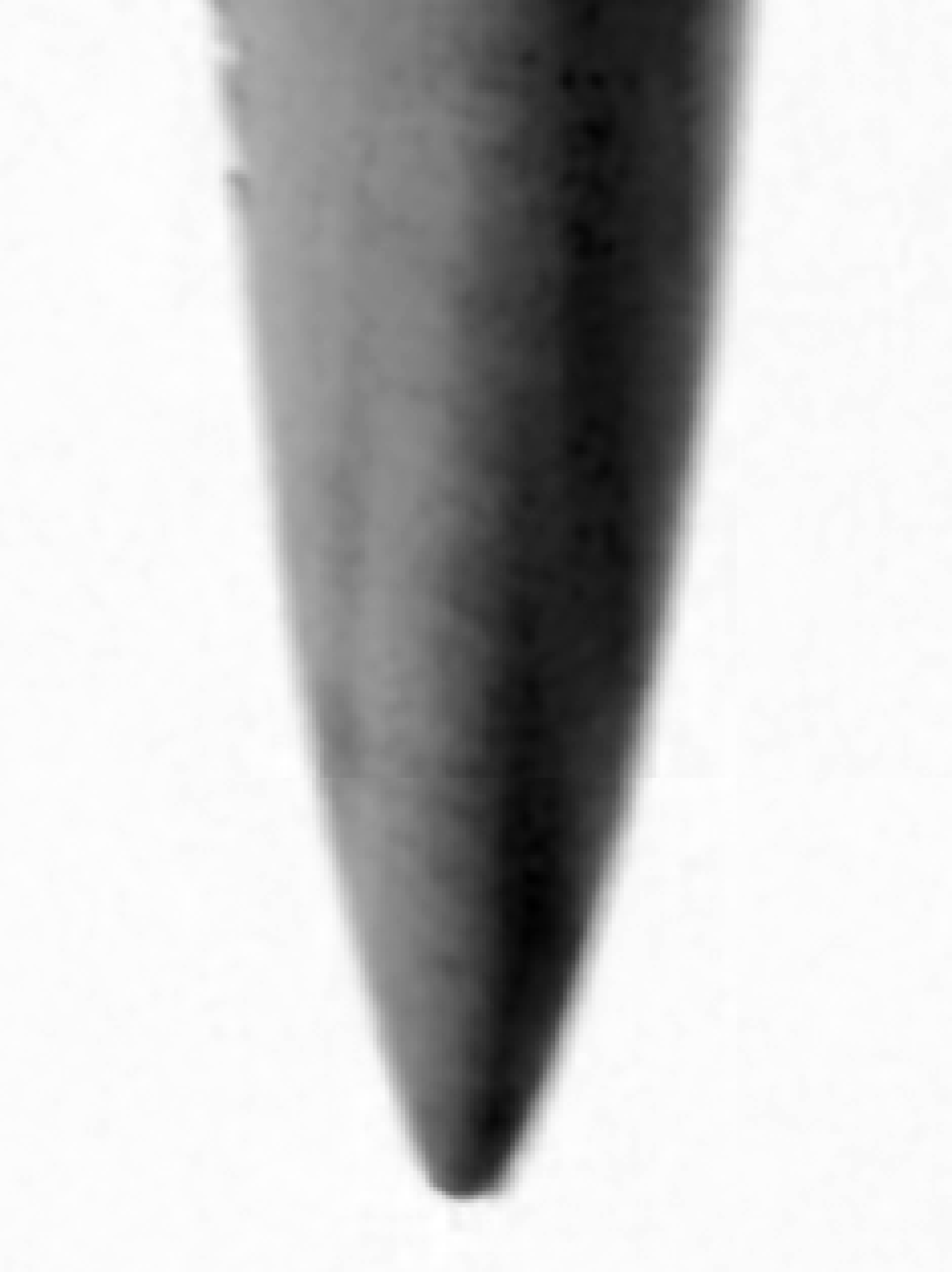}
\caption*{\centering \parbox{\linewidth}{\centering\Huge Zoomed gray-scaled image \\ Lossy blade region\\ \textcolor{white}{.} }}
\end{subfigure}
&
\begin{subfigure}[!h]{0.95\textwidth}
\includegraphics{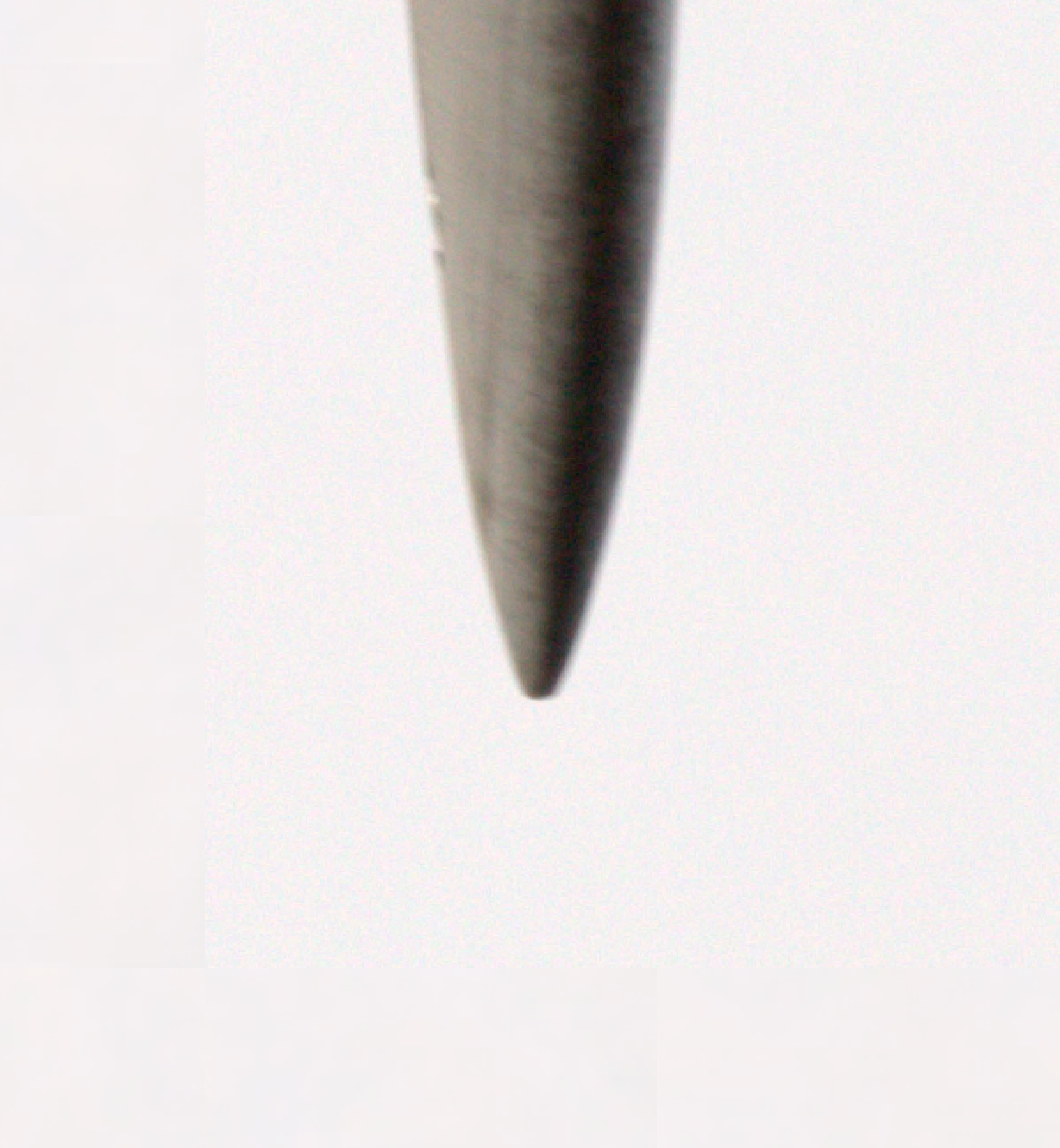}
\caption*{\centering \parbox{\linewidth}{\centering\Huge Blade: 9.5581 bit/px \\ Background: 36.54 / 0.9409 / 0.0197 \\ General: 36.82 / 0.9445 / 0.5381 }}
\end{subfigure}
&
\begin{subfigure}[!h]{0.77\textwidth}
\includegraphics{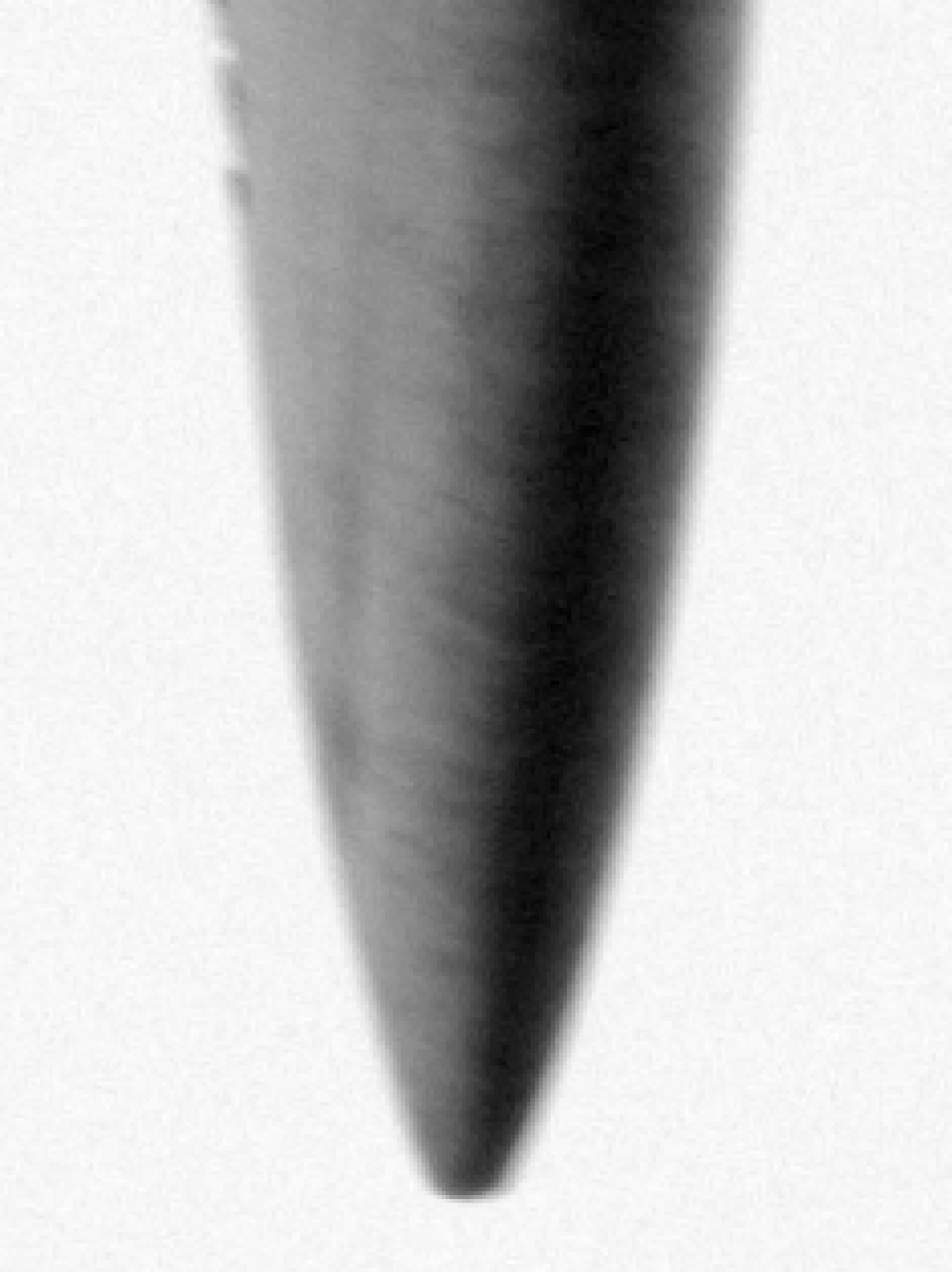}
\caption*{\centering \parbox{\linewidth}{\centering\Huge Zoomed gray-scaled image \\ Lossless blade region \\ \textcolor{white}{.} }}
\end{subfigure}
\end{tabular}}
\end{center} 
\vspace{-0.2cm}
\caption{\textbf{Blade pictures compressed with our proposed ROI algorithm}. The first column showcases the images compressed using the lossy HP+EASN-deep coder with two distinct rate-distortion models on the blade ($[\zeta = 0.1]\cdot$MSE) and background ($[\zeta = 0.01]\cdot$MSE), whereas the third column presents these blade regions compressed losslessly. Second and fourth column represent their left column images augmented and gray-scaled, so the high-frequency components can be visualized and compared. The sub-captions indicate the PSNR, MS-SSIM and bit/px. See Fig.~8 from Supplementary for the original images.}
\label{fig:roi-visualization} \vspace{-0.5cm}
\end{figure*}

We present two blade picture instances after being compressed with our ROI-eML framework in Fig.~\ref{fig:roi-visualization}. The figure displays each image being compressed lossily with two distinct rate-distortion models and losslessly along its blade region. Furthermore, the blade regions are augmented and transformed into gray-scale to better visualize high-level details. In this way, we can remark that the high frequency components of the blade region are no longer present in lossy images. The blade region compressed lossily introduces blurring, losing detail. Lastly, we can observe by zooming the color images on the background region how the texture  due to being compressed in different levels, and the appearance of some artifacts that are discussed in Section III-D from Supplementary.%

\vspace{-0.3cm}
\subsection{Computational Cost} \label{sec:roi-time}

Similarly to the compression performance, the proportion of blade and background regions directly determines the compressing and decompressing inference times. In particular, the total computational cost includes segmenting the image, compressing the background and, lastly, compressing the blade. Runtime aligns with the weighted average of background and blade compression times, based on their respective proportions. The minor overhead from compressing the segmentation mask and loading each model can be disregarded, as does not represent any significant increase. The decompression process follows the same pattern without the segmentation timing.

When both regions are encoded lossily, runtime depends on the two chosen rate-distortion models from HP+EASN-deep (Section~\ref{sec:lossy-time}) and the region proportions. These distinct models spend a really similar time during the encoding, thus, the cost results in the sum of the segmentation (0.089s) and lossy (6.64s, Table~\ref{tab:lossy-compression-time}) algorithms. By contrast, the decompression time varies from 7.86s to 15.35s for the less compressed model, so a typical image decompression of $\frac{2}{3}$ of background region with these two rate-distortion models is roughly $\frac{2}{3} \cdot 7.86s$ plus $\frac{1}{3} \cdot 15.35s$. Hence, the coding time is effectively decreased as we do not require high quality on the background.

If the blade zone is handled losslessly, the overall cost is primarly determined by encoding the blade region, due to the computational burden of Bit-swap. As the background is quickly compressed in lossy mode, a typical image only spends $\frac{1}{3}$ of Bit-swap timing, resulting in 1,896s ($\sim$32min) and 1,455s ($\sim$24min) of compression and decompression time, respectively; compared to around 1h 30min of a lossless compressed image (Table~\ref{tab:lossless-compression-time}). In pictures taken on the blade tip, most of the image is background and the timing is extremely decreased, such is the case for the images in Supplementary Fig. 8, whose ROI compression time is $\sim$10min (10.49\% of blade region) and $\sim$5min (5.43\% of blade region).

Although coding only the blade losslessly reduces running time significantly, further timing reduction is imperative for industrial applications. Thanks to our ROI-eML algorithm, we release the sequential dependency of Bit-swap to compress the blade patches in parallel. Crucially, with enough hardware resources, ROI-eML does perfectly scale. Using just three parallel runs, our ROI algorithm can outpace JXL in terms of speed (Table~\ref{tab:lossless-compression-time}) and maintain its compression performance. 

The amount of parallel runs that our ROI algorithm can perform is bounded, due to limited amount of initial bits from the compressed background. For instance, we require an average of $Bit_{\text{init}}^{\text{optimized}} \approx 25$ initial bit/px on a 64$\times$64 image patch (Section IV-D of Supplementary): a total of 102,400 initial bits. The maximum number of parallel runs depends on the level of compression used. With increased compression, there are fewer available bits, resulting in a reduction of parallel runs. Let us take as an example the HP+EASN-deep model with $1.11$ bit/px and $42.82$ PSNR from Fig.~\ref{fig:lossy-performance}-left or, to be more precise, with $1.14$ bit/px and $42.60$ PSNR in the background region (Section V-A of Supplementary). Each of our 6,704$\times$4,502 images is subdivided into $486$ patches of 256$\times$256 size, due to the mirror padding. This means that each lossy patch is compressed into 74,711 bits. Therefore, each parallel run requires 102,400 / 74,711 = 1.37 compressed image patches for the initial bits or, $2$ patches to avoid splitting the compressed bits of each patch. As a typical image with a portion of $\frac{2}{3}$ of background region has $324$ background patches, we could have a maximum of $162$ parallel runs in our example. Note that the conventional recursive bits-back requires $Bit_{\text{init}}^{\text{recursive}}\approx58$ initial bit/px, implying $4$ background patches per parallel run and, therefore, half of parallel runs.

\vspace{-0.15cm}
\section{Conclusion} \label{sec:conclusion}

This paper addresses the critical challenge of transferring large amount of high-resolution images efficiently in wind turbine scenarios by introducing a novel ROI image compressor. Remarkably, our solution represents a pioneering integration of deep learning techniques, merging segmented image generation with a dual-mode compression strategy encompassing both lossless and lossy coding paradigms. This framework represents a high-quality reconstruction compressor of the blade region, preserving defect detection capabilities.

Extensive evaluation on a comprehensive dataset validate the efficacy of our framework. The introduction of acceptance-ratio curves highlights the superior performance of BU-Netv2+P, attributed to its dense CRF loss and enhanced post-processing. Among various state-of-the-art lossy coders, HP-EASN-deep coding approach demonstrates its efficiency, validated by its encoding efficiency and its low computational cost. Bit-swap proves as the top-performing lossless routine and, by its integration into ROI-eML, it can be effectively designed as parallel algorithm. The synergy of these segmentation and coding capabilities within ROI-eML results in a swift and efficient end-to-end compressor. ROI-eML contributes to enhancing the feasibility and reliability of visual inspections by reducing the data transfer bottleneck and enhancing the quality of blade pictures, positioning it as a potential new compression standard within the entire wind industry.

\noindent \textbf{Acknowledgment.} This work has been supported by the projects 2021 ID1044-0044A of the Innovation Fund Denmark, GRAVATAR PID2023-151184OB-I00 funded by MCIU/AEI/10.13039/501100011033 and by ERDF, UE, and by the project GreenVAR of the Fundación Ramón Areces.

\ifCLASSOPTIONcaptionsoff
  \newpage
\fi

\bibliographystyle{IEEEtran}
\bibliography{References}


\title{End-to-End Image Compression with Segmentation Guided Dual Coding for Wind Turbines\\
--Supplementary Material--} 

\author{Raül~Pérez-Gonzalo\textsuperscript{\orcidaffil{0009-0007-6874-0709}},
Andreas~Espersen\textsuperscript{\orcidaffil{0009-0005-9835-541X}},
Søren~Forchhammer\textsuperscript{\orcidaffil{0000-0002-6698-8870}},~\IEEEmembership{Member,~IEEE},
and Antonio~Agudo\textsuperscript{\orcidaffil{0000-0001-6845-4998}}
 }

\maketitle

\setcounter{equation}{0}
\setcounter{figure}{0}
\setcounter{table}{0}
\setcounter{section}{0}



\section{Wind Turbine Image Data} \label{sec:data}

To train the presented models in order to design our ROI coder, we employed two distinct datasets: a supervised segmentation one for optimizing our BU-Netv2+P, and an unsupervised compression dataset for our lossy and lossless learning-based coders.

\vspace{-0.25cm}
\subsection{Blade Segmentation Dataset} \label{sec:seg-data}
The proposed dataset consists of high-resolution images of $6{,}744 \times 4{,}502$ pixels, captured from various sections of wind turbine blades. This dataset is a curated subset of the compression dataset described in Section~\ref{sec:comp-data}, with corresponding manually annotated masks serving as ground-truth segmentation labels.
Due to be taken in the wild, the images highly vary in appearance, leading to a challenging dataset (see some instances in Fig.~\ref{fig:data}). The blade can diverge in size, shape or illumination conditions. Furthermore, the background is completely distinct depending on the engine used to take the picture. Ground-based images are taken from a robotic arm placed in the floor, thus, the background is generally sky; while drone images are obtained from a closer perspective, capturing generally a vast landscape. Other background variation arises from on/off-shore wind turbines. In general, the expected relation between the background and the blade area is 2:1, but it is worth noticing that in some cases the blade area could be very reduced or, conversely, cover almost entirely the image. 

The training set is comprised of 1,712 images, while the validation and test sets contain 120 and 200 ones, respectively. These images have been gathered from different windfarms and inspection campaigns, ensuring their independence and that the test emulates new data acquired, so we can fairly analyze the generalization of our approach. {More specifically, the validation (6 windfarms) and test (10 windfarms) sets are formed from randomly selecting 20 images per windfarm. Moreover, the training data was selected from a pool of different blade images, prioritizing the under-sampled instances that are hardest to infer. These images cover the blade sections located at the root, tip or max-cord, commonly known as the shoulder of the blade. 

\vspace{-0.25cm}
\subsection{Blade Compression Dataset} \label{sec:comp-data}
We work with a total of 64,438 high-resolution blade images. The image data files are in Canon Raw Format Version 2 format, popularly known as CR2. Each RGB image is 32 MPixels, with a specific image size of 6,744$\times$4,502 pixels.

The image data are split into three sets. The training set is comprised of roughly 80\%, while the validation and test sets have roughly 10\% each one. The split is designed to ensure that images from the same inspection campaign only belong to one set, ensuring that the performance metrics are not biased and properly generalize when receiving new inspection data. During the training procedures, we operate with crops that are randomly selected after loading a set of entire images.

\begin{figure}[!t]
\centering
\includegraphics[width=3.5in]{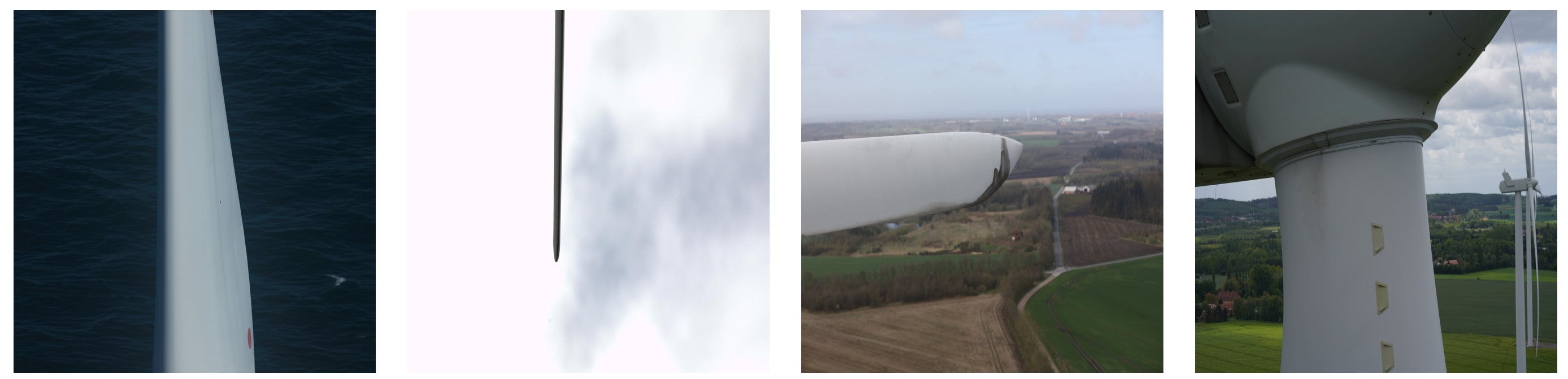}
\caption{\textbf{Four example instances of the blade imagery dataset}. The complexity on this data arises due to the background variation and wild illumination conditions.}
\label{fig:data}
\vspace{-0.5cm}
\end{figure}

\section{BU-Netv2+P}

\subsection{Training Details} \label{sec:training-seg}
The original input images were resized to $256 \times 256$ pixels. The model is trained using a NVIDIA GeForce RTX 3080 Ti. We use Adam solver~\cite{adam} with an initial learning rate of $10^{-4}$, and the size of the mini-batch for gradient descent is equal to 1. A custom scheduler is employed to reduce the learning rate when there is no improvement on the validation loss after three epochs. Early stopping \cite{early} is performed based on the validation accuracy. The CRF regularization term is only trained for the last 15 epochs.  

Each RGB image is min-max normalized. In addition to that, flipping and cropping strategies are applied to squeeze the amount of data available, so the network can learn the desired invariance and robustness properties. 

\subsection{Hyperparameter Search}
The BU-Netv2 parameters were fine-tuned using our validation set to optimize segmentation accuracy. The resulting optimal values are $\tau^{BU}=0.255$, $\gamma=2$, and $\alpha=0.25$ for the focal loss term $\mathcal L_{focal}$, $\eta_{\texttt{loc}}=0.1$ and $\eta_{\texttt{rgb}}=15$ for the Gaussian bandwidths of $\mathcal L_{CRF}$, and $\lambda=0.05$ for the load regularization term.

Regarding the post-processing, the random forest is optimized by investigating which neighboring pixels should be taken as input to boost the performance. The random forest is designed with five tree estimators that have a maximum of four split branches. 

Although the random forest involves some inherent randomness due to bootstrapping and feature sampling, it consistently acts as a denoising step by learning stable patterns across similar images. The variability across runs is negligible: output masks differ by only an average of 0.0067 in pixel values, and this variation is further reduced when ensembles with the BU-Netv2 output.

Instead of picking all the neighboring pixels at a certain distance and the pixel itself, the random forest is only fed with the top, bottom, left and right neighbors at these distance as follows: 

\vspace{-0.35cm}
\begin{equation*} \label{eq:rf-neigh}
     \texttt{neigh}_{n,d}(x_{h,w}) = \bigcup_{i=1}^{n} \{ x_{h+i d,w}, x_{h-i d,w}, x_{h,w+i d}, x_{h,w-i d}\}, 
\end{equation*}
where $n \in \{0,1,\ldots,4\}$ and $d \in \{1,\ldots,5\}$. In so doing, the random forest complexity does not need to be optimized to the input size. Additionally, we reduce the inference time and memory requirements without dropping the segmentation performance. To accomplish the same number of neighboring pixels of all the local pixels, we perform mirror padding.

In Fig.~\ref{fig:seg-rf}, we report the accuracy of the random-forest masks with distinct sets of input pixels, varying the number of neighbors and their spacing. Furthermore, to demonstrate that this step provides a meaningful improvement, we show the accuracy results when the model input does not take any neighboring pixel and after applying the preceding hole-filling step. The result suggests that which neighbors are taken is not significant, as our solution is stable and remains within reasonable bounds. Indeed, the model achieves its maximum performance with a few input pixels, thus, we took as input the local pixel and its most immediate neighbors. 
The input local pixels and the lightweight random forest define a naive model that can be understood as a regularizer of the color intensity of the pixels. Its simplicity indicates that this model is mainly focusing on capturing the pixel intensities to categorize them between blade and background, instead of polishing global features. As we fed the random forest with images taken under similar conditions (same windfarm inspection), it can resolve the single cases where the preceding steps failed, obtaining robustness in the solution.

\begin{figure}[!t]
\centering
\includegraphics[width=3.3in]{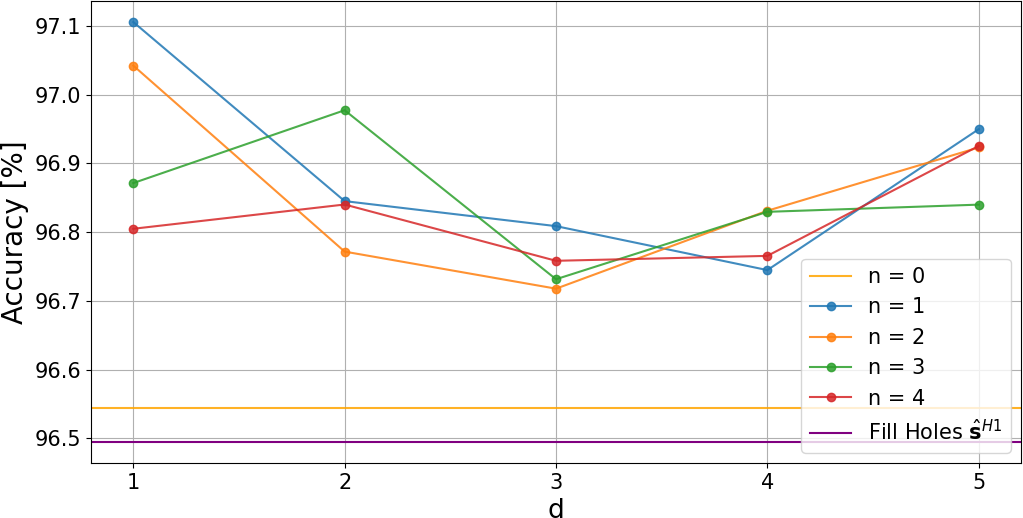}
    \caption{\textbf{Validation accuracy after applying the random forest step with distinct sets of input pixels}. The performance is analyzed in terms of the number of local pixel neighbors taken as input $n$ and the distance $d$ between these input pixels. The purple horizontal line represents the accuracy obtained before applying the random forest step.} 
    \label{fig:seg-rf} \vspace{-0.3cm}
\end{figure}

\subsection{Windfarm Dissimilarity}

As mentioned when introducing the blade dataset, each dataset includes images from distinct windfarms and inspection campaigns. Therefore, in Fig.~\ref{fig:seg-inspections}, we examine how the segmentation performance varies depending on which windfarm the image belongs to. For confidentiality reasons, we have denoted each windfarm by a number.

Figure~\ref{fig:seg-inspections} proves the generability of our segmentation approach, demonstrating that images from all the distinct windfarms have a high performance. Additionally, each windfarm has significant dissimilar performance because test images highly differ depending on the windfarm. For instance, in some windfarms, the images were taken using ground-based equipment and, in others, they were taken by a drone. 

On this basis, the post-processing steps enhance the BU-Netv2 masks, having drastically different effects between distinct windfarms. For instance, hole filling is crucial for Windfarm \#1, while random forest is crucial for Windfarm \#2 and \#7. Hence, it is the post-processing steps as a whole which improves the generability. Notice that the random forest decreases the performance of Windfarm \#4. This decline in the average windfarm accuracy stems from a single image that our algorithm fails to process effectively. 

\begin{figure}[!t]
\centering
\includegraphics[width=3.3in]{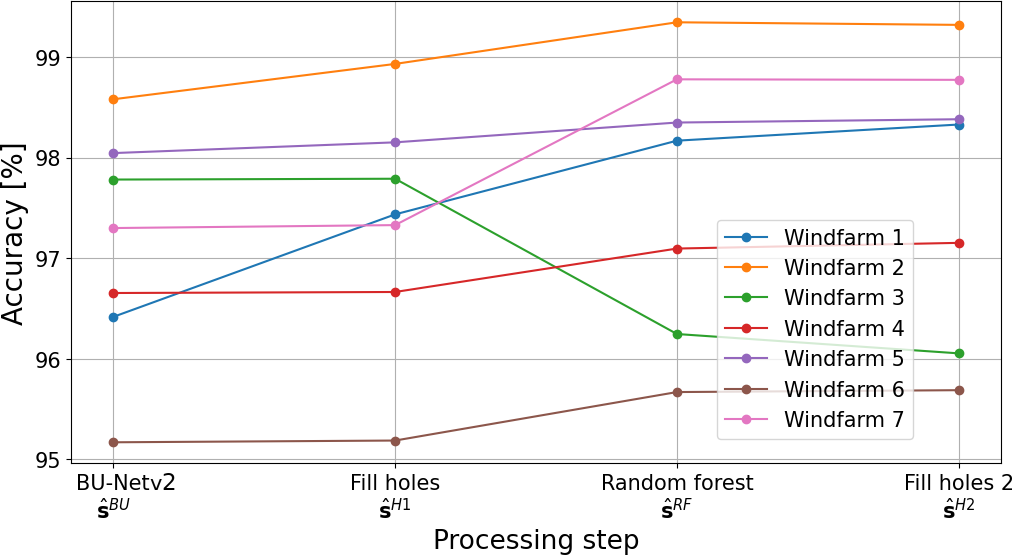}
    \caption{\textbf{Windfarm test performance of each segmentation step}. For readability, the plot does not include every windfarm.}
    \label{fig:seg-inspections} \vspace{-0.3cm}
\end{figure}

\subsection{Qualitative Results} \label{sec:seg-qual}

\begin{figure*}[t!]
\begin{center}
\resizebox{18cm}{!} {
\begin{tabular}{@{}c@{}c|c|c@{}}
\hspace{-0.35cm} \makebox[0.02\textwidth][c]{\raisebox{0.5cm}{\includegraphics[width=0.05\textwidth]{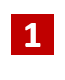}}}&
\hspace{0.01cm} \hspace*{0.02cm} \includegraphics[width=0.285\textwidth]{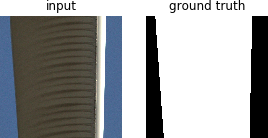}&
\includegraphics[width=0.59\textwidth]{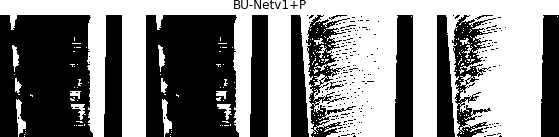}&
\includegraphics[width=0.59\textwidth]{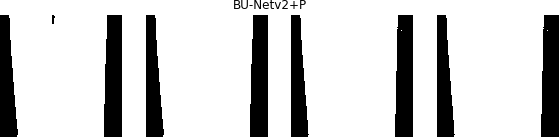}\\ %
\hspace{-0.35cm} \makebox[0.02\textwidth][c]{\raisebox{0.5cm}{\includegraphics[width=0.05\textwidth]{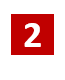}}}&
\hspace{0.01cm} \hspace*{0.02cm} \includegraphics[width=0.285\textwidth]{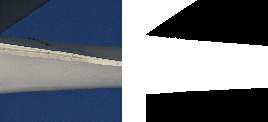}&
\includegraphics[width=0.59\textwidth]{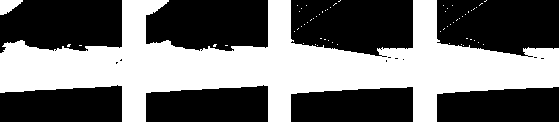}&
\includegraphics[width=0.59\textwidth]{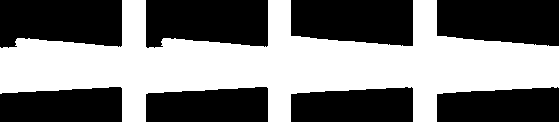}\\ %
\hspace{-0.35cm} \makebox[0.02\textwidth][c]{\raisebox{0.5cm}{\includegraphics[width=0.05\textwidth]{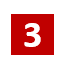}}}&
\hspace{0.01cm} \hspace*{0.02cm} \includegraphics[width=0.285\textwidth]{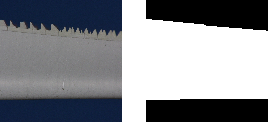}&
\includegraphics[width=0.59\textwidth]{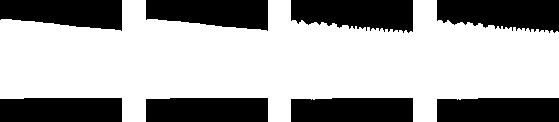}&
\includegraphics[width=0.59\textwidth]{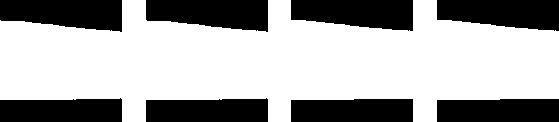}\\ %
\hspace{-0.35cm} \makebox[0.02\textwidth][c]{\raisebox{0.5cm}{\includegraphics[width=0.05\textwidth]{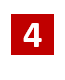}}}&
\hspace{0.01cm} \hspace*{0.02cm} \includegraphics[width=0.285\textwidth]{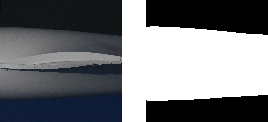}&
\includegraphics[width=0.59\textwidth]{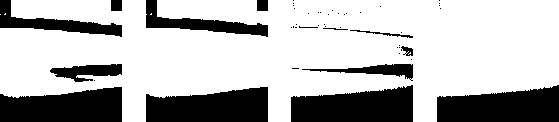}&
\includegraphics[width=0.59\textwidth]{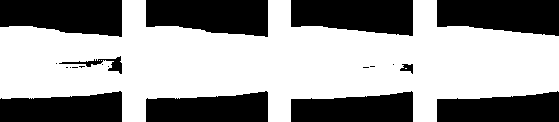}\\ %
\hspace{-0.35cm} \makebox[0.02\textwidth][c]{\raisebox{0.5cm}{\includegraphics[width=0.05\textwidth]{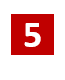}}}&
\hspace{0.01cm} \hspace*{0.02cm} \includegraphics[width=0.285\textwidth]{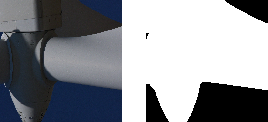}&
\includegraphics[width=0.59\textwidth]{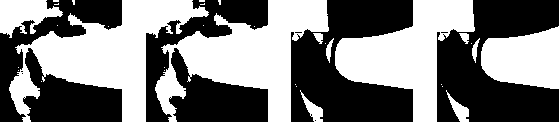}&
\includegraphics[width=0.59\textwidth]{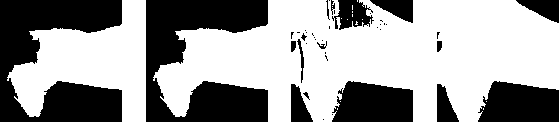}\\ %
\hspace{-0.35cm} \makebox[0.02\textwidth][c]{\raisebox{0.5cm}{\includegraphics[width=0.05\textwidth]{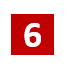}}}&
\hspace{0.01cm} \hspace*{0.02cm} \includegraphics[width=0.285\textwidth]{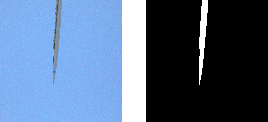}&
\includegraphics[width=0.59\textwidth]{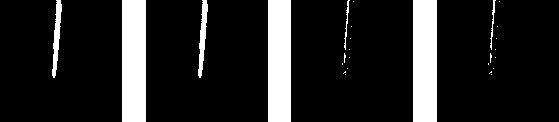}&
\includegraphics[width=0.59\textwidth]{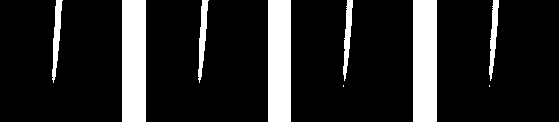}\\ %
\end{tabular}}
\end{center}
\vspace{-0.2cm}
\caption{\textbf{Qualitative comparison evaluation between BU-Netv1+P and our approach BU-Netv2+P}. The figure depicts on the left side the test images $\mathbf{x}$ and its ground truth mask $\mathbf{s}$; on the inner side the segmentation steps of BU-Netv1+P (from left to right columns: BU-Netv1, first hole filling, random forest, second hole filling); and on the right side the  the steps of our proposed segmentation algorithm BU-Netv2+P (from left to right columns: BU-Netv2 ${\hat{\mathbf{s}}}^{BU}$, first hole filling $\hat{\mathbf{s}}^{H1}$, random forest ensemble $\hat{\mathbf{s}}^{RF}$ and second hole filling $\hat{\mathbf{s}}\equiv\hat{\mathbf{s}}^{H2}$).}
\label{fig:seg-visual}
\end{figure*}

\begin{figure*}[t!]
\begin{center}
\resizebox{18.cm}{!} {
\begin{tabular}{@{}c@{}c|c@{}c@{}}
\hspace{-0.35cm} \makebox[0.02\textwidth][c]{\raisebox{0.35cm}{\includegraphics[width=0.03\textwidth]{Plots/numbers/1.png}}}&
\hspace{0.01cm} \includegraphics[width=0.4\textwidth]{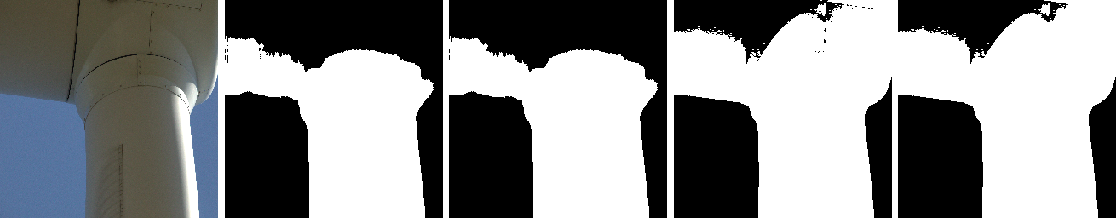}&
\hspace{-0.15cm} \makebox[0.02\textwidth][c]{\raisebox{0.35cm}{\includegraphics[width=0.03\textwidth]{Plots/numbers/3.png}}}&
\hspace{0.01cm} \includegraphics[width=0.4\textwidth]{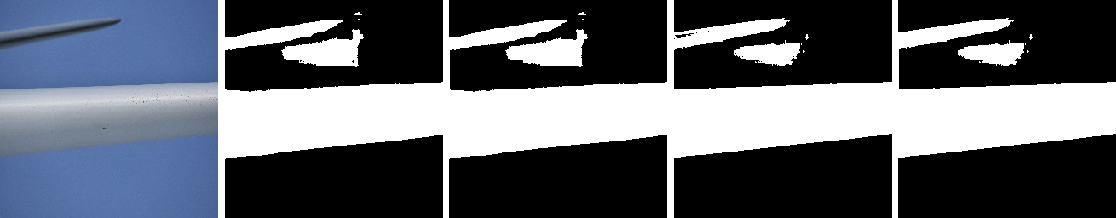}\\
\hspace{-0.3cm} \makebox[0.02\textwidth][c]{\raisebox{0.35cm}{\includegraphics[width=0.03\textwidth]{Plots/numbers/2.png}}}&
\hspace{0.01cm} \includegraphics[width=0.4\textwidth]{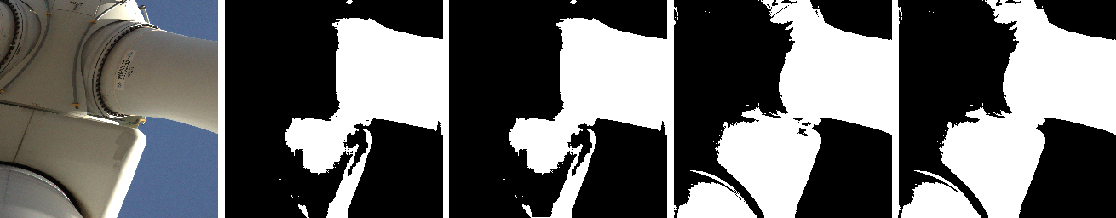}&
\hspace{-0.15cm} \makebox[0.02\textwidth][c]{\raisebox{0.35cm}{\includegraphics[width=0.03\textwidth]{Plots/numbers/4.png}}}&
\hspace{0.01cm} \includegraphics[width=0.4\textwidth]{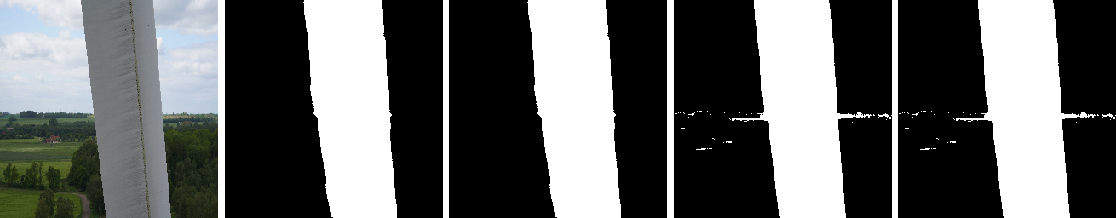}
\end{tabular}}
\end{center}
\vspace{-0.4cm}
\caption{\textbf{Failure cases on test images after running each segmentation step}. First column: input color image $\mathbf{x}$. From second to fifth column: BU-Netv2 ${\hat{\mathbf{s}}}^{BU}$, first hole filling $\hat{\mathbf{s}}^{H1}$, random forest ensemble $\hat{\mathbf{s}}^{RF}$ and second hole filling $\hat{\mathbf{s}}\equiv\hat{\mathbf{s}}^{H2}$ estimations, respectively. On both sides, the same information is displayed.}
\label{fig:seg-failure} \vspace{-0.3cm}
\end{figure*}

In Fig.~\ref{fig:seg-visual}, we illustrate with six instances in which typical scenarios the BU-Netv2+P outperforms its preceding BU-Netv1+P~\cite{PerezGonzaloIcip2023}. The huge gain comes with the BU-Netv2 output, which can be clearly seen on the first and fourth instance. Thanks to the BU-Netv2 capacity to infer significant better masks, the subsequent steps can focus on refining the solution and we do not longer depend on post-processing for duly identifying the segmentation masks. Indeed, the first and fourth instance are not properly refined for the BU-Netv1+P~\cite{PerezGonzaloIcip2023}, which is not the case for BU-Netv2+P. In addition to this, the BU-Netv2 is stronger capturing the shadow regions as depicted in the fifth instance, because its regularization CRF loss promotes continuous solutions.

Another substantial asset from our proposal is adopting an ensemble strategy for the random forest step. In this way, the random forest --which captures the color intensities patterns of a set of images taken under similar conditions-- can balance its prediction with the BU-Netv2 logits, which are inferred in a more in-depth manner. Hence, the segmentation masks of this step can join forces of the two models and enhance region predictions with higher certainty. Accordingly, the random forest ensemble is less vulnerable to mislead shadows in the blade region. This is the case for the second and the fifth instance. But even more importantly, it also does not mislead possible defects which typically do have a distinct color to the rest of the blade, which is the case for the third instance.

Despite that the hole filling steps quantitatively do not greatly boost performance, they remain crucial for achieving a detail-aware solution: the first hole filling step enhances the masks from the first and fourth instance and, most critically, the second hole filling is essential to deliver the remarkable ultimate masks for the fourth and fifth instances.

\vspace{-.4cm}
\subsection{Failure Analysis} \label{sec:seg-failure}
\vspace{-.1cm}
As mentioned, our approach is more robust than the competing methods, however, there is room for improvement in specific cases as shown in Fig.~\ref{fig:seg-failure}. First, despite the BU-Netv2 contains a contiguity regularizer that helps the network, the model is not able to cope with all images on the root. This is due to lack of instances of that kind along with being dissimilar in shape to the rest of the images. Hence, our approach can successfully capture the typical blade region that appears on this type of images, but it struggles with the rotor. That is the case for the first two instances of Fig.~\ref{fig:seg-failure}.

In addition to that, an image could include wind turbine regions at its background. Ideally, we would like to capture those regions only when they belong to the same wind turbine as the foreground blade. Hence, this task is notably challenging. For this reason, the BU-Netv2 can get confused as shown in the third instance on the right side of Fig.~\ref{fig:seg-failure} and, depending on the network mask, the post-processing steps are not sufficient to resolve it.

Lastly, another possible failure that might occur is the presence of small artifacts. This is usually caused by the random forest step. Although the ensemble strategy can address most cases, when some pixel intensities on the background and on the blade region are really similar, the random forest may infer with high confidence that those pixels on the background belong to the blade region. It is in these cases where the ensemble cannot revert it and, therefore, we can see artifacts as illustrated in the fourth instance of Fig.~\ref{fig:seg-failure}.

\subsection{Computational cost}
Figure~\ref{fig:seg-time} presents the average inference runtimes for each step in our algorithm. It also includes the cumulative time, illustrating that the overall running time for segmenting an image is 0.089s.

\begin{figure}[!t]
\centering
\includegraphics[width=3.3in]{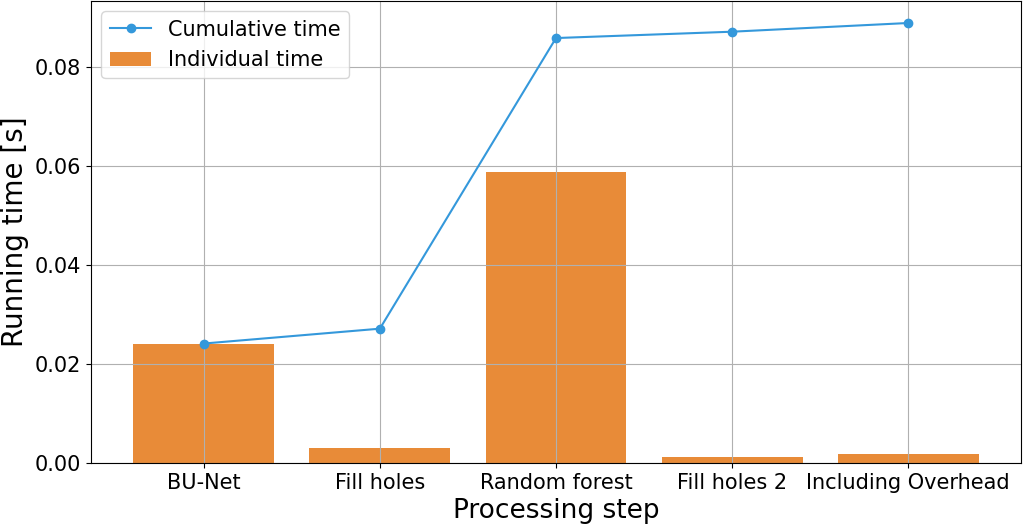}
\caption{\textbf{Computational time in seconds of an individual image for each segmentation step}. Computational time for the BU-Net, hole filling and random forest steps are reported. The blue line represents the cumulative time.}
\label{fig:seg-time}
\vspace{-0.5cm}
\end{figure}

\section{HP+EASN-deep}
\subsection{Training Details} \label{sec:training-lossy}
The model is fed with patches of 256$\times$256 pixels following HP~\cite{hyperprior} with a mini-batch size for gradient descent of 8. The model is trained using a NVIDIA GeForce RTX 3080 Ti. We use the Adam optimizer with an initial learning rate of $10^{-4}$. Two distinct distortion metrics are trained: the squared error $d_2$ and the negative Multi-Scale Structural SIMilarity (MS-SSIM)~\cite{msssim}, which better captures the human perceptual judgments of image quality. To prevent very large gradients, the likelihood $\tilde{p}_{\boldsymbol{\phi}}(\tilde{\mathbf{z}})$ of the rate term is lower bounded by $10^{-9}$. To avoid overfitting, we perform early-stopping and in each epoch we randomly generate distinct image patches by cropping them from the full-resolution images.

\subsection{Discretization of the Latent Space} 
After training, the cumulative density function $c$ is discretized to apply entropy coding to the quantized latents $\bar{\mathbf{z}}_l$. As entropy coders cannot operate with continuous distributions, the latent probability models are discretized to the densities $\bar{p}_{\boldsymbol{\theta}}(\bar{\mathbf{z}}_l)$ for $l \in \{1,2\}$. Every latent $M$-dimensional variable component $\mathbf{z}_l = (z_{l,1}, \ldots, z_{l,M})$  is rounded to the nearest integer value, discretizing the latent space into bins of unit size. Let $\bar{z}_{l,j} \in \mathds{Z}$ be the quantized value of $z_{l,j}$, then we define the discretized probability $\bar{p}_{\boldsymbol{\theta}}(z_{l,j})$ and its multivariate distribution as:

{\small
\begin{align} \label{eq:quantized_factorized_prior}
    \bar{p}_{\boldsymbol{\theta}}(\bar{z}_{l,j}) &= \int_{\bar{z}_{l,j} - 0.5}^{\bar{z}_{l,j} + 0.5} p_{\boldsymbol{\theta}}(z_{l,j}) dz_{l,j} = c(\bar{z}_{l,j} + 0.5) - c(\bar{z}_{l,j} - 0.5) ,\nonumber\\
    \bar{p}_{\boldsymbol{\theta}}(\bar{\mathbf{z}}_l) &= \prod_{j=1}^{M} %
     \bar{p}_{\boldsymbol{\theta}}(\bar{z}_{l,j}) .
\end{align} }

\vspace{-0.1cm}
Due to the impracticability to evaluate every possible integer $\bar{z}_{l,j}$, we calculate the quantized distribution for values between a lower and an upper bound on $\bar{z}_{l,j}$ for each $l$. In this way, if we denote $\texttt{med}(z_{l,j})$ the median of $p_{\boldsymbol{\theta}}(z_{l,j})$, the discretized probability is only computed for latent variables in the range of $[ \texttt{med}(z_{l,j}) - z_{l,\min} , \texttt{med}(z_{l,j}) + z_{l,\max} ]$, where $z_{l,\min}$ and $z_{l,\max}$ are determined so that the integers in the tails of the distribution are the only non-evaluated values. Hence, they are obtained as the largest distance observed along the distinct components $j$ between the lower tail quantile and the median, and between the median and the upper tail quantile. In particular, the tail masses are chosen to be $10^{-9}$. We now provide the concrete definition of $z_{l, \min}$ and $z_{l, \max}$ as:
\begin{align*}
    \begin{split}
    z_{l, \min} &= \max_{j} \{ \texttt{med}(z_{l,j}) - z_{l,j}^{-}  \text{,} \\ 
    &\quad \text{where } z_{l,j}^{-} \text{ is the } z_{l,j} \text{ s.t. } c(z_{l,j})=10^{-9} \} \\
    z_{l, \max} &= \max_{j} \{ z_{l,j}^{+} - \texttt{med}(z_{l,j}) \text{,}  \\ 
    &\quad \text{where } z_{l,j}^{+} \text{ is the } z_{l,j} \text{ s.t. } c(z_{l,j})=1-10^{-9}  \}  .
    \end{split}
\end{align*} 

To ensure replicability, we provide below some insights:

\begin{itemize}
    \item The quantile tails of the prior probability $\{\mathbf{z}_{2}^{-}, \mathbf{z}_{2}^{+}\}$ and the median $\texttt{med}(\mathbf{z}_{2})$ are obtained through optimization via stochastic gradient descent. The numerically stability is guaranteed by the logit function. Moreover, the quantile tails of $\bar{p}_{\boldsymbol{\theta}}(\bar{\mathbf{z}}_1)$ are easily computed through the inverse cumulative function of the Gaussian distribution.
    \item For $\bar{z}_{l,j}$ outside of the ranges, $\bar{p}_{\boldsymbol{\theta}}(\bar{z}_{l,j}) = 10^{-9}$. Thus, outliers of $\bar{z}_{l,j}$ are coded as either $\texttt{med}(z_{l,j}) - z_{l,\min} - 1$ or $\texttt{med}(z_{l,j}) + z_{l,\max} + 1 $.
    \item The resulting coding alphabet for the $l$-th latent $j$-th  component is $\mathcal{A}_{l,j} = \{\texttt{med}(z_{l,j}) - z_{l,\min} - 1, \ldots,   \texttt{med}(z_{l,j}), \ldots,  \texttt{med}(z_{l,j}) + z_{l,\max} + 1 \}$.
    \item We use range coding \cite{rangecoding} as the entropy coder. The range coder precision is set to $16$. 
\end{itemize}

\subsection{Side information}

We have analyzed the amount of side information when compressing a full-resolution image. The side information includes the additional bits of information to successfully reconstruct the original image, which includes the padding dimensions, the original image size and the rate-distortion model that has been used. Furthermore, the side information also includes the additional bits due to the discretization process. Hence, this side information is what differs between the rate term of the rate-distortion loss and the actual bit rate after compressing an image. Figure~\ref{fig:side_information} shows the amount of bit rate used as side information compared to the total bit rate. The side information rate grows linearly, but it only uses a small fraction of the total bit rate which is marginal.

\begin{figure}[!t]
\centering
\hspace{-0.1cm} \includegraphics[width=2.9in]{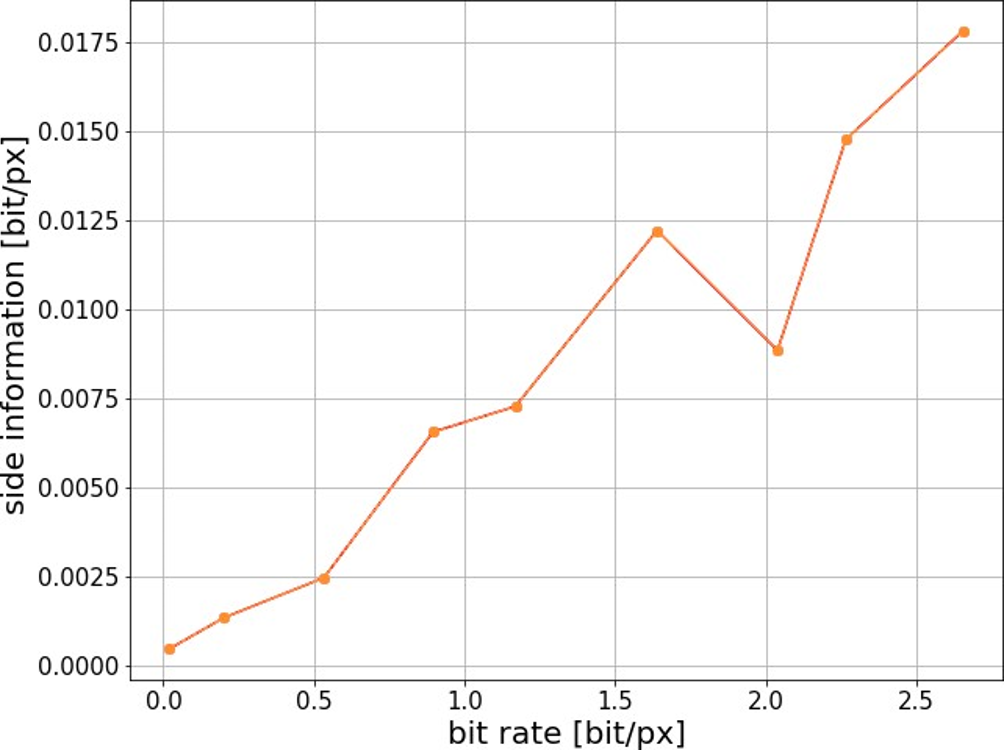}
\caption{\textbf{Additional bits per pixel of information} required to decompress each rate-distortion model of HP+EASN-deep.}
\label{fig:side_information}
\end{figure}

\vspace{-0.2cm}
\subsection{Visualization of the Compressed Images} \label{sec:lossy-visualization}

\begin{figure*}[t!]
\begin{center}
\resizebox{17.8cm}{!} {
\begin{tabular}{@{}c@{}c@{}}
\hspace{-0.2cm} \includegraphics[width=0.4\textwidth]{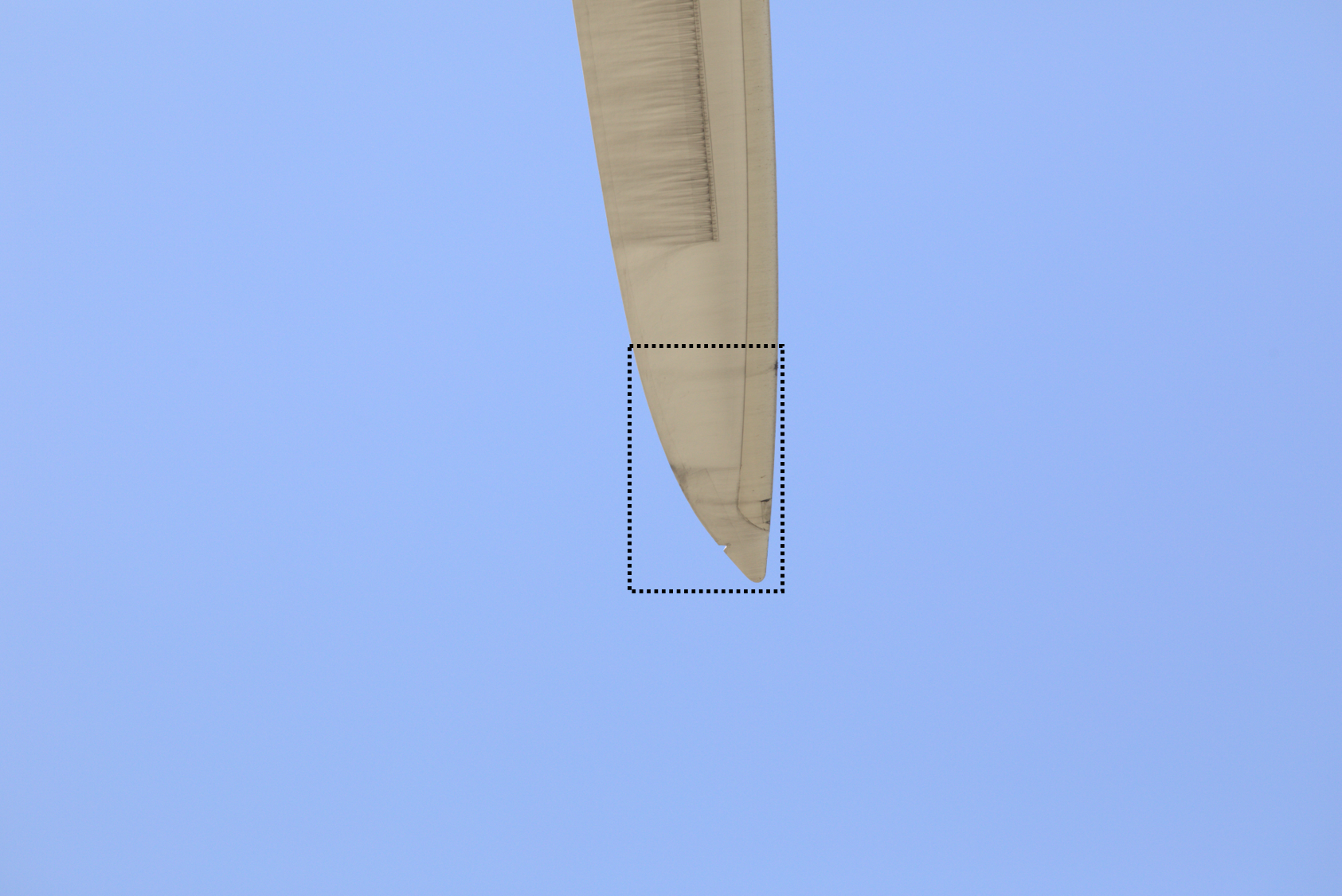}&
\includegraphics[width=0.4\textwidth]{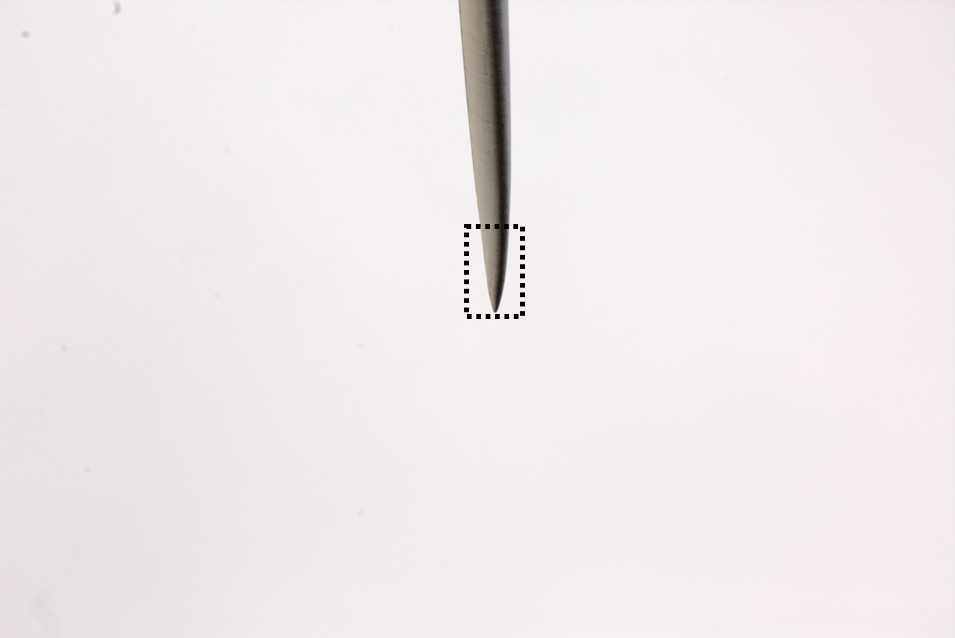}
\end{tabular}}
\end{center}
\vspace{-0.9cm}
\end{figure*}

\begin{figure*}[t!]
\begin{center}
\resizebox{17.8cm}{!} {
\begin{tabular}{@{}c@{}c@{}c@{}c@{}c@{}c@{}}
\begin{subfigure}[!h]{0.4\textwidth}
\includegraphics{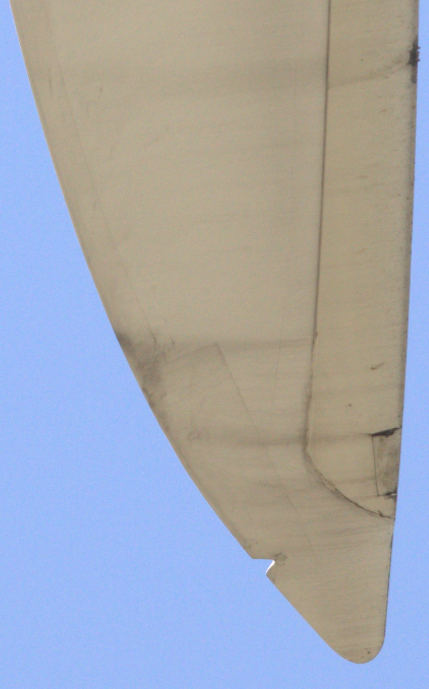}
\caption*{\centering \LARGE{Original image \newline PSNR / MS-SSIM / BPP} }
\end{subfigure}
&
\begin{subfigure}[!h]{0.4\textwidth}
\includegraphics{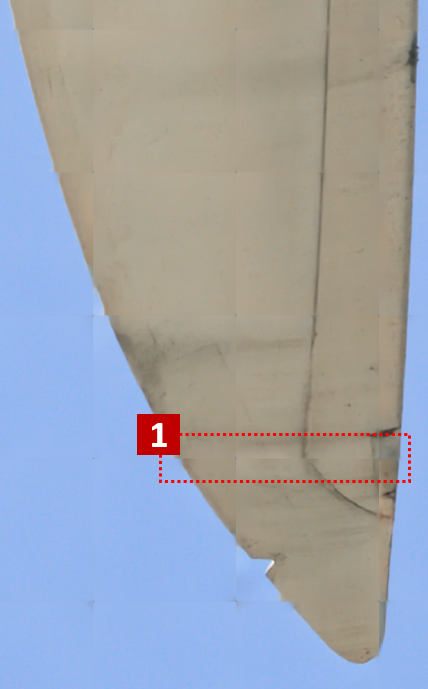}
\caption*{\centering \LARGE{$[\zeta = 10]\cdot(1-$MS-SSIM$)$  \newline 31.9095 / 0.9553 / 0.1312}} 
\end{subfigure}
&
\begin{subfigure}[!h]{0.4\textwidth}
\vspace{-0.07cm}\includegraphics{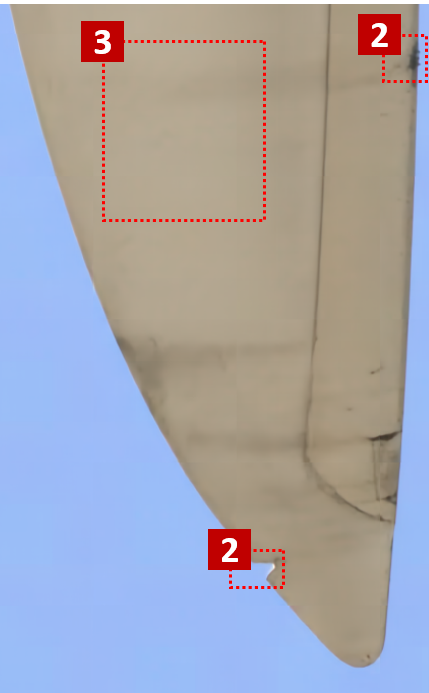}
\caption*{\centering \LARGE{$[\zeta = 0.01]\cdot$MSE  \newline 35.7147 / 0.9326 / 0.01341} } 
\end{subfigure}
&
\begin{subfigure}[!h]{0.4\textwidth}
\includegraphics{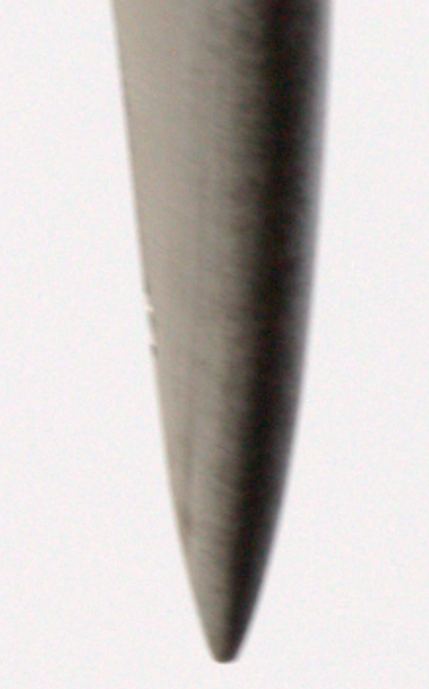}
\caption*{\centering \LARGE{Original image \newline PSNR / MS-SSIM / BPP} }
\end{subfigure}
&
\begin{subfigure}[!h]{0.4\textwidth}
\includegraphics{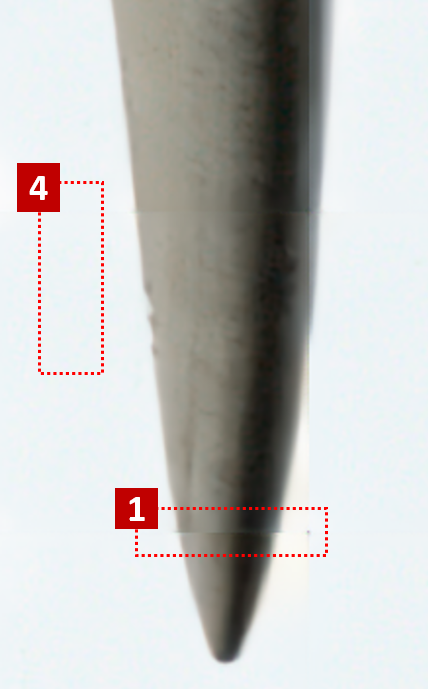}
\caption*{\centering \LARGE{$[\zeta = 10]\cdot(1-$MS-SSIM$)$  \newline  31.2675 / 0.9567 / 0.1370}  }
\end{subfigure}
&
\begin{subfigure}[!h]{0.4\textwidth}
\vspace{-0.17cm}\includegraphics{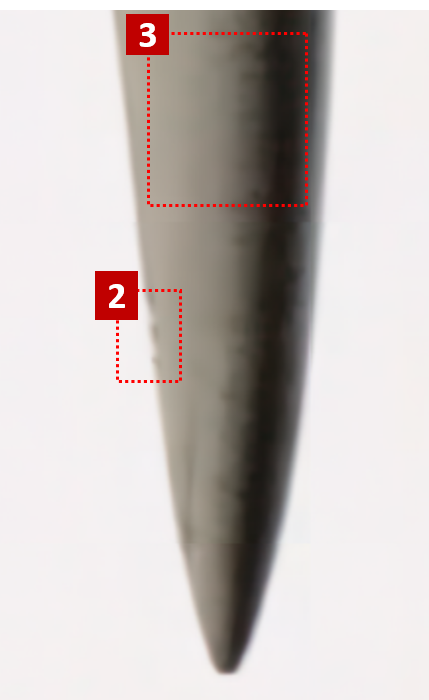}
\caption*{\centering \LARGE{$[\zeta = 0.01]\cdot$MSE  \newline 36.1810 / 0.9369 / 0.0177}  }
\end{subfigure}

\end{tabular}}
\end{center}
\vspace{-0.2cm}
\caption{\textbf{Quality assessment of two blade pictures with high compression rates}. The first row showcases the complete original images, whereas the second row presents a zoomed-in image accompanied by two different compression rates. Sub-captions indicate the distortion loss term and the compression performance. Red regions indicated with numbers pinpoint the following artifacts: 1) blocking artifacts, 2) smoothed contours, 3) smoothed textures and 4) visual color scale change.} \vspace{-0.35cm}
\label{fig:lossy-visualization}
\end{figure*}

We would like to study which compression artifacts could disrupt the manual detection of blade defects for HP+EASN-deep. To this end, we present two blade pictures that have been compressed using low bit rates with HP+EASN-deep. The pictures have been divided into individual 256$\times$256 patches that are compressed independently. Note that for low bit rates, the compression performance of HP+EASN-deep is inferior to some traditional coders such as BPG or VTM. Indeed, from a practical point of view, the industry would not tolerate such a low quality.

Both pictures have been compressed using two rate-distortion models trained over distinct distortion loss terms: $[\zeta = 0.01]\cdot$MSE and $[\zeta = 10]\cdot(1-$MS-SSIM$)$. 
Since each model was trained using a different distortion metric, the reconstructed pictures reflect these differences.  For example, although the image optimized for MS-SSIM has a much higher bit rate and higher MS-SSIM, its PSNR is significantly lower. This is expected, as the MS-SSIM optimized model does not prioritize MSE minimization. An analogous behaviour can be seen in the image optimized for MSE.

The two selected pictures contain tiny surface defects, thus, we require that those defects are recognizable after decompression. It is also necessary that the contaminated regions are not wrongly detected as a false positive defect. Figure~\ref{fig:lossy-visualization} illustrates the quality of these blade pictures after conducting a high compression. Using red rectangles, we have highlighted the typical compression artifacts that could appear.

In general terms, these pictures compressed with our method lose detail for low bit rates. The edges and texture of the blade are smoothed, eliminating and blurring many small patterns. The contours and sharpness of many of the edges are closed, producing smoothed and convex objects. These modifications can directly impact on the perception of tiny defects (see red rectangles pinpointed with a 2), such as misleading its category or therefore its severity. The worse situation, of course, would be to confuse a defect with some dirt.

Furthermore, blocking artifacts are generated because the pictures have been cropped into image patches that are independently compressed. The two pictures illustrate how artifacts vary with bit rate. The image optimized for MSE, which has a much lower bit rate, shows minimal blocking artifacts typically introduced by patch-based encoding. This is because such artifacts are hardly visible due to increased blurring.

Finally, it is important to note that the visual perception of the color scale in the image may differ significantly, although the distortion metric does not discern from the original image to a great extent. This is caused because MS-SSIM does not prioritize the color space difference as MSE does. 

To sum up, HP+EASN-deep progressively simplifies the contours and features of the image when performing a high compression. This may give an artificial appearance and may affect the defect detection performance. 

\subsection{Traditional Codec Settings}

This subsection details the settings used to evaluate the performance of standard lossy image compression codecs: JPEG2000~\cite{J2K}, WebP~\cite{webp}, BPG~\cite{bpg}, and VTM~\cite{vtm}, for consistent comparison with our proposed method.

For JPEG2000, we use the official and reference implementation \texttt{OpenJPEG} (v2.5.0). Compression is done using:
\begin{HighlightVerbatim}
opj_compress -i input.png -o out.j2k -q PSNR 
\end{HighlightVerbatim}
where \texttt{PSNR} is the desired peak signal-to-noise ratio target, with tested values in the range $[35.5, 37, 38.5, 40, 42]$. Decoding is handled by:
\begin{HighlightVerbatim}
opj_decompress -i out.j2k -o output.png
\end{HighlightVerbatim}

For WebP, we use \texttt{libwebp} version 1.2.2. Compression is performed with:
\begin{HighlightVerbatim}
cwebp -q QP input.png -o output.webp
\end{HighlightVerbatim}
where \texttt{QP} indicates the quality factor, chosen from $[15, 40, 80, 85, 90, 95]$. The compressed image is decompressed using:
\begin{HighlightVerbatim}
dwebp output.webp -o output.png
\end{HighlightVerbatim}

For BPG, we use \texttt{libbpg} version 0.9.8, using:
\begin{HighlightVerbatim}
bpgenc -o output.bpg -q QP -f 444 -e x265 
-c ycbcr -b 8 input.jpg
\end{HighlightVerbatim}
We evaluate quality levels with \texttt{QP} values of $[20, 22.5, 25, 30, 40, 50]$ using 4:4:4 chroma subsampling. Decoding is executed with:

\begin{HighlightVerbatim}
bpgdec -o output.png output.bpg
\end{HighlightVerbatim}

For VTM (VVC Test Model), version 9.1 is used. The input images are first converted from RGB to YUV444 format before encoding with:
\begin{HighlightVerbatim}
EncoderApp -i input.yuv -o output.yuv -q QP 
-c encoder_intra_vtm.cfg  -b output.bin 
--InputChromaFormat=444 --InputBitDepth=8
-fr 1 -f 1 -wdt W -hgt H 
\end{HighlightVerbatim}
The parameter \texttt{QP} is set to $[22, 25, 30, 35, 40, 45, 55]$, and \texttt{W} and \texttt{H} correspond to the original image dimensions. The decoding process is performed using:
\begin{HighlightVerbatim}
DecoderApp -b input.bin -o output.yuv -d 8
\end{HighlightVerbatim}
Finally, the decoded YUV444 images are converted back to RGB to evaluate its reconstruction quality.

\subsection{Computational Cost}

Figure~\ref{fig:lossy-time} compares compression and decompression times for full-resolution images across various state-of-the-art lossy compression methods. Traditional coders like BPG~\cite{bpg}, WebP~\cite{webp}, and JPEG2000~\cite{J2K} offer fast, consistent runtimes with compression times around 4-5 seconds and slightly higher decoding times, reflecting their efficient, optimized pipelines.

Learning-based methods show greater variation. Lightweight models such as HP~\cite{hyperprior} and HP+EASN-deep achieve low latency, with compression under 7 seconds and decompression near 12.8 seconds, making them practical for real-time applications. In contrast, autoregressive approaches like GMM-Attn~\cite{gmm} and JA~\cite{ja} incur significantly higher costs (e.g., GMM-Attn: 252.81s compression, 508.58s decompression). ELIC~\cite{elic}, MLIC++~\cite{mlic++} and LIC-TCM~\cite{lic-tcm} provide a trade-off between complexity and speed.

VTM~\cite{vtm}, while offering top-tier compression performance compared to traditional coders, suffers from the highest compression time (up to 9943s) due to its extensive rate-distortion optimization. Despite this, its decoding speed remains comparatively fast. Its extreme computational cost stems from exhaustive rate-distortion optimization and large search space, which scale poorly with image resolution and bitrate.

Compression time generally increases with bit rate, but the growth behavior varies by method. For VTM, BPG, and WebP, the upward-curving lines in the log-scale plot suggest superlinear, such as near exponential growth with bit rate. Conversely, models like HP+EASN-deep show linear scaling, as reflected in their straight-line trends and slopes reported in Table III from the main manuscript. This reflects low computational complexity and consistent encoding efficiency across bit rates, making HP+EASN-deep more scalable for practical applications in resource-limited environments.

Overall, HP+EASN-deep offers a strong balance between speed and quality, making it especially suited for industrial settings like wind turbine blade inspection, where runtime efficiency is critical.

\begin{figure*}[!t]
\centering
\resizebox{18cm}{!}{
\begin{tabular}{@{}c@{}}   
\includegraphics[width=3.3in]{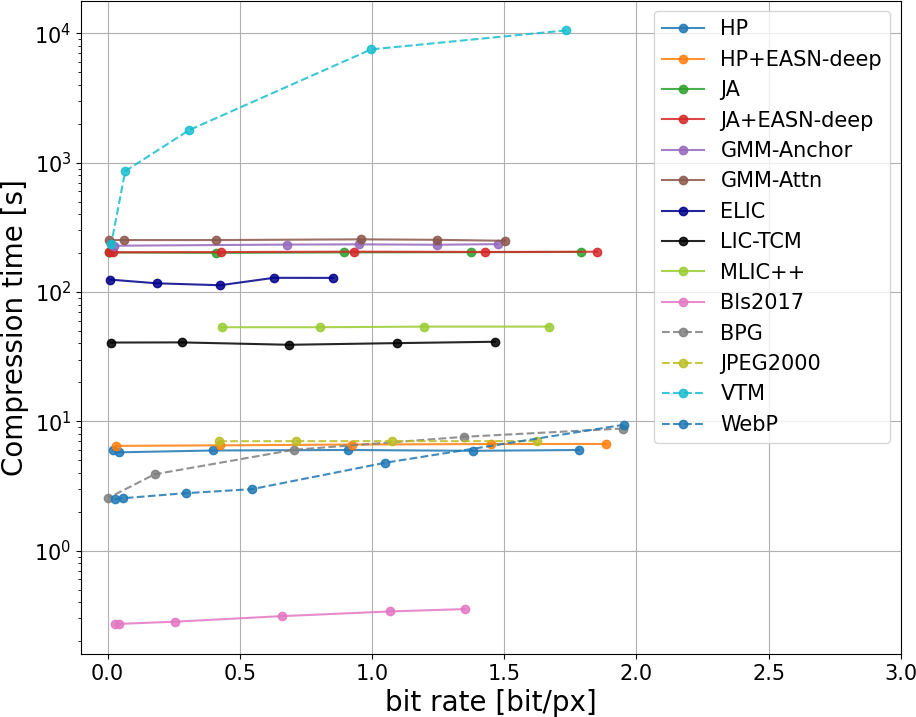} 
\includegraphics[width=3.3in]{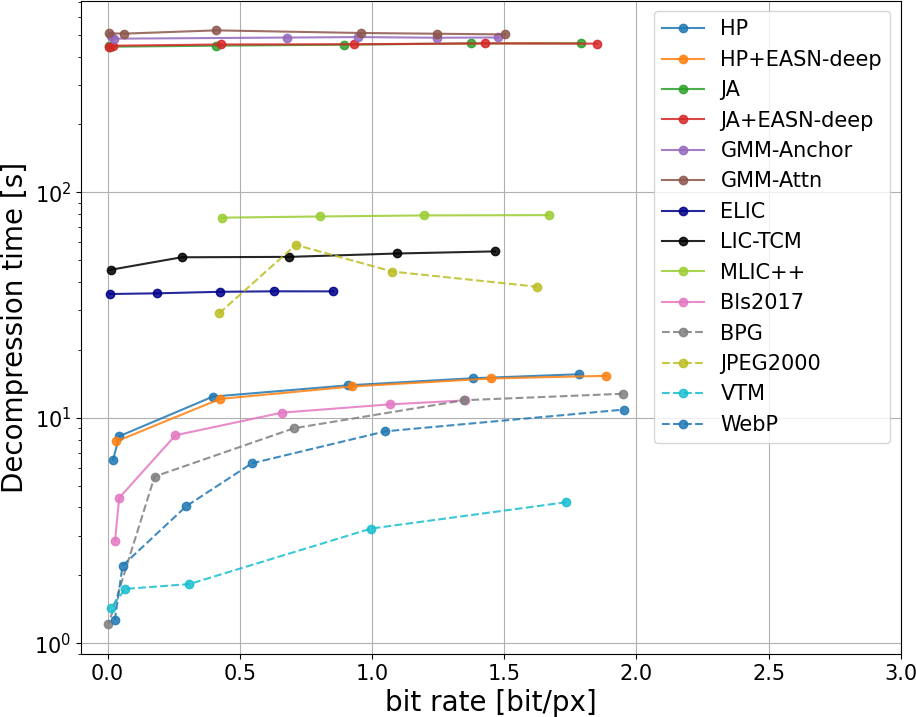}
\end{tabular}}    
\caption{\textbf{Compression and decompression time of a full-resolution image} for the distinct state-of-the-art lossy coders.} 
\label{fig:lossy-time}
\end{figure*}

\section{Bit-swap}

\subsection{Training Details} \label{sec:bitswap-training}

Before feeding the input images to the network, the images are squeezed following a multi-scale architecture~\cite{squeeze}. The mini-batch size for gradient descent is 128. The model is trained using a NVIDIA GeForce RTX 3080 Ti. We use the AdamW~\cite{adamw} optimizer with an initial learning rate of $10^{-4}$. The learning rate is decayed after $10^{5}$ gradient iterations with a scale factor of $0.99995$. This learning rate schedule is stopped when it decays down to the minimal value of $5 \cdot 10^{-5}$. The gradient norm is clipped to 1. To prevent posterior collapse we use the free-bits technique~\cite{free-bits}, the net bit rate of each latent variable is lower bounded by 1. 

To avoid overfitting, we randomly crop an image patch from an original full-resolution image in each epoch. We use weight normalization~\cite{weightnormalization} to restrict the Euclidean norm of the trainable parameters by a predefined scalar, and early-stopping. 

\subsection{Optimized Recursive Bits-back Algorithm} 

The detailed algorithm of our lossless routine is depicted in Algorithm~\ref{alg:bit-swap}. Specifically, note that immediately after decoding a latent variable $\mathbf{z}_l$ using the inference distribution
$q_{\boldsymbol{\phi}} (\mathbf{z}_{l} | \mathbf{z}_{l-1})$, we have access to the corresponding likelihood $p_{\boldsymbol{\theta}} (\mathbf{z}_{l-1} | \mathbf{z}_{l})$. Therefore, we can permute some
entropy operations of the conventional bits-back coding algorithm and perform an encoding operation after each decoding.

\begin{algorithm}[!t]
\caption{Optimized Recursive Bits-back coding \newline Compression} \label{alg:bit-swap}
\begin{algorithmic}[1]
\State \textbf{Given}: data $\mathcal{D} = \{\mathbf{x}_1, \ldots, \mathbf{x}_{|\mathcal{D}|}\}$, distributions $q_{\boldsymbol{\phi}}(\mathbf{z}_{l} | \mathbf{z}_{l-1}), p(\mathbf{z}_L),  p_{\boldsymbol{\theta}}(\mathbf{z}_{l-1} | \mathbf{z}_{l})$ for $l \in \{1, \ldots, L \}$, \item[] stack-like entropy encoder
\item[]
\State \textbf{Initialize}: bitstream
\Repeat
\State take $\mathbf{x} \in \mathcal{D}$
\State decode $\mathbf{z}_{1} \sim q_{\boldsymbol{\phi}}(\mathbf{z}_{1} | \mathbf{x})$ 
\State encode $\mathbf{x}$ with $p_{\boldsymbol{\theta}}(\mathbf{x} | \mathbf{z}_{1})$
\For {$l = 1$ to $L-1$}
\State decode $\mathbf{z}_{l+1} \sim q_{\boldsymbol{\phi}}(\mathbf{z}_{l+1} | \mathbf{z}_{l})$ 
\Comment{\footnotesize Grants access to \vspace{-0.1cm} \newline \rightline{$p_{\boldsymbol{\theta}}(\mathbf{z}_{l} | \mathbf{z}_{l+1})$ \hspace{0.55cm}}}
\normalsize
\State encode $\mathbf{z}_{l}$ with $p_{\boldsymbol{\theta}}(\mathbf{z}_{l} | \mathbf{z}_{l+1})$
\EndFor
\State encode $\mathbf{z}_{L}$ with $p(\mathbf{z}_L)$
\Until $\mathcal{D} = \varnothing$
\State \textbf{Send}: bitstream 
\end{algorithmic}
\end{algorithm}

\subsubsection{Discretization of the Latent Space} Every latent variable $\mathbf{z}_l = (z_{l,1}, \ldots, z_{l,2^P})$ gets discretized into $2^P$ bins, where $P$ is a predefined precision (in our case, $P=10$). All bins in each latent layer and each component $z_{l,j}$ have identical widths $\delta_{l,j}$ following a uniform binning~\cite{bb-ans}. Let $B(z_{l,j})$ be the bin that contains $z_{l,j}$, then we define the discretized $\bar{z}_{l,j}$ of $z_{l,j}$ as the centroid of the bin $B(z_{l,j})$. Thereby, the discretized probability of the inference $\bar{q}_{\boldsymbol{\phi}}(z_{l,j} | \mathbf{z}_{l-1})$ and its multivariate distribution are expressed by $\bar{q}_{\boldsymbol{\phi}}(\bar{z}_{l,j}| \mathbf{z}_{l-1}) = \int_{B(z_{l,j})} q_{\boldsymbol{\phi}}(z_{l,j} | \mathbf{z}_{l-1}) dz_j$ and $\bar{q}_{\boldsymbol{\phi}}(\bar{\mathbf{z}}_{l}| \mathbf{z}_{l-1}) = \prod_{j=1}^{2^P} %
     \bar{q}_{\boldsymbol{\phi}}(\bar{z}_{l,j}| \mathbf{z}_{l-1})$, respectively.

Consequently, the discretized inference model becomes sufficiently smooth for a large number of discretization bins $ \bar{q}_{\boldsymbol{\phi}}(\bar{{z}}_{l,j}| \mathbf{z}_{l-1}) \approx q_{\boldsymbol{\phi}}(\bar{{z}}_{l,j}| \mathbf{z}_{l-1}) \delta_{l,j} $. Note that the initial bits required grows with the number of discretization bins as $ -  \log \bar{q}_{\boldsymbol{\phi}} ({z}_{l,j} | \mathbf{z}_{l-1}) \approx  - \log q_{\boldsymbol{\phi}} ({z}_{l,j} | \mathbf{z}_{l-1}) - \log \delta_{l,j}  $, making that the optimized recursive bits-back scheme becomes even more necessary. Except for $p_{\boldsymbol{\theta}}(\mathbf{x} | \mathbf{z}_{1})$ that is already a discrete probability, the probability mass functions corresponding to the prior and likelihood distributions are also discretized. 

\subsection{Asymmetric Numeral Systems}

Asymmetric Numeral Systems (ANS)~\cite{ANS} is a recently developed entropy coding algorithm. Its variant rANS encodes a sequence of symbols as a natural number $s \in \mathds{N} $, such that the length of its binary representation closely matches the theoretical entropy of the probabilistic model $p$.

To illustrate this idea, we present a naive example with a 2-symbol alphabet $\mathcal{A} = \{ a_1, a_2\} $ with associated equiprobable probabilities $p_1 = p_2 = 0.5$. Let $s^t \in \mathds{N}$ denote the state after encoding the first $t$ symbols of a given source message, where $s^0 = 1$. A valid rANS scheme would be: 

\vspace{-0.45cm}
\begin{gather*}
    a_1 : s^t \longrightarrow 2s^{t-1} \hspace{0.62cm} \\
    a_2 : s^t \longrightarrow 2s^{t-1} + 1 .
\end{gather*}

This mapping just adds a $0$ to the binary compressed representation to encode $a_1$ and, a $1$ to encode $a_2$. Therefore, it is easily decodable: if the state $s^t$ is an even number, then the last encoded symbol is $a_1$. Otherwise, if the state $s^t$ is odd, we will decode $a_2$. After finding
out which was the last encoded symbol, the previous state $s^{t-1}$ can be easily calculated by:

\vspace{-0.45cm}
\begin{gather*}
    a_1 (s^t \text{ even}) : s^{t-1} \longrightarrow \frac{s^t}{2} \hspace{0.55cm} \\
    a_2 (s^t \text{ odd}) : s^{t-1} \longrightarrow \frac{s^t - 1}{2} .
\end{gather*}

An alternative operation for obtaining the last symbol that was encoded regardless of $s^t$ is $\floor{ \frac{s^t}{2} }$, where $\floor{ \cdot }$ denotes the floor function. Note that we have partitioned the natural numbers in two sets (the even and the odd numbers), depending on which last symbol was encoded, and the sets alternate such that they occur with probability approximately equal to $p_i$ in $\mathds{N}$. This is the cornerstone that allows that the binary length of $s^t$ is close to the entropy $H(p)$.

In the general case, where we have an arbitrary number of symbols in the alphabet $\mathcal{A} = \{a_1, \cdots, a_{| \mathcal{A} |} \}$ and its symbol $a_i$ occurs with probability $p_i$, we also partition the natural numbers in $| \mathcal{A} |$ sets and associate each set $S_i \subset
\mathds{N}$ to the symbol $a_i$. Following the example design, each natural number will belong to one set $S_i$ denoting that the last encoded symbol is $a_i$ and, more precisely, the elements of this subset $S_i$ must appear in $\mathds{N}$ with probability $p_i$.

This is materialized by scaling up the probabilities $f_i \propto p_i^{-1}$, sharing the same proportional rate between all $f_i$, plus some minor deviations to ensure $f_i \in \mathds{N}$. The proportional ratio is chosen such that the scaling up probabilities sum to a preset multiplier $M$: $\sum_{i=1}^{| \mathcal{A} |} f_i = M$. Additionally, we define several subsets $\{K_1, K_2, \ldots\}$ that take ranges of M consecutive natural numbers. Specifically, $K_1$ will take the first $M$ natural numbers, the next $M$ numbers will belong to $K_2$, and so on. 

Then, in every partition $K_j$, we assign the first $f_1$ numbers ($\approx M p_1$) to $S_1$ and constitute $S_{j1}$, the second $f_2$ numbers to $S_2$ and constitute $S_{j2}$, etc. Thus, the set $S_i$ of associated numbers with the last encoded symbol $a_i$ is composed by $S_i = \bigcup_{j=1}^{\infty} S_{ji}$ and, truly, all $S_i$ partition the natural numbers. Moreover, the elements of $S_i$ occur in $\mathds{N}$ with a probability approximately equal to $p_i$.

Let $s^{t-1}$ be the current initial state. The rANS algorithm encodes a symbol $a_i$ and obtains its corresponding natural number $s_t$ by selecting the $s^{t-1}$-th occurrence in the set $S_i$. This is accomplished by the following formula:

\vspace{-0.45cm}
$$ s^t = \text{rANS}_{\text{encode}}(a_i, s^{t-1}) = M \floor{\frac{s^{t-1}}{f_i}} + B_i + (s^{t-1} \mod f_i) ,$$
where $B_i = \sum\limits_{i'=1}^{i-1} f_{i'}$ is the accumulated scaled up probability function.

Conversely, to decode an initial state $s^t$, we need to figure out which subset $S_i$ it belongs to. Notice that rANS is decodable since any natural number can be uniquely identified with one of the symbols $a_i$. Subsequently, once we know which was the last encoded symbol $a_i$, we can recover the previous state $s^{t-1}$ by doing a look-up for $a_i$ in the subset $S_i$. Indeed, the index of $a_i$ in $S_i$ corresponds to the state $s^t$ that preceded $s^{t-1}$. Similarly to the encoder, this operation can be accomplished by:

\begin{align*}
    (a_i, s^{t-1}) &= \text{rANS}_{\text{decode}}(s^t) \\
    &= \bigg( \argmax_{ 1 \leq i \leq | \mathcal{A} | }\{B_i < R \}, f_i \floor{\frac{s^t}{M}} + R - B_i \bigg),
\end{align*} 
where $R = s^t \mod M $. Note that the decoder returns both the symbol and the state. \newline

This entropy coder is an optimal compression scheme by construction. If rANS encodes a symbol $a_i$, the natural number $s^t$ that is obtained is in the order of $\frac{s^{t-1}}{p_i}$. So, let $s^0$ denote the initial state with an insignificant code length, after encoding a sequence of symbols $a_{i_1}a_{i_2}\ldots a_{{i_N}}$, we will end up with a state $s^{i_N} \approx \dfrac{s_0}{p_{i_1}p_{i_2}\ldots p_{i_N}}$. Hence, the corresponding bit length is: 

$$ \log s^{i_N} \approx \log s^0 + \sum_{j=i_1}^{i_N} \log \frac{1}{p_{j}} . $$

By dividing by the number of symbols encoded $N$, we obtain the average code length which clearly approaches the entropy of $p$ when $N$ is large enough.

If we compare rANS with arithmetic coding~\cite{arithmetic}, we can observe that they differ in the order in which messages are decoded. Arithmetic coding works as a queue structure, i.e., the last symbols that are encoded are the last symbols to be decoded. On the other hand, rANS is stack-like, due to decoding in the opposite order that it encodes. Since rANS only requires a few arithmetic modular operations, it has replaced arithmetic coders as the reference entropy coder for its stunning computation speed.

\subsection{Initial Bits Impact} \label{sec:init_bits}

The effect of the initial bits is analyzed by monitoring the bitstream length in each coding step. Figure~\ref{fig:cum-bpp} presents the average cumulative bit rate for an image patch of 32$\times$32 pixels and $L=4$ as an example. The figure compares the two recursive bits-back algorithms: the conventional (left figure) and optimized version (middle figure). Both methods end with the same net bitstream length. But, as the conventional algorithm firstly performs the decoding operations, the cumulative bit rate reaches the most negative value after sequentially decoding all the latent variables. In particular, the conventional recursive bits-back requires an average of $Bit_{\text{init}}^{\text{recursive}}= 58.46$ bit/px initial bits, while the optimized version reduces this number to $Bit_{\text{init}}^{\text{optimized}} =25.78$ bit/px - more than double of difference.

\begin{figure*}[!t]
\centering
\resizebox{18cm}{!}{
\begin{tabular}{@{}c@{}}   
\includegraphics[width=3.3in]{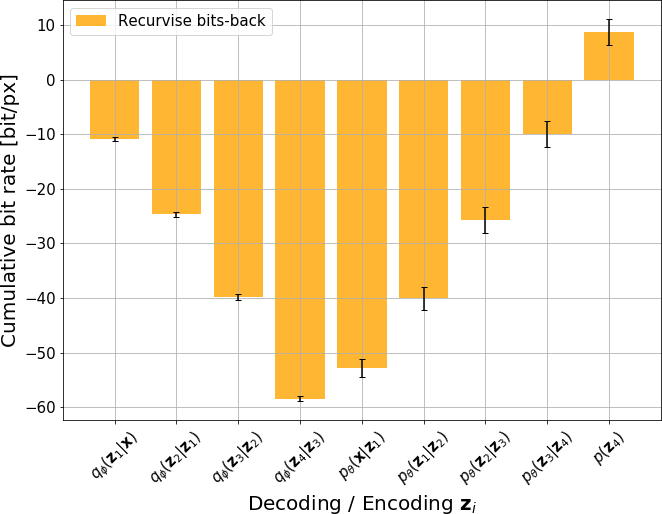} 
\includegraphics[width=3.3in]{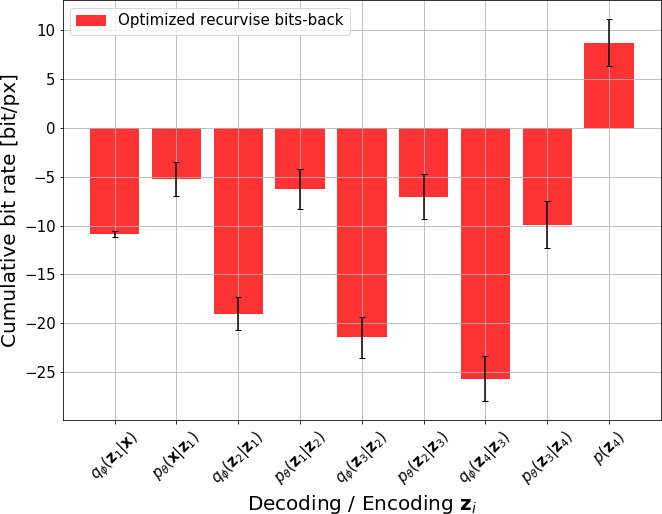}
\includegraphics[width=3.3in]{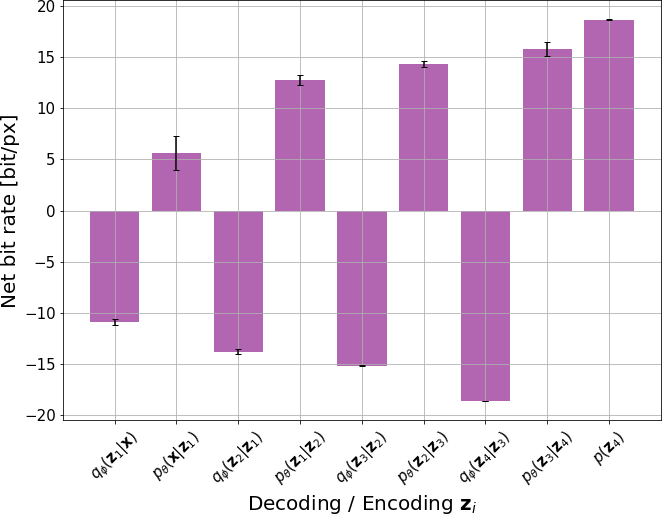}
\end{tabular}}    
\caption{\textbf{Cumulative average bit rate} of the distinct coding steps when applying the recursive bits-back algorithm (left) or the optimized version (middle). \textbf{Average net bit rate} of the distinct coding steps (right). The minimum negative bit rate denotes the required initial bits for a single image patch.} 
\label{fig:cum-bpp}
\end{figure*}

The bitstream length varies more when coding deeper latent variables. Therefore, bit-swap is transmitting most of the image information to the deepest latent variables. For instance, the prior is the distribution that increases  the bitstream length more; when encoding $\mathbf{z}_4$. This behavior is also observed in the learning curves depicted in Section~\ref{app:learning-curves} and is further presented in Section~\ref{app:net-bpp}, which shows the average net bit rate of the distinct coding steps.

The influence of the initial bits on the bit rate can also be analyzed for full-resolution images, whose image patches are encoded sequentially in raster order, such that the initial random bits can be taken from the previous image patch. Figure~\ref{fig:init-bpp} presents how many patches are required to obtain a bit rate close to the net bit rate, i.e., the ELBO which is represented in the purple dotted line. The plot compares the conventional recursive bits-back scheme and the optimized version. Due to the optimized recursive bits-back starts with lower initial bits, the effect of the initial bits is quickly canceled. The conventional version requires more patches to accomplish the same performance.

\begin{figure}[!t]
\centering
\hspace{-0.1cm} \includegraphics[width=2.9in]{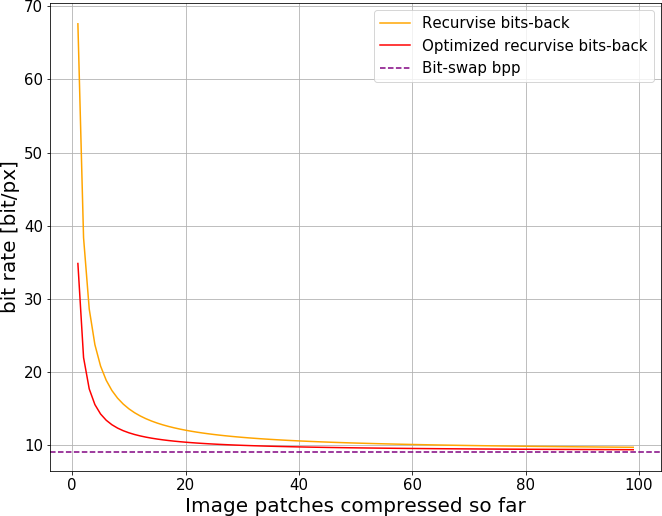}
\caption{\textbf{Bit rate as a function of the number of image patches} that have been sequentially compressed.}
\label{fig:init-bpp}
\end{figure}

\subsection{Net Bit Rate per Latent Distribution}  \label{app:net-bpp}

Figure~\ref{fig:cum-bpp}-right illustrates the net bit rate of each coding operation for $L=4$ and an image patch of 32$\times$32. Note that the order of the encoding/decoding operations does not condition the net bit rate of each coding step. For instance, the resulting bit rate after encoding $\mathbf{x}$ using $p_{\boldsymbol{\theta}}(\mathbf{x}|\mathbf{z}_1)$ is equal for the conventional recursive bits-back and for the optimized version.

The variability in bitstream length is more pronounced when coding deeper latent variables. This highlights the predominant role of the model in conveying image information to the deepest latent variables. Notably, the prior distribution contributes the most to the increase in bitstream length.

\subsection{Learning Behavior} \label{app:learning-curves}

\begin{figure}[!t]
\centering
\resizebox{8.5cm}{!}{
\begin{tabular}{@{}c@{}}   
\includegraphics[width=2.85in]{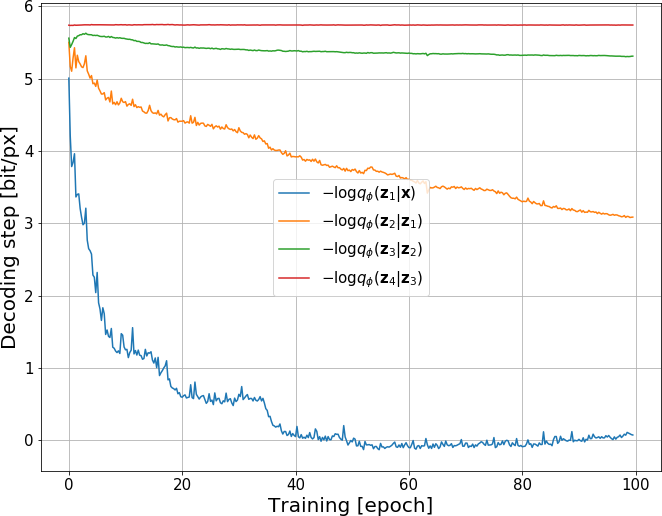} \\
\includegraphics[width=2.9in]{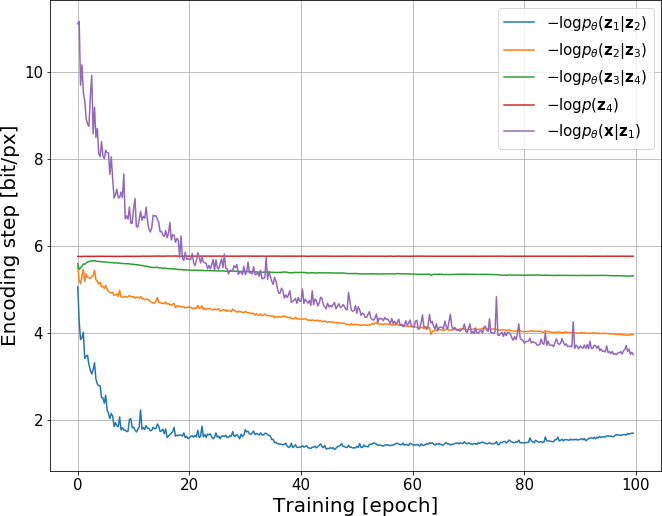} 
\end{tabular}}           
\caption{\textbf{Learning curves of each individual distribution} when training on the validation dataset. Notice the inference curves are minimized when our desire is to maximize them. \textbf{Top:} Inference distributions (decoding steps). \textbf{Bottom:} Likelihood and prior distributions (encoding steps).}
\label{fig:bitswap-learning-indiv}
\end{figure}

By breaking down the loss into each coding distribution, we can examine the behavior of the variational inference optimization problem in more depth. To illustrate the results better, we analyze Bit-swap for $L=4$ latent variables and a patch size of 32$\times$32. Figure~\ref{fig:bitswap-learning-indiv}-top shows the learning curves for the decoding distributions, i.e., the inference distributions $q_{\boldsymbol{\phi}}(\mathbf{z}_{l+1} | \mathbf{z}_{l})$ for $l \in \{0, 1, 2, 3 \}$, $ \mathbf{z}_{0} = \mathbf{x}$. Figure~\ref{fig:bitswap-learning-indiv}-bottom shows the learning curves for the encoding distributions, i.e., the prior $p(\mathbf{z}_{4})$ and likelihood distributions $p_{\boldsymbol{\theta}}(\mathbf{z}_{l} | \mathbf{z}_{l+1})$ for $l \in \{0, 1, 2, 3 \}$, $ \mathbf{z}_{0} = \mathbf{x}$. Each decoding distribution is nested to an encoding distribution (indicated with the same colors in Figure~\ref{fig:bitswap-learning-indiv}), because decreasing (increasing) the likelihood of one implies decreasing (increasing) the likelihood of its nested distribution. For instance, $q_{\boldsymbol{\phi}}(\mathbf{z}_{1} | \mathbf{x})$  is nested to $p_{\boldsymbol{\theta}}(\mathbf{z}_{1} | \mathbf{z}_{2})$. These relationships impede that the decoding distributions are maximized and at the same time the encoding distributions are minimized. For this reason, the inference distributions learning curves decreases throughout training, when we would expect the opposite. Notice that $p_{\boldsymbol{\theta}}(\mathbf{z}_{1} | \mathbf{z}_{2})$ is not nested to any distribution, so it has more freedom in its optimization. 

The likelihood of deeper latent variables is harder to optimize. Indeed, the likelihood of the nested distributions $q_{\boldsymbol{\phi}}(\mathbf{z}_{4} | \mathbf{z}_3)$ and $p(\mathbf{z}_{4})$ is stable during the whole training. The incapability to optimize the deepest latent variable $\mathbf{z}_4$ may suggest that the model ignores it, resulting in a trivial inference distribution $q_{\boldsymbol{\phi}}(\mathbf{z}_4 | \mathbf{z}_3)$ that collapses to the prior $p(\mathbf{z}_4)$ - what is known as the posterior collapse phenomenon. This may indicate that the optimum number of latent variables is smaller than $L=4$. Therefore, the free-bits technique is not successfully mitigating posterior collapse for a 4-layer model. 

Note that $- \log q_{\boldsymbol{\phi}}(\mathbf{z}_1 | \mathbf{x})$ takes negative values, which seems counterintuitive as no symbol can be encoded in a bitstream of negative length. However, $q_{\boldsymbol{\phi}}(\mathbf{z}_1 | \mathbf{x})$ denotes a continuous density distribution, so it can take any positive value, implying that $- \log q_{\boldsymbol{\phi}}(\mathbf{z}_1 | \mathbf{x})$ can be negative. Indeed, until the distributions are not discretized, it should not be understood that the distributions correspond to the bitstream length.

\subsection{Ablation Study}

We investigated the input patch size $PS$ in order to optimize the receptive field of the network, and the number of latent variables $L$ for optimizing the model's performance. Through hyperparameter fine-tuning over the validation set, we can observe that for $PS=32$ and $64$, the optimal is $L=2$ as depicted in Fig.~\ref{fig:bitswap-crop-nz}, with $PS=64$ in the lead. As expected, when increasing the $PS$ to 128, we require more latent variables, because the model needs to capture a higher complexity of the data distribution when dealing with larger inputs. A larger image size could imply that there are more subtle variations in the data that need to be represented by the latent variables. In particular the compression performance reaches its optimal for $PS=128$ when $L=3$. However, for such $PS=128$, the bit rate is always higher than for $PS=32$ and $64$.

The number of training parameters of Bit-swap is chosen similar for any $L$, resulting always in a quite complex and deep network. This complexity caused certain instability during the training, which was controlled through the batch size and the learning rate; apart from the techniques mentioned in the training details from Section~\ref{sec:bitswap-training}. Nonetheless, this unstable behavior caused a distinct general pattern between the different $PS$. For instance, we observe in Fig.~\ref{fig:bitswap-crop-nz} that $L=3$ achieves a lower performance than $L=4$ for $PS=32$, which is not the case for $PS=64$.

\begin{figure}[!t]
\centering
\includegraphics[width=2.9in]{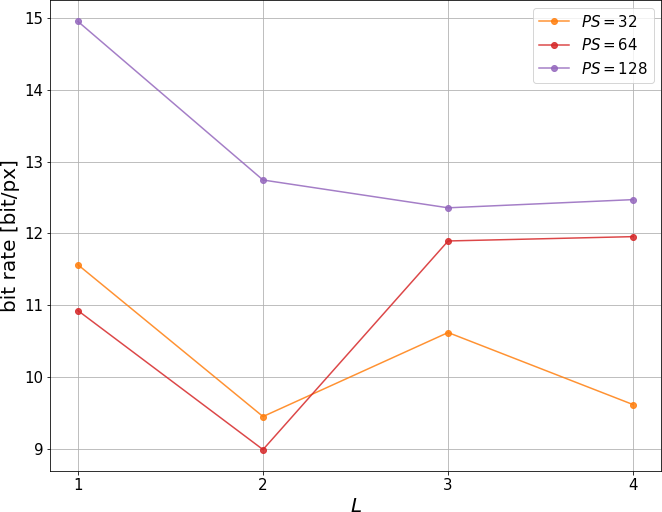}
\caption{\textbf{Ablation study of the number of latent variables $L$ and the patch size $PS$} over the Bit-swap model.} \vspace{0.cm}
\label{fig:bitswap-crop-nz}
\end{figure}

\begin{figure}[!t]
\centering
\includegraphics[width=2.9in]{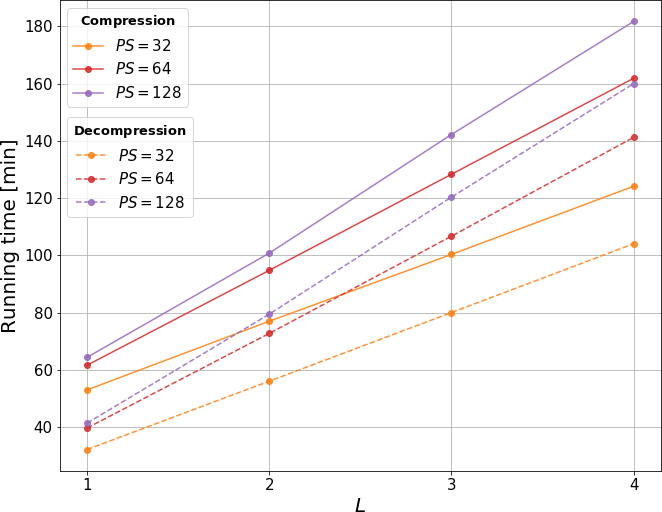}
\caption{\textbf{Compression and decompression times in terms of the number of latent variables $L$ and the patch size $PS$} over the Bit-swap model.} 
\label{fig:bitswap-crop-nz-time}
\vspace{-.3cm}
\end{figure}

We analyze the effect of the patch size $PS$ and the number of latent variables $L$ in Fig.~\ref{fig:bitswap-crop-nz-time}. The computational cost is mainly determined by the inference time of the encoder $h_{\boldsymbol{\phi}}$ and decoder $g_{\boldsymbol{\theta}}$ transforms, which runs on the GPU, and by ANS entropy coder, which runs in the CPU. The decompression time is always lower than the compression one, because ANS requires less operations for decoding. In addition, despite maintaining a constant amount of parameters for any $L$, an increased quantity of latent variables results in greater computational requirements, since ANS is executed more times and there is more GPU and CPU memory swapping.

\begin{figure*}[!t]
\centering
\resizebox{17.8cm}{!}{
\begin{tabular}{cc}  
\includegraphics[width=2.in]{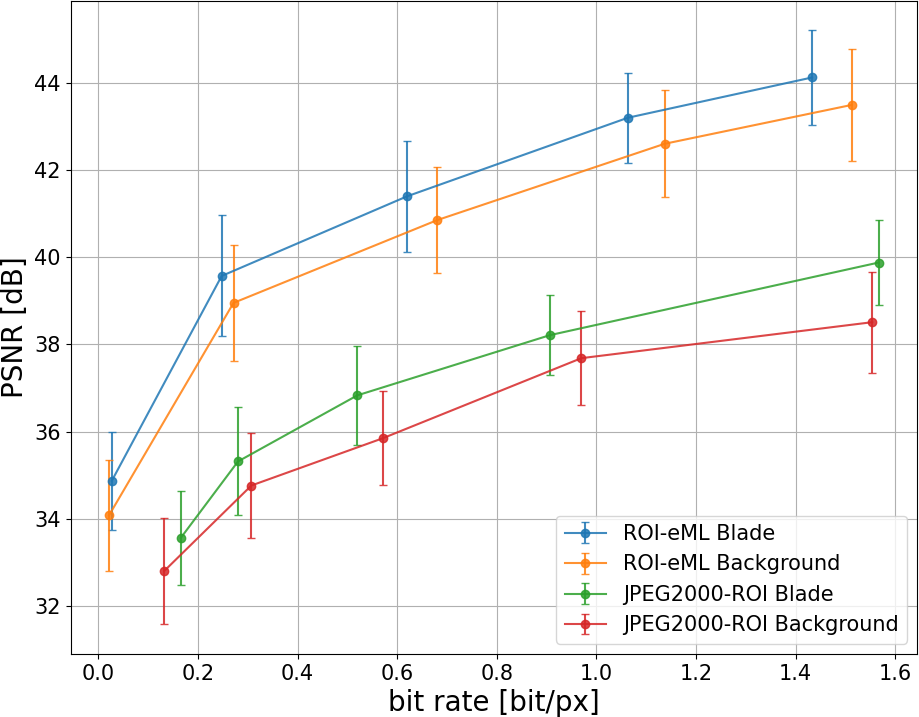}
\includegraphics[width=2.in]{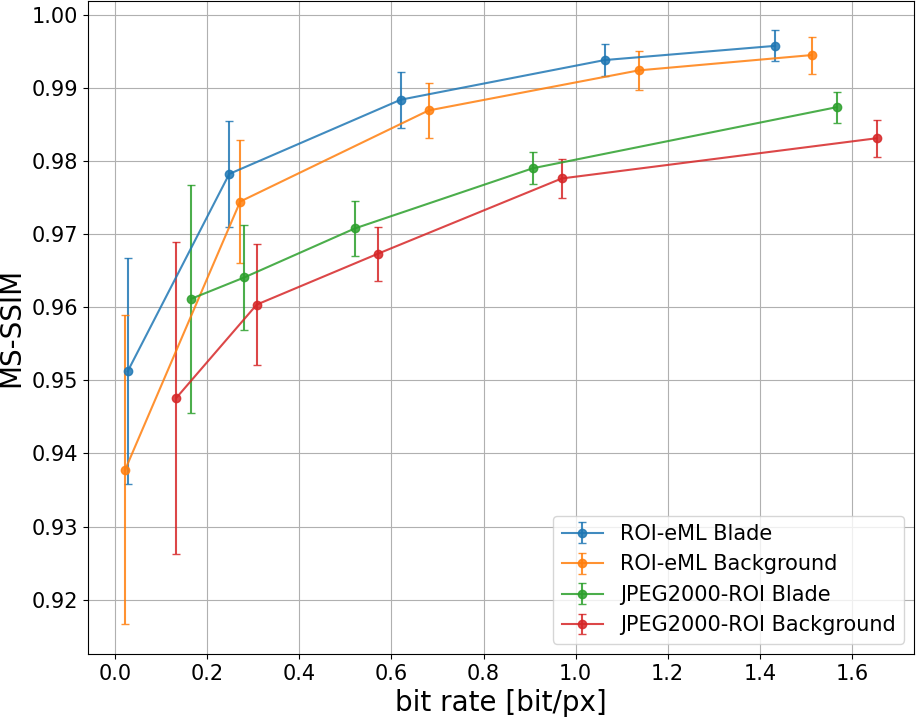}
 \end{tabular}}     
 \vspace{-.3cm}
\caption{\textbf{Rate-distortion curves per region of ROI-eML}; HP+EASN-deep trained for squared error. The error bars indicate the std distortion on the compression performance. PSNR (left) and MS-SSIM (right) quality measures.} \vspace{-.3cm}
\label{fig:roi-curves}
\end{figure*}

\subsection{Traditional Codec Settings}

This subsection details the settings used to evaluate the performance of standard lossless image compression codecs: PNG~\cite{png}, JPEG2000~\cite{J2K}, WebP~\cite{webp}, BPG~\cite{bpg}, and JXL~\cite{jxl}.

For PNG, the images are converted from Canon RAW (CR2) format using \texttt{rawpy} version 0.24.0, with 8-bit bit per pixel allocation:

\begin{HighlightVerbatim}
with rawpy.imread("input.CR2") as img_cr2:
    img = img_cr2.postprocess(output_bps=8)
img = Image.fromarray(img, 'RGB')
\end{HighlightVerbatim}

Then, the images are stored as-is in the PNG format using \texttt{PIL.Image.save()}. No additional compression is applied beyond the default lossless encoding.

For JPEG2000, we use the \texttt{OpenJPEG} library (v2.5.0) in default lossless mode. The compression and decompression are performed with:

\begin{HighlightVerbatim}
opj_compress -i input.png -o output.j2k
opj_decompress -i output.j2k -o output.png
\end{HighlightVerbatim}

For WebP, we use \texttt{libwebp} version 1.2.2. Lossless compression is enforced with the \texttt{-lossless} flag:

\begin{HighlightVerbatim}
cwebp input.png -o output.webp -q 100 -m 6 
-lossless dwebp output.webp -o output.png
\end{HighlightVerbatim}

For BPG, lossless mode is enabled using the \texttt{-lossless} flag. We use \texttt{libbpg} version 0.9.8:

\begin{HighlightVerbatim}
bpgenc -lossless input.png -o output.bpg
bpgdec output.bpg -o output.png
\end{HighlightVerbatim}

For JPEG XL, we use \texttt{libjxl} tools \texttt{cjxl} and \texttt{djxl}, setting quality to 100 and effort level to 9 for maximum compression efficiency:

\begin{HighlightVerbatim}
cjxl input.png output.jxl -q 100 -e 9 --quiet
djxl output.jxl output.png --quiet
\end{HighlightVerbatim}

\begin{table}[!t]
\centering
\begin{tabular}{cccc}
\toprule
     Coding & \multicolumn{1}{c}{Region} &  \multicolumn{1}{c}{Bit rate} &  \multicolumn{1}{c}{Bit rate} \\
    scheme & \multicolumn{1}{c}{} &   \multicolumn{1}{c}{mean} & \multicolumn{1}{c}{std}  \\ \midrule
   JPEG2000-ROI~\cite{J2K} & Blade &  10.06 & 1.68  \\
   JPEG2000-ROI~\cite{J2K}  & Background & 10.31 & 1.72 \\
   
   ROI-eML & Blade & 8.84  &  1.68 \\
   ROI-eML  & Background & 9.06 & 1.72 \\
\bottomrule
\end{tabular}
\caption{\textbf{Lossless compression performance comparison of ROI-eML per region}, implemented by means of Bit-swap.} \label{tab:roi-lossless}
\end{table}

\section{ROI-eML}

\subsection{Compression Performance per Region}  \label{app:roi-region}

To further analyze the regional behavior of the proposed dual-compression scheme, we report separate rate–distortion (RD) curves for the blade (ROI) and the background regions.  

Fig.~\ref{fig:roi-curves} compares the performance of ROI-eML against the JPEG2000-ROI~\cite{J2K} baseline using the HP+EASN-deep model trained with a squared error objective.  For each method, the PSNR and MS-SSIM are computed independently inside and outside the ROI. The plots clearly show that ROI-eML achieves a higher reconstruction quality in the ROI (and non-ROI) region for the same total bitrate compared to JPEG2000-ROI, indicating a more efficient bit allocation. 
The graph shows that the blade region is compressed more efficiently than the background. This is because the background of an image has more variability, while the blade region is fairly homogeneous. Moreover, the blade region is always present, thus, the network is capable to learn how to compress this area. 

The behavior is consistent in the lossless regime. Table~\ref{tab:roi-lossless} summarizes the bit rates achieved by Bit-swap for both regions. As the lossy case, the background requires slightly more bits due to its higher structural variability, while the blade region, being more homogeneous and consistently present across the dataset, is compressed more efficiently. Overall, ROI-eML achieves lower mean bit rates than JPEG2000-ROI~\cite{J2K} in both regions, confirming the benefit of the learned representation in selectively optimizing compression according to spatial relevance.

\end{document}